\documentclass{article} 
\usepackage{iclr2026_conference,times}

\usepackage{amsmath,amsfonts,bm}

\def\eqref#1{equation~\ref{#1}}
\def\Eqref#1{Equation~\ref{#1}}

\def\1{\bm{1}}

\DeclareMathAlphabet{\mathsfit}{\encodingdefault}{\sfdefault}{m}{sl}
\SetMathAlphabet{\mathsfit}{bold}{\encodingdefault}{\sfdefault}{bx}{n}

\usepackage[utf8]{inputenc} 
\usepackage[T1]{fontenc}    
\usepackage{hyperref}       
\usepackage{url}            
\usepackage{booktabs}       
\usepackage{amsfonts}       
\usepackage{nicefrac}       
\usepackage{microtype}      
\usepackage{xcolor}         
\usepackage{amsmath,amssymb,mathrsfs}
\usepackage{subcaption}
\usepackage{comment}
\usepackage{graphicx}
\usepackage{bm}
\usepackage[capitalize]{cleveref}
\usepackage{float}
\usepackage{multirow}

\newcommand{\simage}{s^{\rm image}}
\newcommand{\stext}{s^{\rm text}}
\newcommand{\bfull}{b_{j}(s)}
\newcommand{\bbase}{b_{j}^{\rm base}}
\newcommand{\bimage}{b_{j}^{\rm image}}
\newcommand{\btext}{b_{j}^{\rm text}}
\newcommand{\bsynergy}{b_{j}^{\rm cross}}
\newcommand{\afull}{a_{j}(s)}
\newcommand{\abase}{a_{j}^{\rm base}}
\newcommand{\aimage}{a_{j}^{\rm image}}
\newcommand{\atext}{a_{j}^{\rm text}}
\newcommand{\asynergy}{a_{j}^{\rm cross}}
\newcommand{\tfull}{\theta_{i}(s)}
\newcommand{\tbase}{\theta_{i}^{\rm base}}
\newcommand{\timage}{\theta_{i}^{\rm image}}
\newcommand{\ttext}{\theta_{i}^{\rm text}}
\newcommand{\tsynergy}{\theta_{i}^{\rm cross}}
\newcommand{\rij}{r_{i,j}}
\newcommand{\rijs}{r_{i,j,s}}
\newcommand{\zijs}{z_{i,j,s}}
\newcommand{\mmirt}{{M2}IRT}
\newcommand{\mmmirt}{{M3}IRT}

\newcommand{\ie}{i.\,e.\ }

\newcommand{\SU}[1]{\textcolor{black}{#1}}

\title{Evaluating Cross-Modal Reasoning Ability and Problem Characteristics with Multimodal Item Response Theory}

\author{
\textbf{Shunki Uebayashi$^{1}$, 
Kento Masui$^{2}$, 
Kyohei Atarashi$^{1}$ , 
Han Bao$^{3,4}$, 
Hisashi Kashima$^{1}$,}\\
\textbf{ 
Naoto Inoue$^{2}$, 
Mayu Otani$^{2}$, 
Koh Takeuchi$^{1}$} \\
$^{1}$Kyoto University 
$^{2}$CyberAgent. 
$^{3}$The Institute of Statistical Mathematics  
$^{4}$Tohoku University
}

\iclrfinalcopy 
\begin{document}

\maketitle

\begin{abstract}
Multimodal Large Language Models (MLLMs) have recently emerged as general architectures capable of reasoning over diverse modalities.
Benchmarks for MLLMs should measure their ability for cross‑modal integration. However, current benchmarks are filled with shortcut questions, which can be solved using only single modality, and thereby yielding unreliable rankings.
For example, in vision-language cases, we can find the correct answer without either the image or the text.
These low-quality questions unnecessarily increase the size and computational requirements of benchmarks.
We introduce a multi-modal and multidimensional item response theory framework (\mmmirt{}) that extends classical IRT by decomposing both model ability and item difficulty into image‑only, text‑only, and cross‑modal components. \mmmirt{} estimates cross‑modal ability of MLLMs and each question’s cross‑modal difficulty, enabling compact, high‑quality subsets that better reflect multimodal reasoning. 
Across 24 VLMs on three benchmarks, \mmmirt{} prioritizes genuinely cross‑modal questions over shortcuts and preserves ranking fidelity even when 50\% of items are artificially generated low‑quality questions, thereby reducing evaluation cost while improving reliability. \mmmirt{} thus offers a practical tool for assessing cross‑modal reasoning and refining multimodal benchmarks. 

\end{abstract}

\section{Introduction}
\label{sec:intro}
Multimodal Large Language Models (MLLMs)~\citep{10.1093/nsr/nwae403} have recently emerged as general architectures capable of reasoning over diverse modalities. A prominent subclass, Visual–Language Models (VLMs), jointly process images and text and are expected to support downstream tasks that require cross‑modal reasoning~\citep{Jiang_2023_CVPR}, such as medical image diagnosis and industrial inspection~\citep{10445007}. Consequently, rigorous and trustworthy multimodal benchmarks are essential for practitioners to choose appropriate models~\citep{NEURIPS2024_2f8ee6a3,yue2024mmmu}.

Benchmarks for MLLMs should measure their ability for cross‑modal integration.
However, current benchmarks are often filled with shortcut questions that can be solved using only single modality (e.g., answerable from text alone or image alone). 
For example, in vision-language cases, we can find the correct answer without either the image or the text.
These low-quality questions unnecessarily increase the size and computational requirements of a benchmark and yields unreliable rankings~\citep{yue2024mmmu}.
As the pool of candidate models grows, evaluating thousands of mixed‑quality questions per model becomes increasingly costly, while single‑modality shortcuts further obstacle evaluating the cross‑modal reasoning ability.

Item Response Theory (IRT) is a principled framework for assessing subject ability and item difficulty~\citep{doi:10.1177/0013164498058003001}.
Without knowing the questions and answers, IRT estimates the ability and difficulty as parameters to predict the records of success or failure of a subject on an item.
These parameters allow us to construct a compact subset of items tailored to each subject using Computerized Adaptive Testing (CAT)~\citep{9c2d3927-7193-3702-9179-0a57b6a5e7e0,cat}.
Recent work on LLM has leveraged IRT, where they considered LLM as subject and questions as items, to construct compact and essential subsets of text questions from benchmarks~\citep{polo2024tinybenchmarksevaluatingllmsfewer}. 
However, classical IRT is agnostic to the modality of inputs and thus contains only a single latent ability or difficulty parameter.
IRT cannot determine whether success on a multimodal item reflects true cross-modal reasoning or others.

To address the limitations, we introduce MultiModal and Multidimensional Item Response Theory (\mmmirt{}), and its variant called \mmirt{}. Our proposed methods simply extend classical IRT by decomposing both model ability and item difficulty into three latent components: image‑only, text‑only, and cross‑modal integration. 
This decomposition allows us to (i) estimate each VLM’s cross‑modal ability and (ii) quantify each question’s cross‑modal difficulty. 
Using these estimates, our proposed methods identifies genuinely cross‑modal items and enables compact, high‑quality benchmark subsets that better reflect multimodal reasoning while reducing evaluation cost.

\begin{figure}[t]
    \centering
    \begin{subfigure}[b]{0.32\textwidth}
        \centering
        \includegraphics[width=\textwidth]{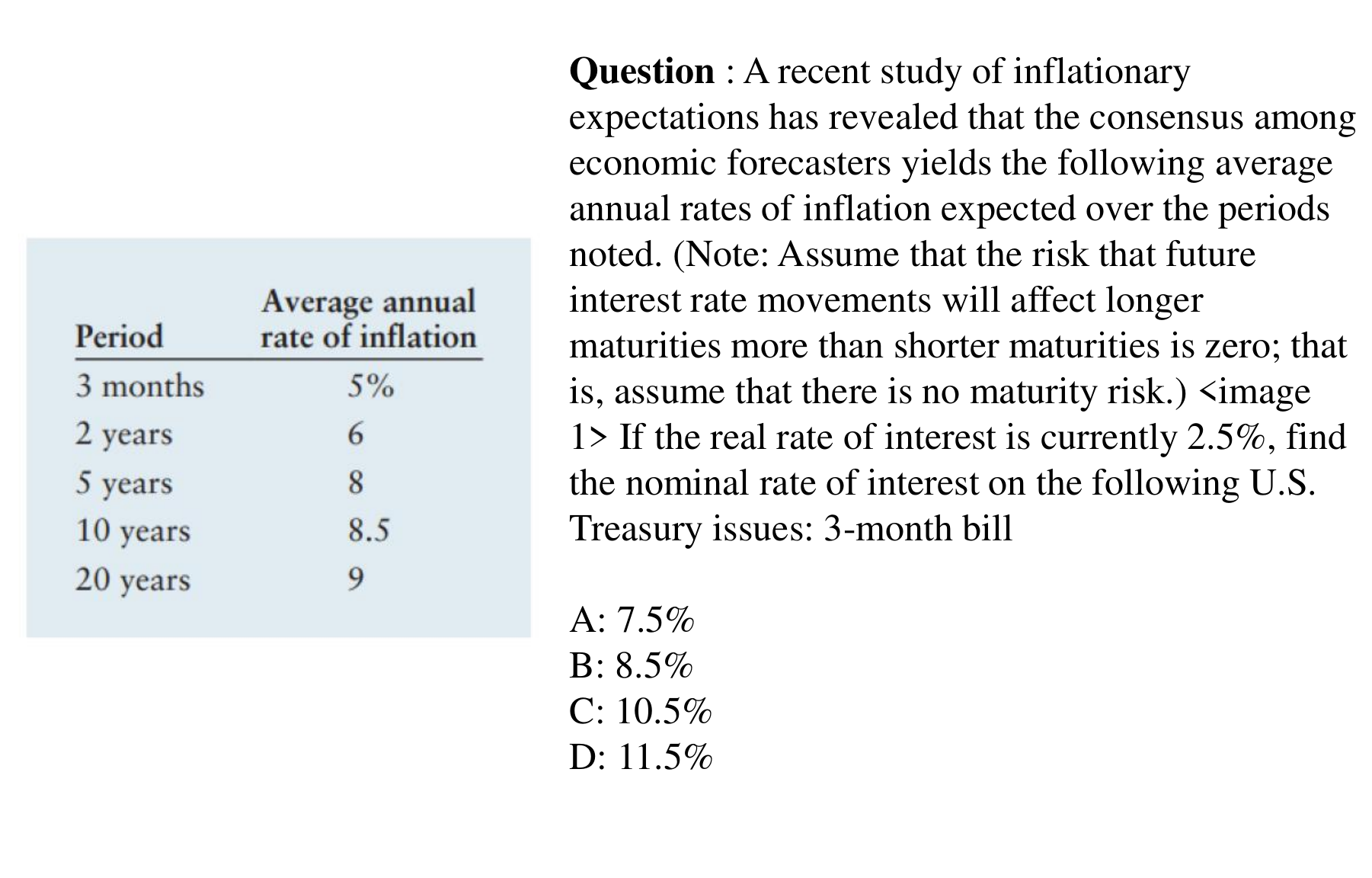}
        \caption{MMMU (Highly quality)}
        \label{fig:high-synergy-3ds-mmmu}
    \end{subfigure}
    \begin{subfigure}[b]{0.32\textwidth}
        \centering
        \includegraphics[width=\textwidth]{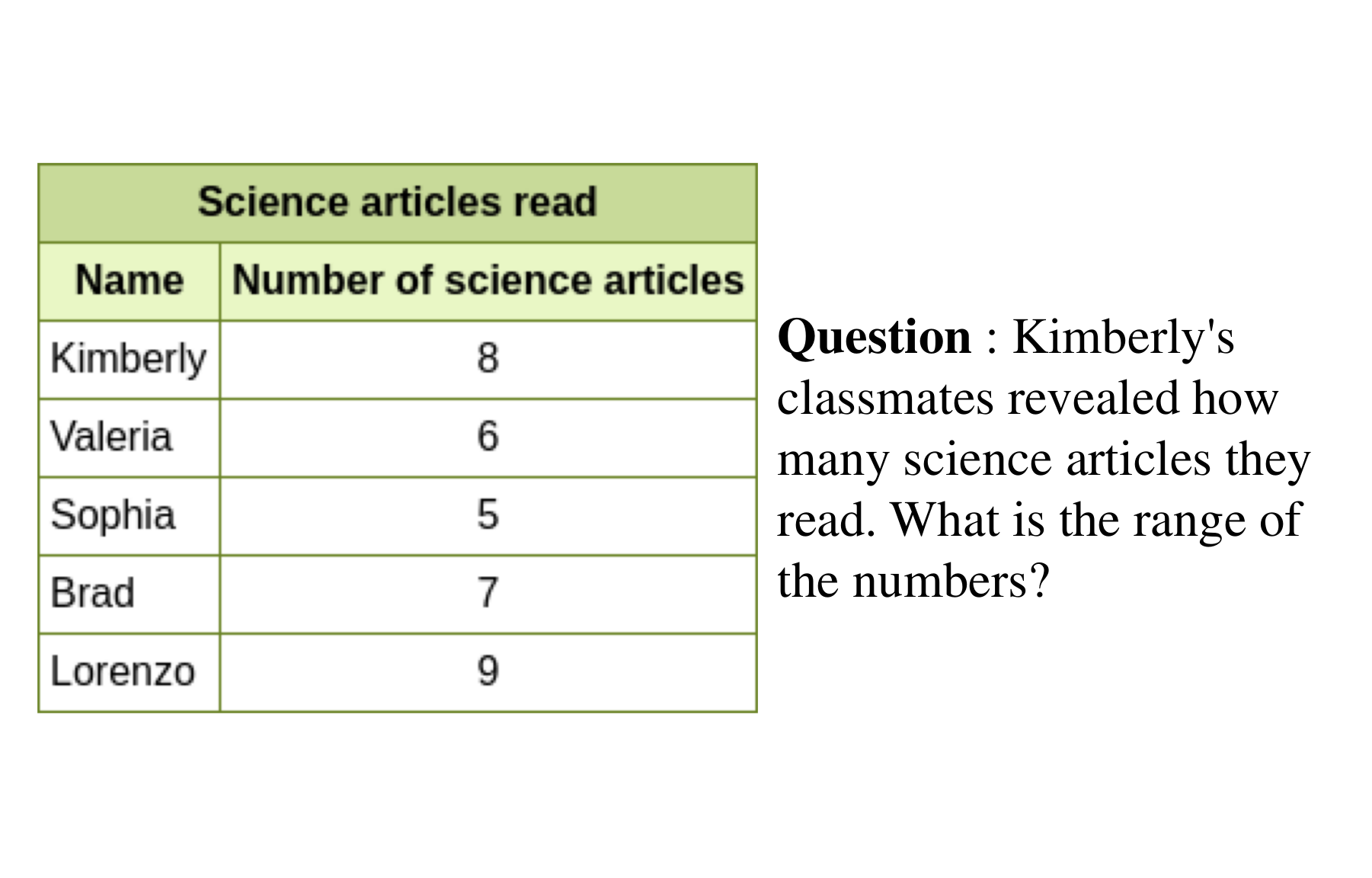}
        \caption{Mathvista (High quality)}
        \label{fig:high-synergy-3ds-mathvista}
    \end{subfigure}
    \begin{subfigure}[b]{0.32\textwidth}
        \centering
        \includegraphics[width=\textwidth]{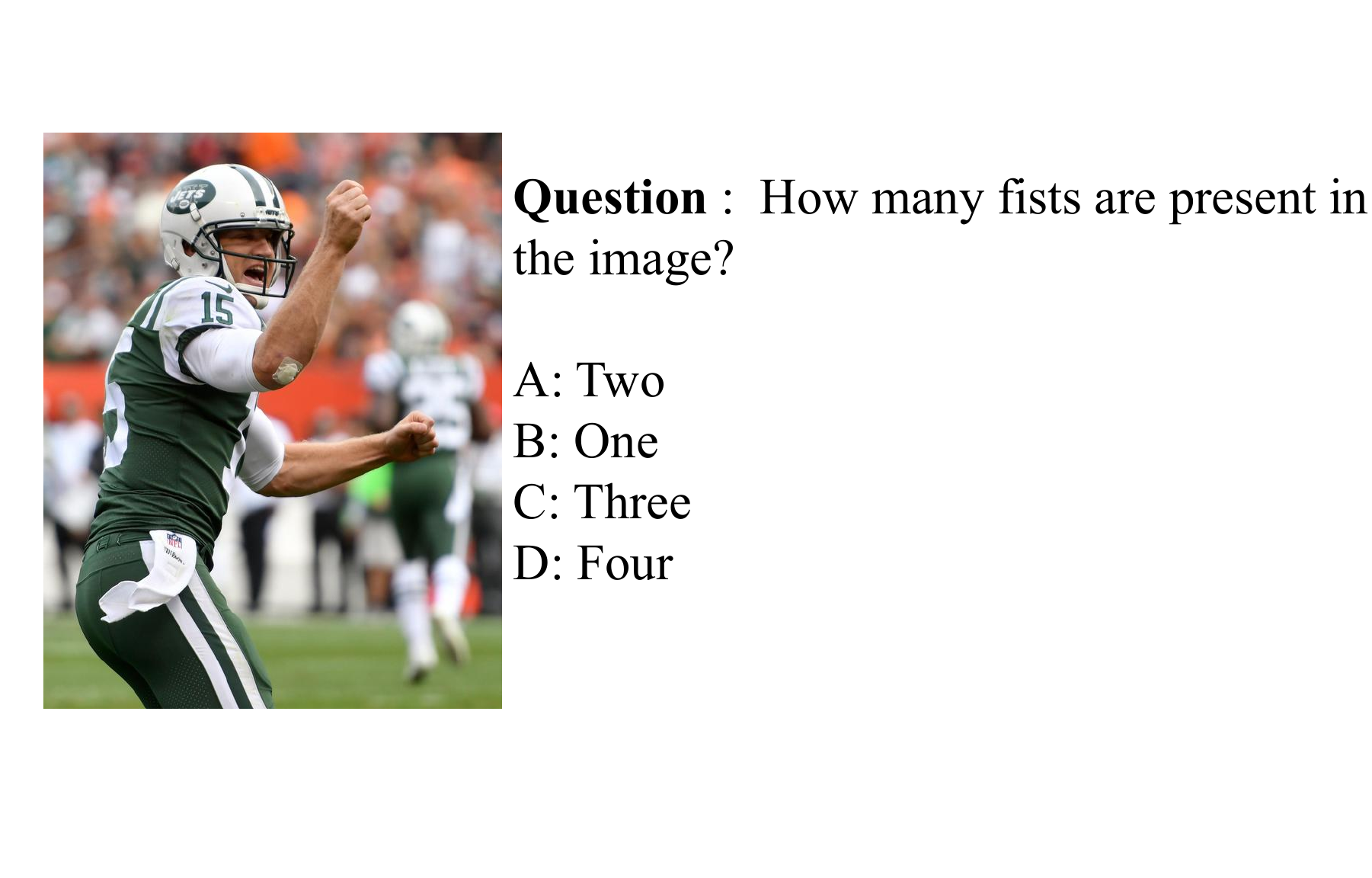}
        \caption{SEED-Bench (High quality)}
        \label{fig:high-synergy-3ds-vqa}
    \end{subfigure}
    \centering
    \begin{subfigure}[b]{0.32\textwidth}
        \centering
        \includegraphics[width=\textwidth]{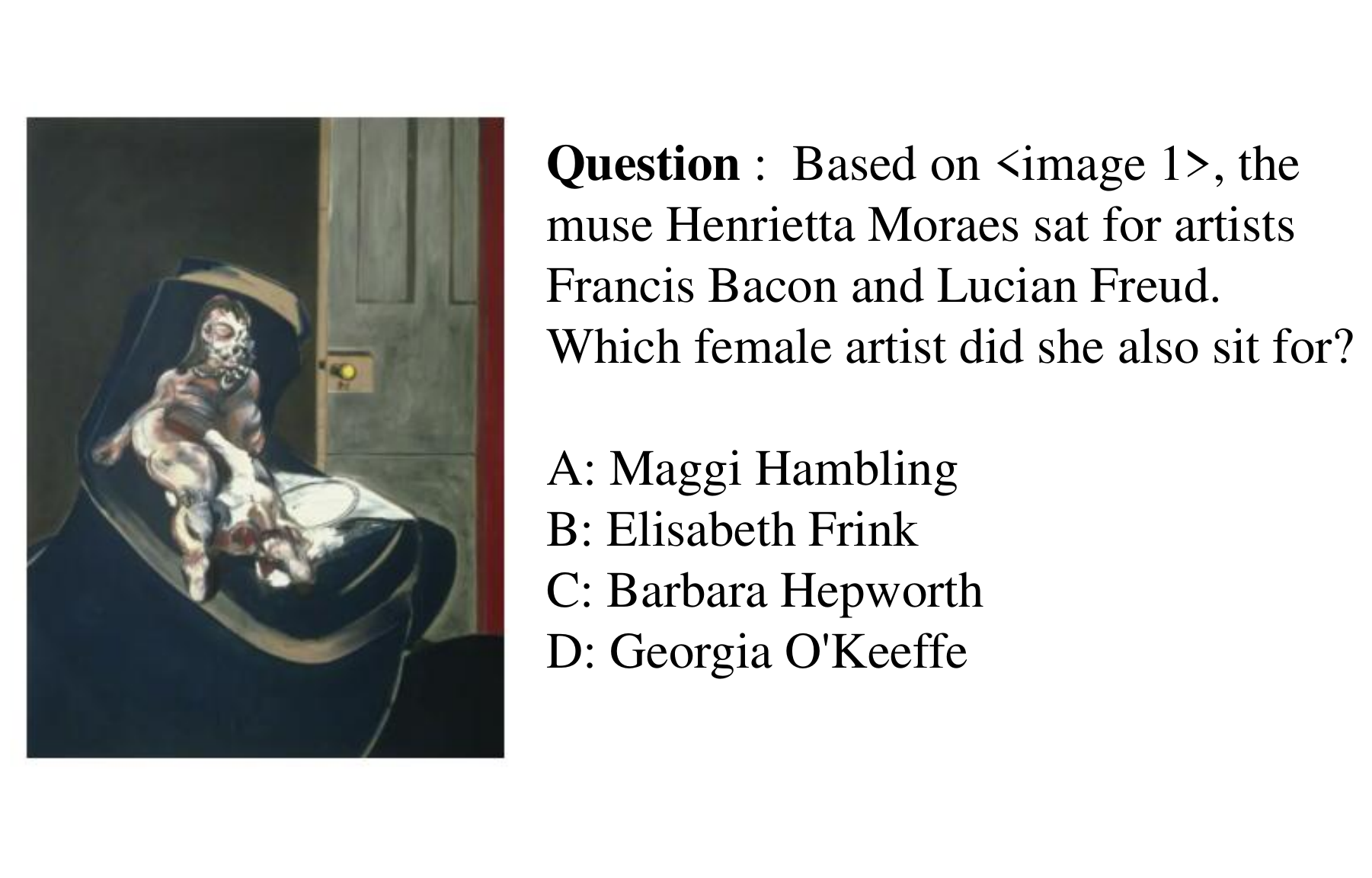}
        \caption{MMMU (Low quality)}
        \label{fig:low-synergy-3ds-mmmu}
    \end{subfigure}
    \begin{subfigure}[b]{0.32\textwidth}
        \centering
        \includegraphics[width=\textwidth]{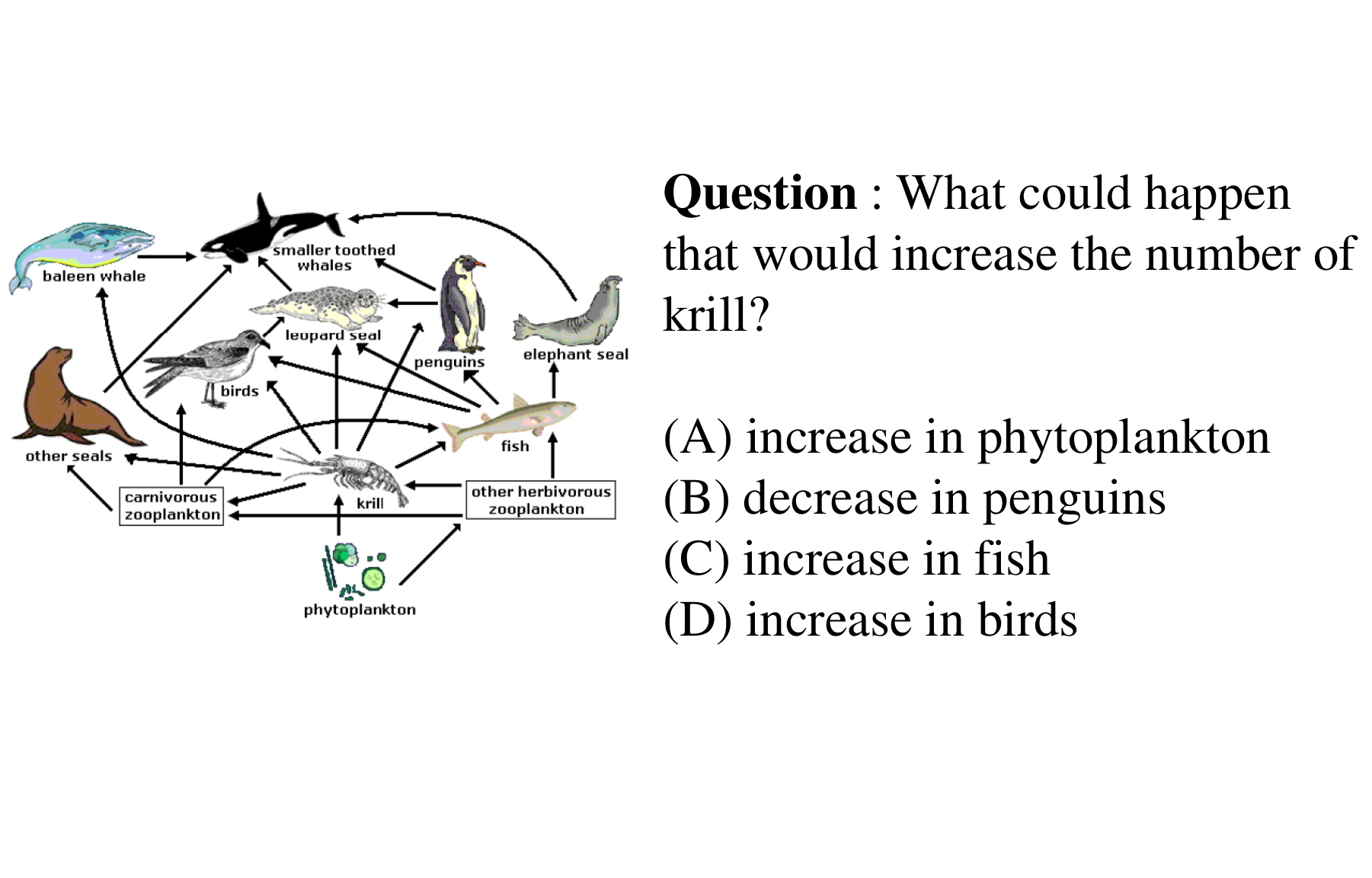}
        \caption{Mathvista (Low quality)}
        \label{fig:low-synergy-3ds-mathvista}
    \end{subfigure}
    \begin{subfigure}[b]{0.32\textwidth}
        \centering
        \includegraphics[width=\textwidth]{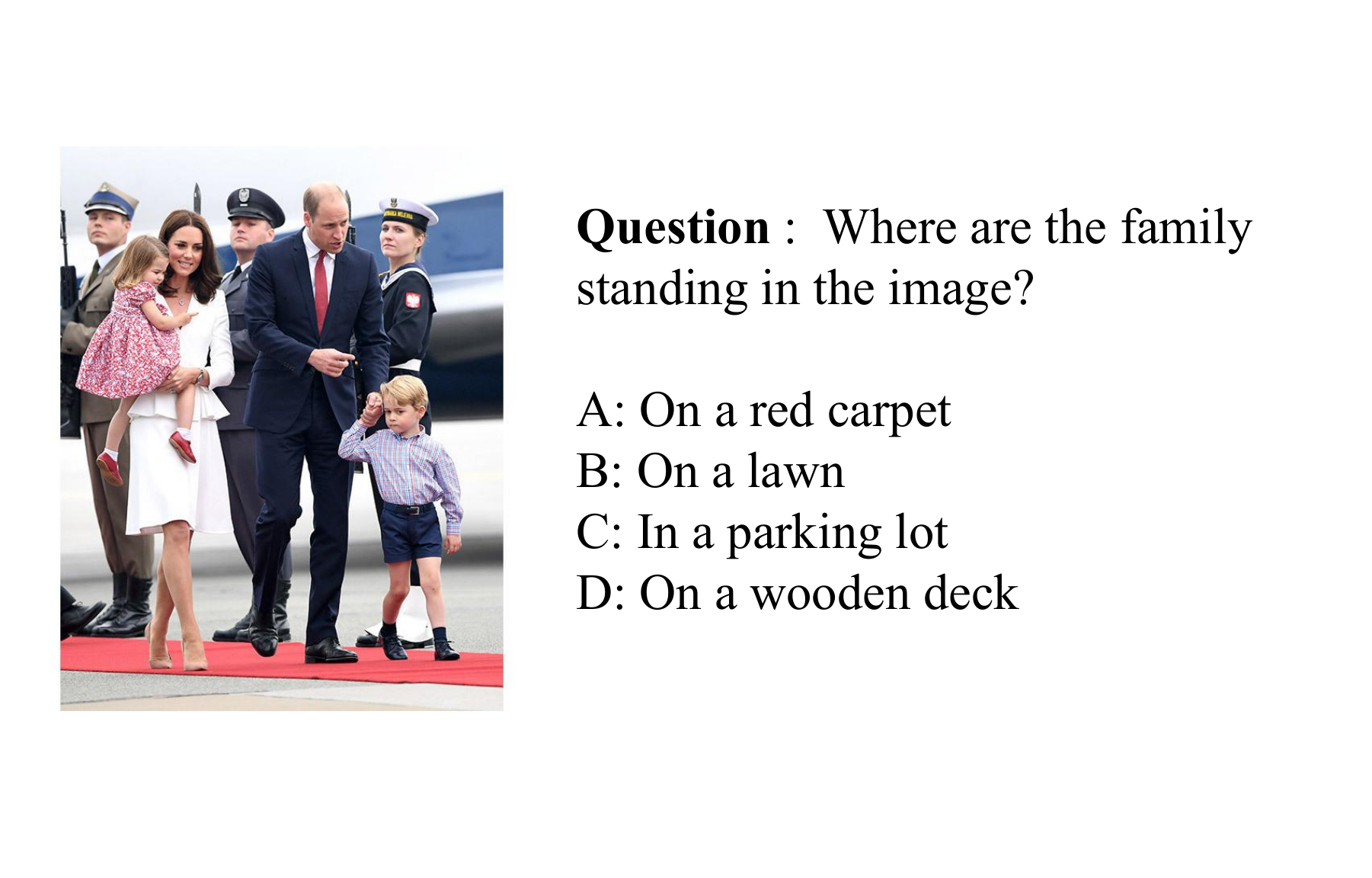}
        \caption{SEED-Bench (Low quality)}
        \label{fig:low-synergy-3ds-vqa}
    \end{subfigure}
    \caption{Questions with the highest or lowest cross-modal difficulty $\bsynergy$ detected by \mmmirt{}. 
    Questions with high cross-modal difficulty require both modalities to find the correct answer. However, those with low difficulty allow us to solve using only the image or text.
    }
    \label{fig:problem-synergy}
\end{figure}

We conduct extensive experiments with 24 VLMs across three benchmarks.
We construct semi-synthetic benchmarks by generating simple low-quality questions through the swapping of image or text from the original questions to introduce artificial shortcut or unsolvable questions.
We obtain the answers of VLMs and make datasets indicating successes and false.
We employ \mmmirt{}, \mmirt{}, and other methods including IRT to refine our semi-synthetic benchmarks.
First, we qualitatively observe that \mmmirt{} prioritizes truly cross‑modal items over shortcuts and preserves ranking fidelity even when 50\% of the items are replaced with artificially generated low‑quality questions.
Representative highly and lower cross-modal difficulty items identified by \mmmirt{} are shown in \Cref{fig:problem-synergy}.

Second, we conducted experiments to extract subsets of questions from the dataset as a high-quality problem-discovery task.
We quantitatively evaluate the degree of ranking reconstruction for VLMs obtained from a small number of subsets of varying sizes, as well as the proportion of simple low-quality questions included in these small subsets. The former enables high performance with fewer items.
The results show that our proposed framework nearly reconstructs the original ranking using only a 10\% subset across all datasets, while also reducing the proportion of low-quality questions to less than half that of existing methods.


Our contributions\footnote{Our code and data are available at \url{https://github.com/CyberAgentAILab/M3IRT}.}
 are threefold:
\begin{enumerate}
    \item We propose \mmmirt{}, which explicitly models modality‑specific (image‑only, text‑only) and cross‑modal components of both item difficulty and model ability for multimodal evaluation.
    \item We show that \mmmirt{} yields compact, high‑quality subsets that emphasize cross‑modal reasoning and maintain reliable model rankings at substantially reduced computational cost.
    \item Through experiments with 24 VLMs across three benchmarks, we demonstrate that \mmmirt{} is robust to large fractions of low‑quality items (up to 50\%) and provides interpretable characterizations of both benchmarks and models.
\end{enumerate}

\section{Related Work}
\label{sec:related}
Recent VLM evaluation has relied on large, static benchmarks such as MMMU~\citep{Yue_2024_CVPR_MMMU}, MathVista~\citep{lu2024mathvista},  SEED‑Bench~\citep{Li_2024_CVPR_seed}, EMMA~\citep{hao2025can} and CCHall~\citep{zhang-etal-2025-cchall}. 
These efforts shift the center of evaluation toward integration itself rather than isolated unimodal skills.
Static expansions such as MMBench~\citep{liu2024mmbench} broaden ability coverage but still exposed to low-quality question contamination and leakage. Several dynamic or live evaluation approaches have emerged such as VLB/FLEX~\citep{yang2025dynamic} proposes to automatically generate both image and text. MAC~\citep{jiang2025mac} and LiveXiv~\citep{
shabtay2025livexiv} automatically constructs VQA from current news and papers.
While valuable, these benchmarks still exposed to the risk of contaminating low-quality questions, such as shortcuts.

Existing methods for single-modal benchmarks can be categorized into Non-IRT-based and IRT-based approaches.  
First, Non-IRT-based approaches include question clustering that selects representative questions from clustering results, such as active testing with multi-stage sampling~\citep{huang2024activetestinglargelanguage}, tailored benchmark creation~\citep{yuan2025onesizefitsalltailoredbenchmarksefficient}, LLM predictability exploration~\citep{ye2023predictablelargelanguagemodel}, and anchor points~\citep{vivek-etal-2024-anchor}.  
Adaptive sampling dynamically selects questions based on current assessments of a model's performance, including SubLIME~\citep{xu2024dataefficientevaluationlarge}, Dele~\citep{saranathan2024dele}, and methods that model inter-example dependencies~\citep{li2024activeevaluationacquisitionefficient}.  
FlashEval~\citep{Zhao_2024_CVPR_FlashEval} was recently proposed, offering a novel evolutionary algorithm for text-to-image generation.  
However, they have not considered whether a question demand the cross-modal integration or not.

Item Response Theory (IRT)~\citep{Lord1980Applications}, originating in psychometrics, provides simultaneous modeling of subject (model) ability and item (question) parameters (e.g., difficulty, discrimination).  
The application of IRT has expanded to NLP~\citep{lalor-etal-2016-building-irt-nlp}, dialogue~\citep{hirai2023applying}, and recommendation systems~\citep{liu2023what}.  
In the LLM domain, IRT has been leveraged to reduce benchmark volumes; \ie, MetaBench~\citep{kipnis2025metabenchsparsebenchmark} distills a sparse benchmark from several benchmarks, and TinyBenchmarks~\citep{polo2024tinybenchmarksevaluatingllmsfewer} provides an efficient cluster-based sampling method.  
IRT has also been employed for adaptive sampling/testing of LLMs; for example, dynamic test adjustment based on model performance~\citep{zhuang2023fromstaticbenchmarkstoadaptivetesting}, CAT-based cognitive ability measurement~\citep{zhuang2025positionaievaluationlearn}, human chatbot evaluation, training of difficulty-calibrated question generators~\citep{jiang2025raisingbarinvestigatingvalues}, and automated model evaluation~\citep{autorageval2024}.
\section{Background}
\label{sec:irt}
Consider a collection of MLLMs, treated as subjects and indexed by $M = \{1,\dots,m\}$, and a multimodal benchmark with questions treated as items and indexed by $N = \{1,\dots,n\}$.
For each subject–question pair $(i,j)$, let $\rij \in \{0,1\}$ indicate whether subject $i$ answers question 
$j$ correctly ($\rij=1)$ or not ($\rij=0$). We denote the resulting response matrix by $R=\{r_{i,j}\}_{(i,j)\in M\times N}$.
Our objective is to assess the cross‑modal abilities of the MLLMs and the difficulty of the questions, and to identify a compact subset $\hat{N} \subset N$ consisting of items that demand strong cross‑modal reasoning.

Item Response Theory (IRT) is a family of latent variable models that jointly infer subject ability and item characteristics from observed response data~\citep{doi:10.1177/0013164498058003001}. 
Given only the pattern of correct or incorrect responses, IRT estimates ability and difficulty parameters and predicts the probability that a subject will answer a given item correctly. 
We use the two-parameter logistic (2PL) model, which can be viewed as a logistic regression with item-specific slope and threshold:
\begin{align}
\label{equation:irt}
\Pr(r_{i,j}=1 \mid \theta_i,a_j,b_j) 
= \sigma\!\big(a_j(\theta_i - b_j)\big),
\end{align}
where $\sigma(x) = 1/(1 + \exp(-x))$ is the sigmoid function.
For each subject \(i\), we define an ability parameter \(\theta_i \in \mathbb{R}\); higher values indicate a greater propensity to answer difficult items correctly. 
For each item \(j\), we define a discrimination parameter \(a_j > 0\) and a difficulty parameter \(b_j \in \mathbb{R}\). 
Larger \(a_j\) means the probability of a correct response is more sensitive to changes in ability, whereas smaller \(a_j\) implies weaker sensitivity. 
As the difficulty \(b_j\) increases, greater ability is required to achieve a high probability of a correct response.
IRT has been applied to CAT~\citep{9c2d3927-7193-3702-9179-0a57b6a5e7e0} to select test questions from an item pool to estimate a subject ability.
Namely, we randomly initialize a student ability, select a question with the maximum Fisher information for a current ability, get an answer, and update a subject ability. We repeat this procedure.

Multi-dimensional IRT (MIRT) is a method that extends IRT to consider the relationship between models and questions in a more complex manner~\citep{reckase2009multidimensional}.
This method supposes a $d$-dimensional latent parameter space.
The ability vector for subject $i$ is $\bm{\theta}_i \in \mathbb{R}^{d}$, and the difficulty and discriminative vectors for question $j$ as $\bm{a}_j, \bm{b}_j \ \in \mathbb{R}^{d}$.
MIRT parametrizes the probability for providing a correct answer for a pair of $(i,j)$ and finds maximum likelihood estimator:
\begin{align}\label{eq:mirt}
  \hat{P}(r_{i,j}=1) = \sigma\bigl(\bm{a}_j^{\top}\bm{\theta}_i - \bm{b}_j\bigr),~
  \hat{P}(r_{i,j}=0) = 1 - \hat{P}(r_{i,j}=1).
\end{align}


\section{Proposed Method}
\label{sec:proposed}
\begin{figure}[t!]
    \centering
    \includegraphics[width=0.85\linewidth]{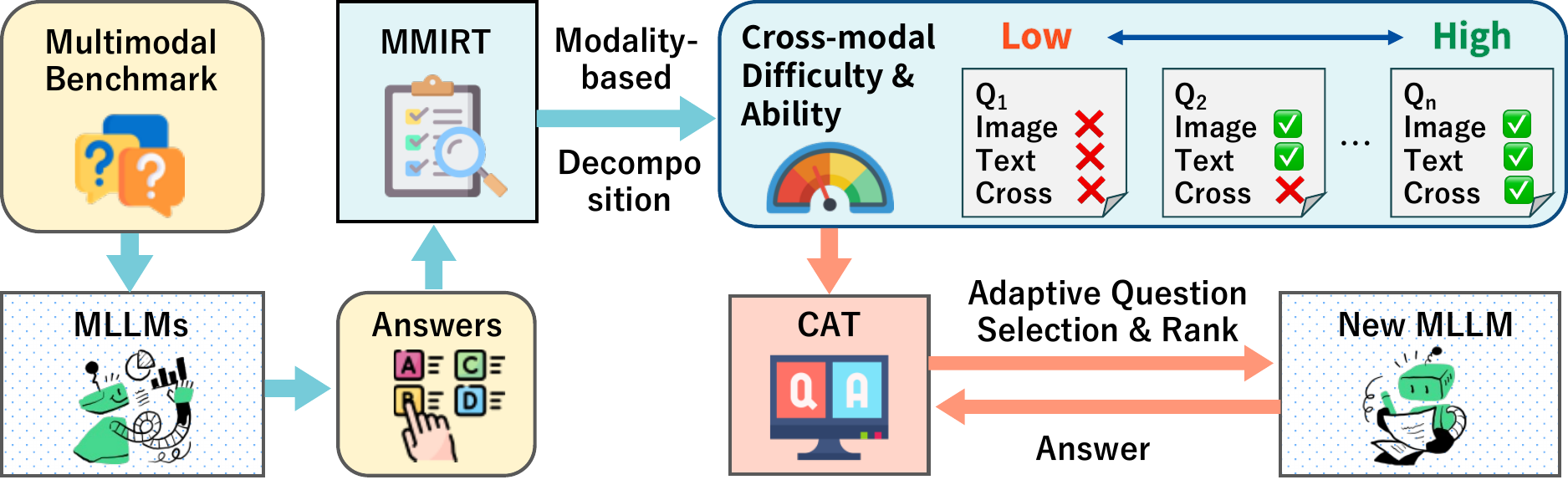}
    \caption{\mmirt{} investigates the modality-specific and cross-modal difficulties of questions that enables to contract a tailored, compact, and high-quality subset for evaluating a new MLLM.}
    \label{fig:teaser}
\end{figure}

To assess modality‑specific and cross‑modal properties of MLLMs and multimodal benchmarks, we introduce the decomposition of the standard IRT parameters into latent components.
Building on this decomposition, we introduce MultiModal Item Response Theory (\mmirt{}) and Multidimensional MultiModal Item Response Theory (\mmmirt{}) as extensions of classical IRT and MIRT.
We also develop a procedure for selecting a compact subset of benchmark items tailored to these models.
\Cref{fig:teaser} illustrates the overall framework.
Although the method is applicable to arbitrary modalities (e.g., action, audio), this paper primarily focuses on vision and language.

\subsection{Modality-based Decomposition of IRT Parameters}
We assume that an MLLM has modality‑specific abilities as well as an ability to integrate information across modalities. Likewise, each multimodal question exhibits modality‑specific and cross‑modal characteristics that can determine whether a subject can provide the correct answer.

In the vision–language setting, we define binary indicators $\simage, \stext \in \{0,1\}$ to represent the modalities present in a question:
$\simage=1$ if an image is provided and $\stext=1$ if text is provided; otherwise, the indicator is $0$.
Let $s = (\simage,\stext) \in S = \{(0,0), (0,1), (1,0), (1,1)\}$ denote a format of representing a question.
When $(\simage,\stext) = (0,0)$, the stimulus are withheld and the subject answers using only a guess from introductions or the multiple‑choice options. 

We assume each subject has a base reasoning ability that, depending on the input format $s$, combines with image‑specific, text‑specific, and cross‑modal integration abilities. For subject $i$, denote the base, image, text, and cross‑modal abilities by $\tbase, \timage, \ttext, \tsynergy \in [0, q]$, respectively, where $q \geq 0$ is a shared upper bound that balances their scales. 
Given a question $j$ and its modality indicator $s$, we define the ability of a subject $j$ as follows:
\begin{align}\label{eq:ability}
 \tfull = \tbase + \simage \timage + \stext \ttext + \simage \stext \tsynergy.
\end{align}
The second and third terms contribute when an image or text is present, respectively; the fourth term contributes only when both are present. This construction naturally extends to additional modalities.

We view answering as exploiting hints provided by the item. For item $j$, let $\bbase$, $\bimage$, $\btext$, $\bsynergy \in [0, q]$ be the base, image, text, and cross-modal difficulties, respectively, using the same upper bound $q\geq 0$. 
We define the difficulty $\bfull$ of question $j$ given the indicator $s$ as
\begin{align}\label{eq:difficulty}
 \bfull = \bbase - \simage \bimage - \stext \btext - \simage \stext \bsynergy.
\end{align}

Similarly, let $\abase \in [0, q]$ be the base discrimination, let $\aimage, \atext, \asynergy \in [0, q]$ capture the contributions from image, text, and cross‑modal integration. The discrimination becomes
\begin{align}\label{eq:discrimination}
 \afull = \abase + \simage \aimage + \stext \atext + \simage \stext \asynergy.
\end{align}

In a general setting, we define indicators to represent the all modalities in a benchmark, and extend parameters applicable to represent combinations of modalities.

\subsection{Multimodal Item Response Theory (\mmirt{})}
To capture cross‑modal behavior, we control which modalities are provided, thus each subject answers each item under the four input formats corresponding to all $s \in S$.
For each subject–question-format combination $(i,j,s)$, let $\rijs \in \{0,1\}$ indicate whether subject $i$ answers question $j$ given the format indicator $s$ correctly ($\rijs=1)$ or not ($\rijs=0$). 
We denote full response set as the resulting response tensor by $R'=\{\rijs\}_{(i,j,j)\in M\times N \times S}$.

\mmirt{} extends the logistic IRT model in \Eqref{equation:irt}. 
Given discrimination $\afull$, difficulty $\bfull$, and ability $\tfull$, we define $\zijs = \afull(\tfull-\bfull)$ and introduce \mmirt{} as follows:
\begin{align}\label{eq:probability_mmirt}
\hat{P}(\rijs=1) = \sigma\bigl(\zijs \bigr)
\quad \text{and} \quad
    \hat{P}(\rijs=0) = 1 - \hat{P}(\rijs=1).
\end{align}
This parameterization captures the modality‑aware behavior of subject $i$ on item $j$.

\subsection{Multimodal Multi-dimensional Item Response Theory (\mmmirt{})}
\mmmirt{} extends the logistic MIRT model in \Eqref{eq:mirt} with the modality‑based decomposition. 
We modify the decomposed components into vectors. For subject $i$, define the ability vector
$\bm{\theta}_i = [\tbase, \timage, \ttext, \tsynergy]^\top$.
For item $j$, define the discrimination and difficulty vectors
$\bm{a}_j = [\abase, \aimage, \atext, \asynergy]^\top$,~
$\bm{b}_j = [\bbase, \bimage, \btext, \bsynergy]^\top$.
For convenience, we introduce a format indicator vector
$\bm{s} = [1, -\simage, -\stext, -\simage \stext]^\top$,
where the negative signs align with the subtractive role of the modality terms in \Eqref{eq:difficulty} and with the decomposition in \Eqref{eq:discrimination}.
From these vectors, we define $\zijs' = \bm{a}_j^{\top}  \mathrm{diag}(\bm{s})  \bm{\theta}_i - \bm{s}^{\top} \bm{b}_j$.
We propose \mmmirt{} as follows:
\begin{align}\label{eq:probability_mmmirt}
\hat{P}(\rijs=1) = \sigma\bigl(\zijs' \bigr)
\quad \text{and} \quad
\hat{P}(\rijs=0) = 1 - \hat{P}(\rijs=1).
\end{align}
Here, $\mathrm{diag}(\bm s)$ is the diagonal matrix whose diagonal elements are $\bm s$.
The probabilistic model~\eqref{eq:probability_mmirt} is a variant of multi-dimensional IRT with the parametrization $\zijs$.
This parametrization takes in the modality-aware nature of subject $i$ when answering multimodal question $j$.


\subsection{Learning \mmmirt{} using Stochastic Gradient Descent}\label{subsec:learning-mmirt}
Instead of the EM algorithm commonly used in IRT, we estimate \mmmirt{} parameters with stochastic gradient descent (SGD).
Let a training dataset as $R'' \subset R'$.
Given $R''$ and the Bernoulli model in \Eqref{eq:probability_mmirt}, the negative log‑likelihood is the negative log likelihood of is
\begin{align}
    \mathcal{L}(\Theta) = - \sum_{(i,j,s) \in R''} \left( \rijs \log \hat{P}(\rijs=1) + 
    (1-\rijs) \log \hat{P}(\rijs=0)\right),
\end{align}
where the parameters set is $\Theta = \{\{\bm{a}_j\}_{j\in{N}}, \{\bm{b}_j\}_{j\in{N}}, \{\bm{\theta}_i\}_{i\in{M}}\}$.
We minimize $\mathcal{L}(\Theta)$ busing mini-bach SGD, \(\hat{\Theta} = \mathop{\rm arg min}_{\Theta} \mathcal{L}(\Theta)\).
We can estimate \mmirt{} in a similar manner.
Note that our approach does not require a dense response matrix: \mmirt{} and \mmmirt{} can be learned from partially observed data like a tensor completion, reducing the cost of evaluating MLLMs and benchmarks.

\subsection{Computer Adaptive Test with \mmirt{} and \mmmirt{}}\label{sec:cat}
We integrate \mmirt{} and \mmmirt{} with classical Computerized Adaptive Testing (CAT)~\citep{9c2d3927-7193-3702-9179-0a57b6a5e7e0} to adaptively select an informative subset of items $\hat{N} \subseteq N$, guided by Fisher information.
For \mmirt{} model, the Fisher information of item $j$ for subject $i$ under format $s$ is
\textbf{\begin{align}
    I_{i,j} = \hat{P}(\rijs=1)\hat{P}(\rijs=0) (\afull)^2,
\end{align}}
where $\hat{P}(\rijs=1)$ is given by \Eqref{eq:probability_mmirt}.
For the multidimensional \mmmirt{} model, the Fisher information matrix for item $j$ at ability $\theta$ is
\begin{align}
    \bm{I}_{i,j} = \hat{P}(\rijs=1)\hat{P}(\rijs=0) ({\rm diag}(\bm{s})\bm{a}_j) ({\rm diag}(\bm{s}){\bm{a}_j})^{\top}.
\end{align}

We adopt the D‑optimality criterion~\citep{mirt-cat-mulder} to minimize estimation uncertainty by maximizing the determinant of the cumulative information. Let 
$U_i \subseteq N$ be the set of items not yet answered by subject $i$.
At stage $t$, given the cumulative information matrix $\bm{I}_{i}^{(t-1)}$, we select the next item and update:
\begin{align}
j^* = \underset{j \in U_i}{\operatorname{argmax}} \det\left(\bm{I}_i^{(t-1)} + \bm{I}_{ij}\right),~ \bm{I}_i^{(t)} = \bm{I}_i^{(t-1)} + \bm{I}_{ij^*}.
\end{align}
Iterating this rule yields a subset that is maximally informative for estimating the subject’s ability.

\section{Experiment}
\label{sec:exp}
\subsection{Datasets and Baselines}
\label{subsubsec:dataset}
We employed three benchmarks for VLMs in this experiment.
\textsc{\textbf{MMMU}}~\citep{Yue_2024_CVPR_MMMU} is designed to evaluate the reasoning capabilities of VLM through undergraduate-level questions in diverse disciplines such as art and design, business, and science. We used $900$ questions in the validation set.
\textsc{\textbf{MathVista}}~\citep{lu2024mathvista} evaluates mathematical reasoning capabilities through questions involving visual context including puzzle figures and graphs. We used $1000$ questions of the test-min set.
\textsc{\textbf{SEED-Bench}}~\citep{Li_2024_CVPR_seed} is a large-scale benchmark designed to comprehensively evaluate the multimodal abilities. 
We used $1000$ questions from \textbf{L1} and \textbf{L2} sets.

To simulate the presence of questionable samples in real-world datasets, we constructed a synthetically contaminated benchmark.
We made semi-synthetic benchmarks by generating simple low-quality questions through the swapping of image or text from the original questions. This process introduces artificial shortcut or unsolvable questions.
We compile a benchmark contaminated with 50\% low-quality questions. 
We provide a detailed description of our data generation process in Appendix~\ref{appendix:low-question}.
To create more realistic low-quality questions, methods such as modifying text and options using LLM or  adding noise to images could be considered. 
Since such methods make the experiment overly complex, we excluded them.
Note that our method learns ability and problem characteristics from whether VLMs answer questions correctly, even if there are different types of low-quality questions, the estimation results are unlikely to change.

We collected responses from 24 VLMs, including the GPT-4.1 series, Gemini-2.0 series, and Claude-3.7 series, as well as open-source models such as Qwen-2.5-vl~\citep{Qwen2.5-VL}, Llama-3.2~\citep{meta2024llama}, and Pixtral~\citep{agrawal2024pixtral}.
On \textsc{SEED-Bench}, since Claude-sonnet-3 became unavailable at the start of the experiments on SEED-Bench, the experiments on SEED-Bench were conducted with 23 models other than Claude-sonnet-3.

We use four baseline methods in our experiments. \textbf{Random} selects subset questions at random. \textbf{IRT} uses a Fisher information-based subset selection estimated by IRT~\citep{reckase2009multidimensional}.
\textbf{MIRT} uses a Fisher information-matrix-based subset selection estimated by MIRT~\citep{reckase2009multidimensional}.
\textbf{TinyBenchmarks}~\citep{polo2024tinybenchmarksevaluatingllmsfewer} is an IRT-based problem selection method for benchmark refinement in LLM.
\textbf{FlashEval}~\citep{Zhao_2024_CVPR_FlashEval} is a SOTA to select prompts for image generation. We extended FlashEval to deal with VLM benchmarks by regarding questions as prompts.

We implemented our proposed method with PyTorch~\citep{paszke2019pytorch}, and used Adam optimizer~\citep{kingma2014adam} whose learning rate was $0.01$.
We used a grid search to select hyperparameter $q$ from ${2,4,8,16}$.
We selected the optimal hyperparamters based on the highest AUC in predicting the correctness of the VLMs' responses on the validation dataset.
We provide the detailed explanation of the experimental setting in Appendix~\ref{appendix:experiment}.

\subsection{Multimodal Difficulty and Ability Decomposition}
\begin{figure*}
    \centering
    \begin{subfigure}[b]{0.48\textwidth}
        \centering
        \includegraphics[width=\textwidth]{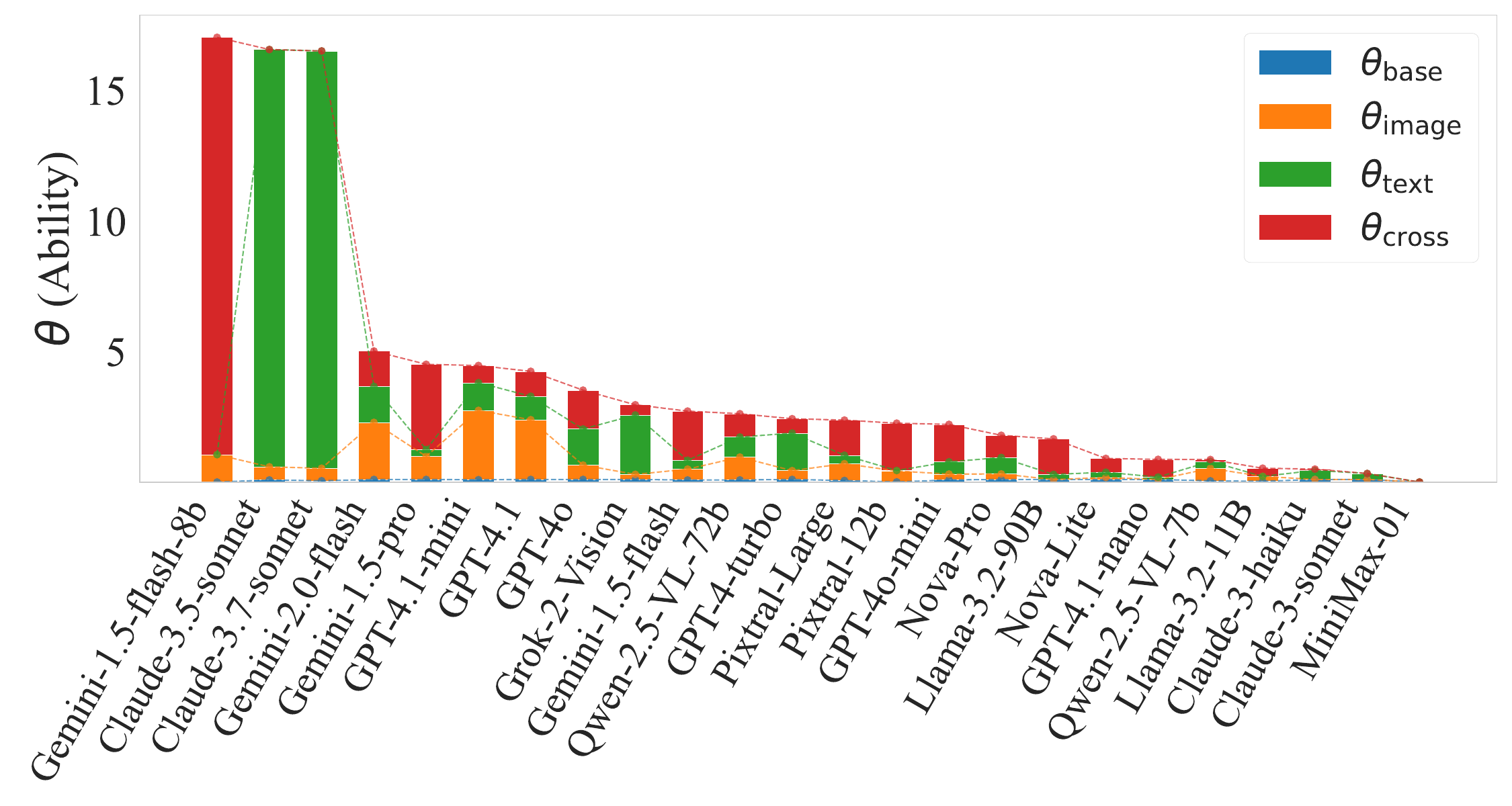}
        \caption{\textsc{MMMU}}
        \label{fig:theta-inappropriate-mmmu}
    \end{subfigure}
    \begin{subfigure}[b]{0.48\textwidth}
        \centering
        \includegraphics[width=\textwidth]{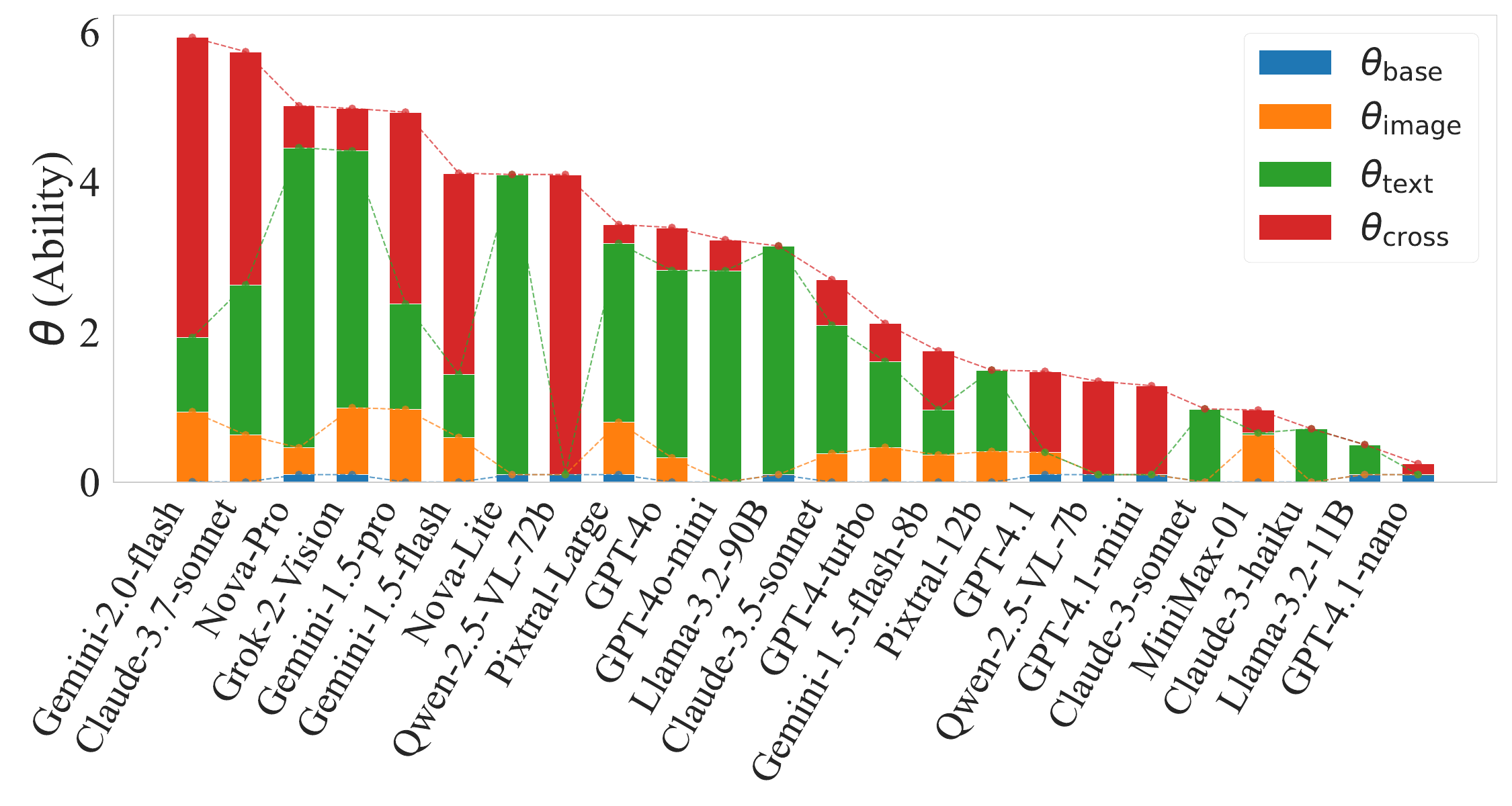}
        \caption{\textsc{MathVista}}
        \label{fig:theta-inappropriate-mathvista}
    \end{subfigure}
    \caption{Distributions of $\theta$ estimated by \mmmirt{} sorted in descending order.}
    \label{fig:theta-inappropriate}
\end{figure*}

\mmmirt{} estimates the extent to which a question requires cross-modal reasoning, represented by difficulty $\bsynergy$. 
This facilitates the identification of questions that truly benefit model's cross-modal capability assessment. 
\Cref{fig:problem-synergy} shows examples of questions with high and low $\bsynergy$.
The questions with low $\bsynergy$ are judged that they can be solved only with images or text.
For example, the bottom one in \textsc{MMMU} can be answered based on knowledge of artists without looking into the image.
On the other hand, the questions with high $\bsynergy$ cannot be solved if either the image or the text is missing.
For example, the one in \textsc{MathVista} requires reading the numerical values in the table that cannot be confirmed only by the question text.
Similarly, if only images are provided, it is not clear what is being asked about in the table. We provide more examples in \cref{appendix:b}.

\mmmirt{} also estimates the extent to which the reasoning ability for each modality contributes to the VLM performance.
\Cref{fig:theta-inappropriate} shows the decomposed reasoning abilities of VLMs.
The top-performing model on \textsc{MMMU} exhibits high ($\tsynergy$), suggesting strong cross-modal reasoning capabilities. On the other hand, the second and third best-performing models demonstrate high textual reasoning ability ($\ttext$) but limited cross-modal reasoning capability. This analysis suggests that these latter VLMs rely heavily on text understanding when solving the MMMU benchmark, rather than effectively integrating visual information.
In \textsc{MathVista}, most VLMs have high $\ttext$. This may reflect \textsc{MathVista}'s emphasis on text understanding. Most VLMs also exhibit moderate $\timage$, suggesting that they also leverage the visual ability to process diagrams and graphs. 
The result for \textsc{SEED-Bench} is shown Appendix~\ref{appendix:b}.

\subsection{Benchmark Refinement}
\label{subsec:subset}
\begin{figure*}
    \centering
    \begin{subfigure}[b]{0.32\textwidth}
        \centering
        \includegraphics[width=\textwidth]{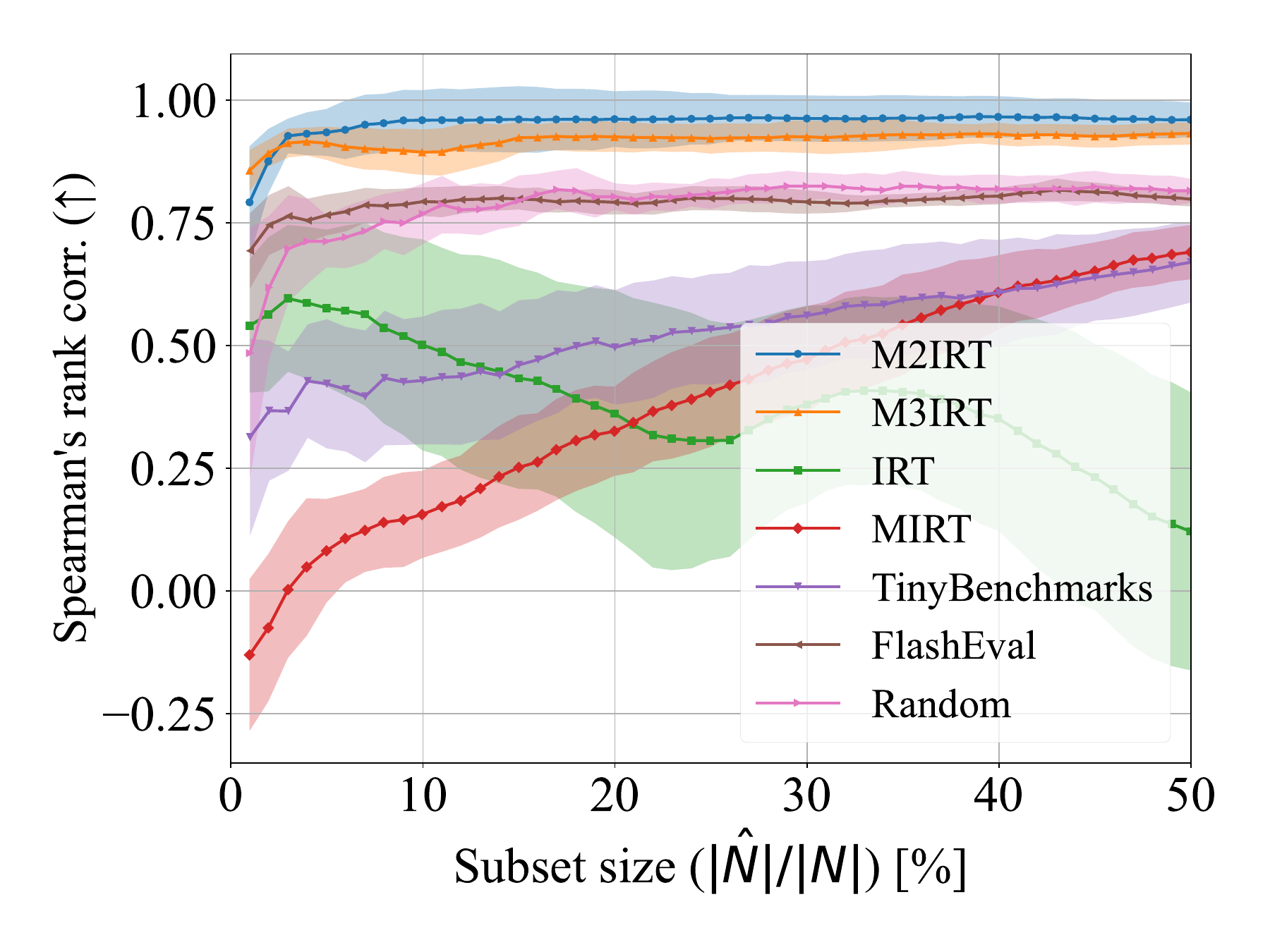}
        \caption{\textsc{MMMU}}
        \label{fig:subset-rank-comp-mmmu}
    \end{subfigure}
    \begin{subfigure}[b]{0.32\textwidth}
        \centering
        \includegraphics[width=\textwidth]{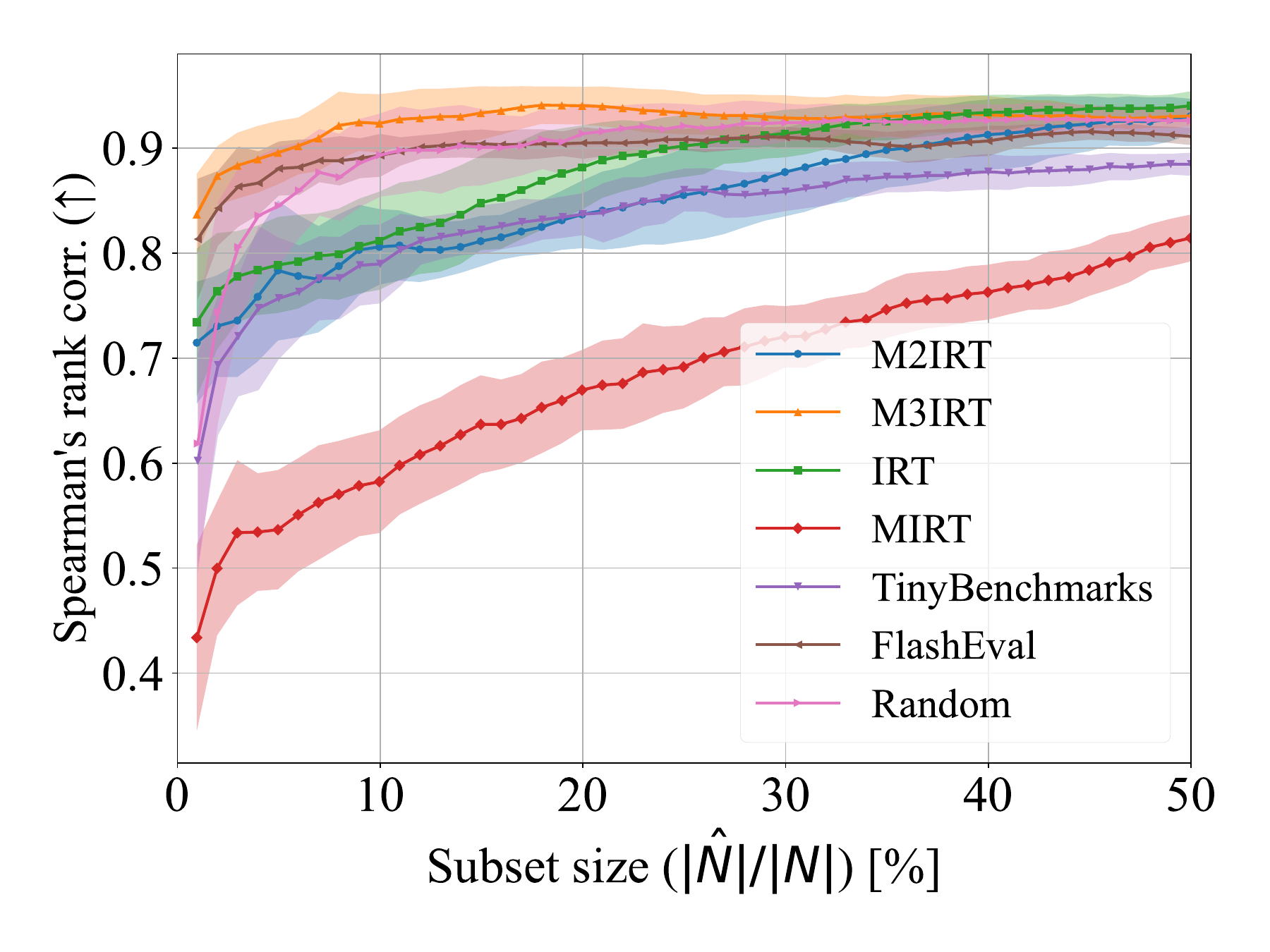}
        \caption{\textsc{MathVista}}
        \label{fig:subset-rank-comp-mathvista}
    \end{subfigure}
    \begin{subfigure}[b]{0.32\textwidth}
        \centering
        \includegraphics[width=\textwidth]{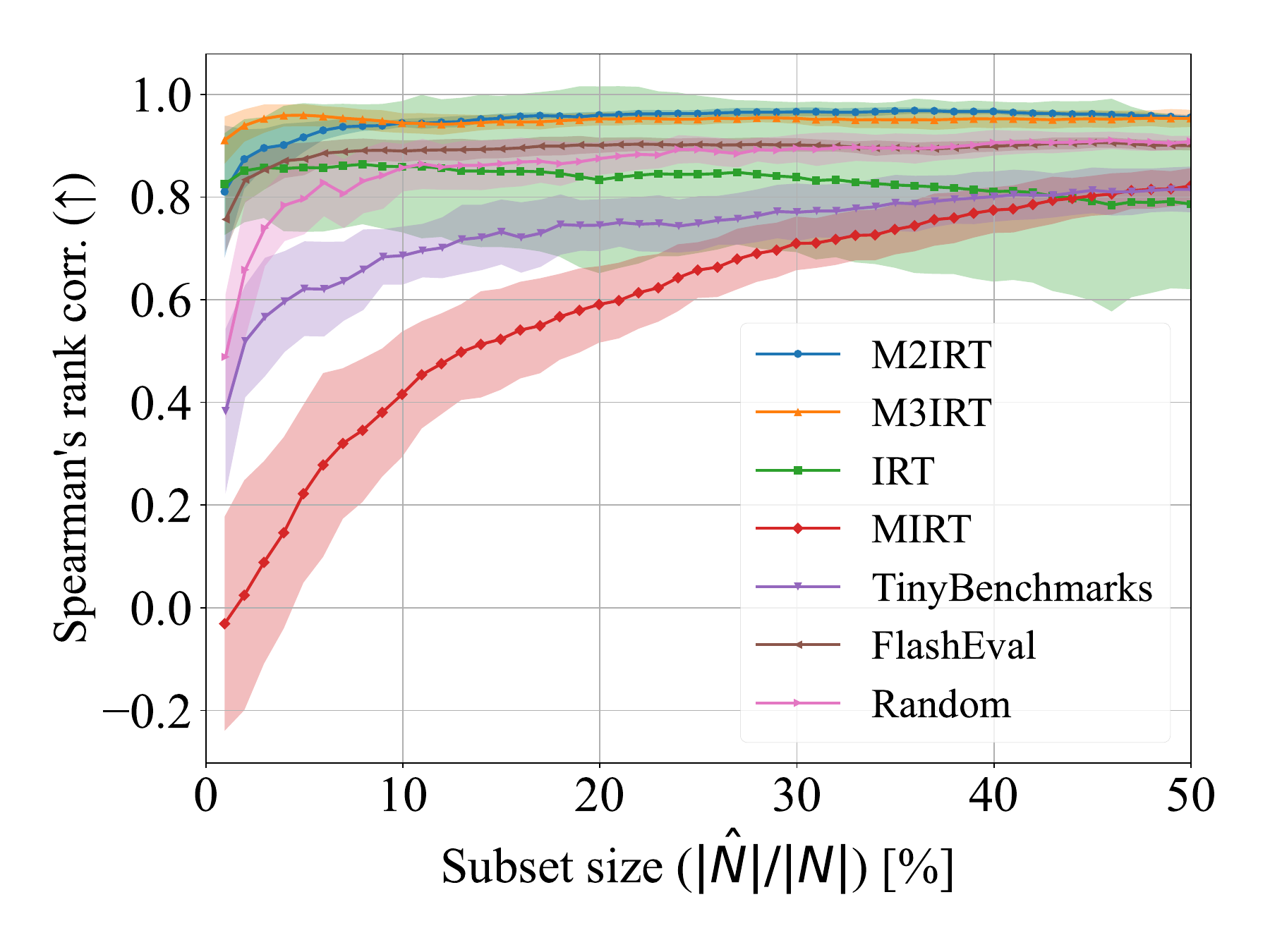}
        \caption{\textsc{SEED-Bench}}
        \label{fig:subset-rank-comp-seed}
    \end{subfigure}
    \caption{The average and standard deviation of Spearman's rank correlations between model rankings on the original benchmark and those estimated on extracted question subsets with different sizes.}
    \label{fig:subset-rank-comp}
\end{figure*}
\begin{figure*}
    \centering
    \begin{subfigure}[b]{0.32\textwidth}
        \centering
        \includegraphics[width=\textwidth]{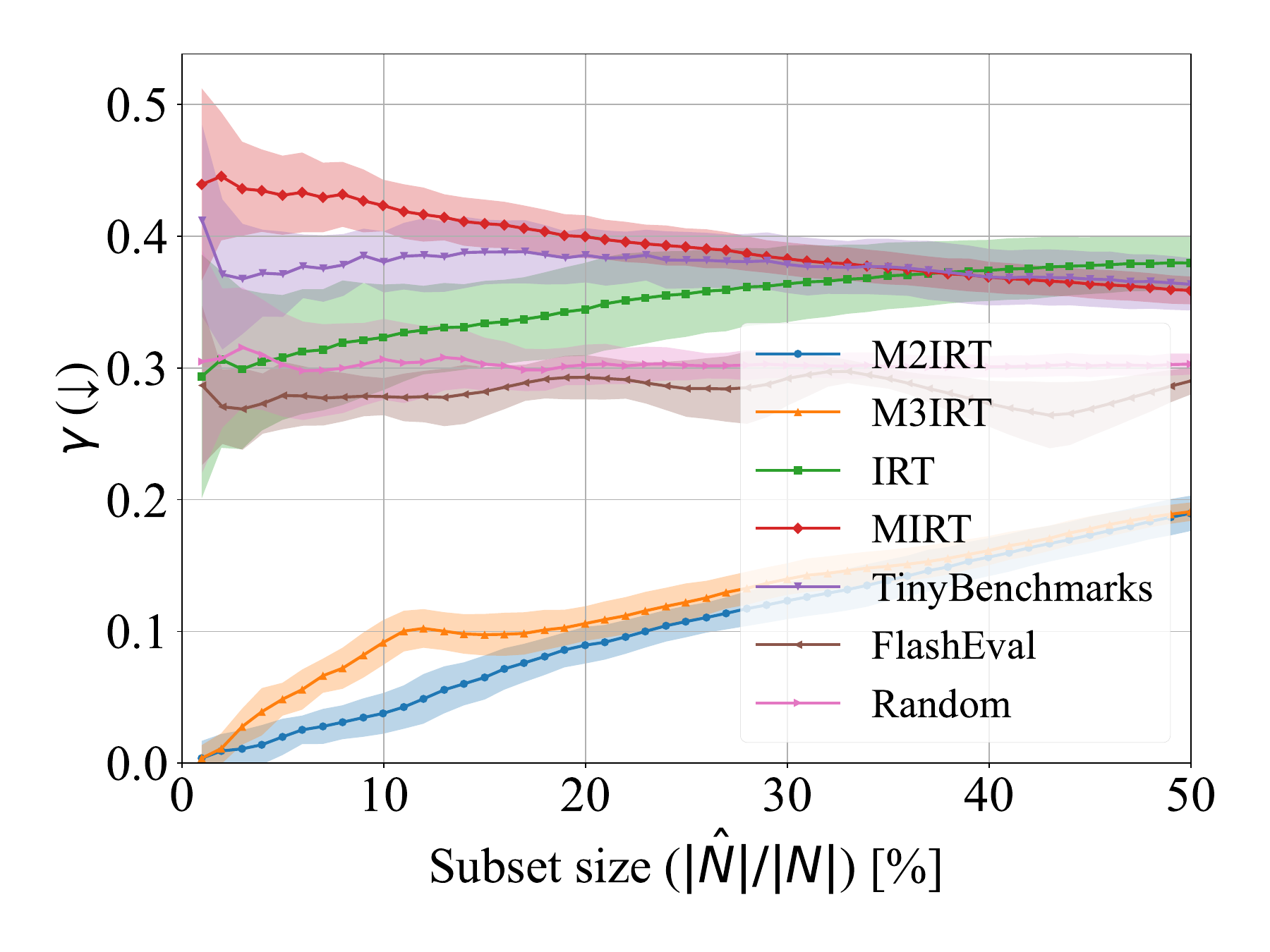}
        \caption{\textsc{MMMU}}
        \label{fig:subset-contami-tate-mmmu}
    \end{subfigure}
    \begin{subfigure}[b]{0.32\textwidth}
        \centering
        \includegraphics[width=\textwidth]{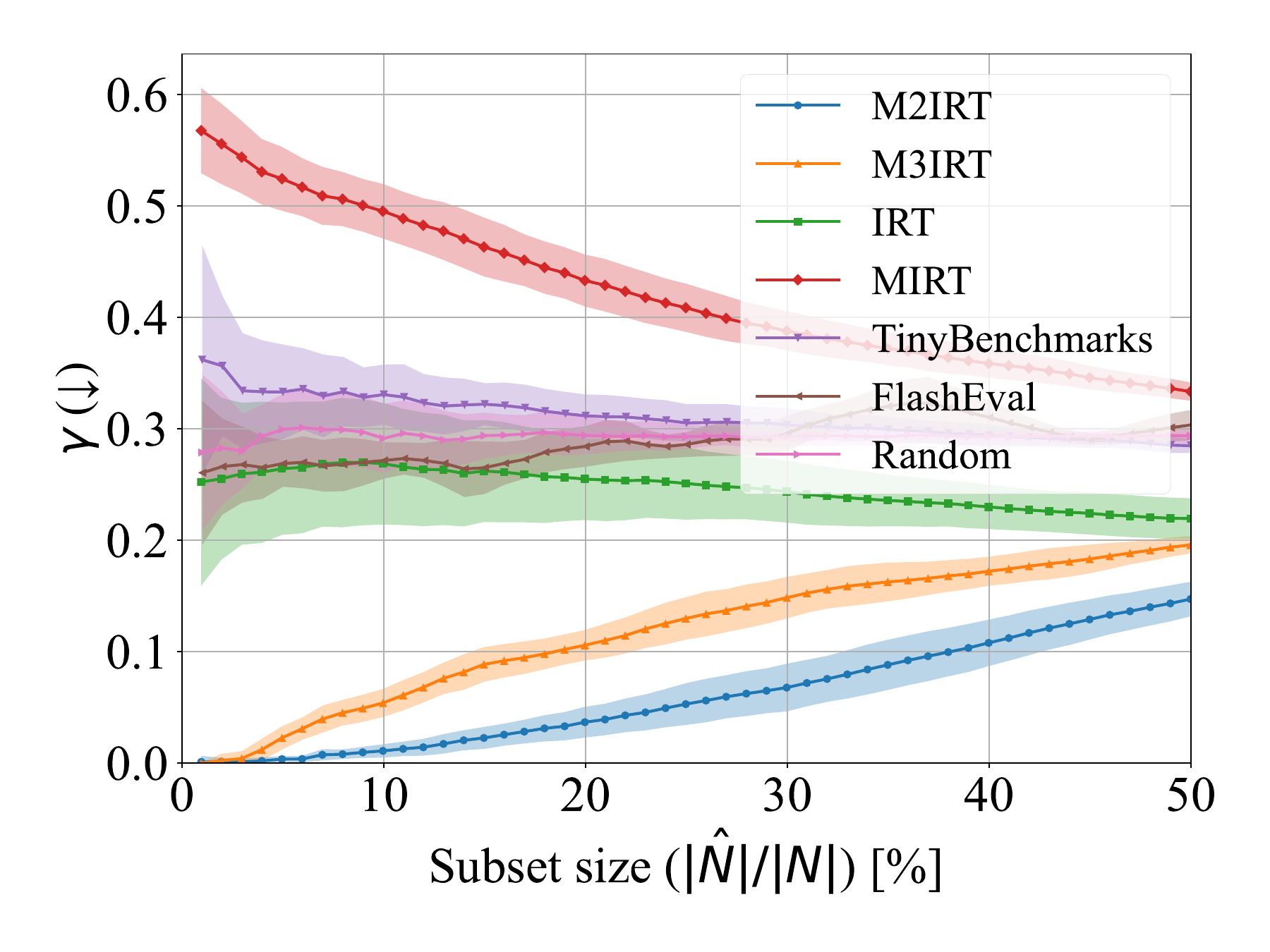}
        \caption{\textsc{MathVista}}
        \label{fig:subset-contami-tate-mathvista}
    \end{subfigure}
    \begin{subfigure}[b]{0.32\textwidth}
        \centering
        \includegraphics[width=\textwidth]{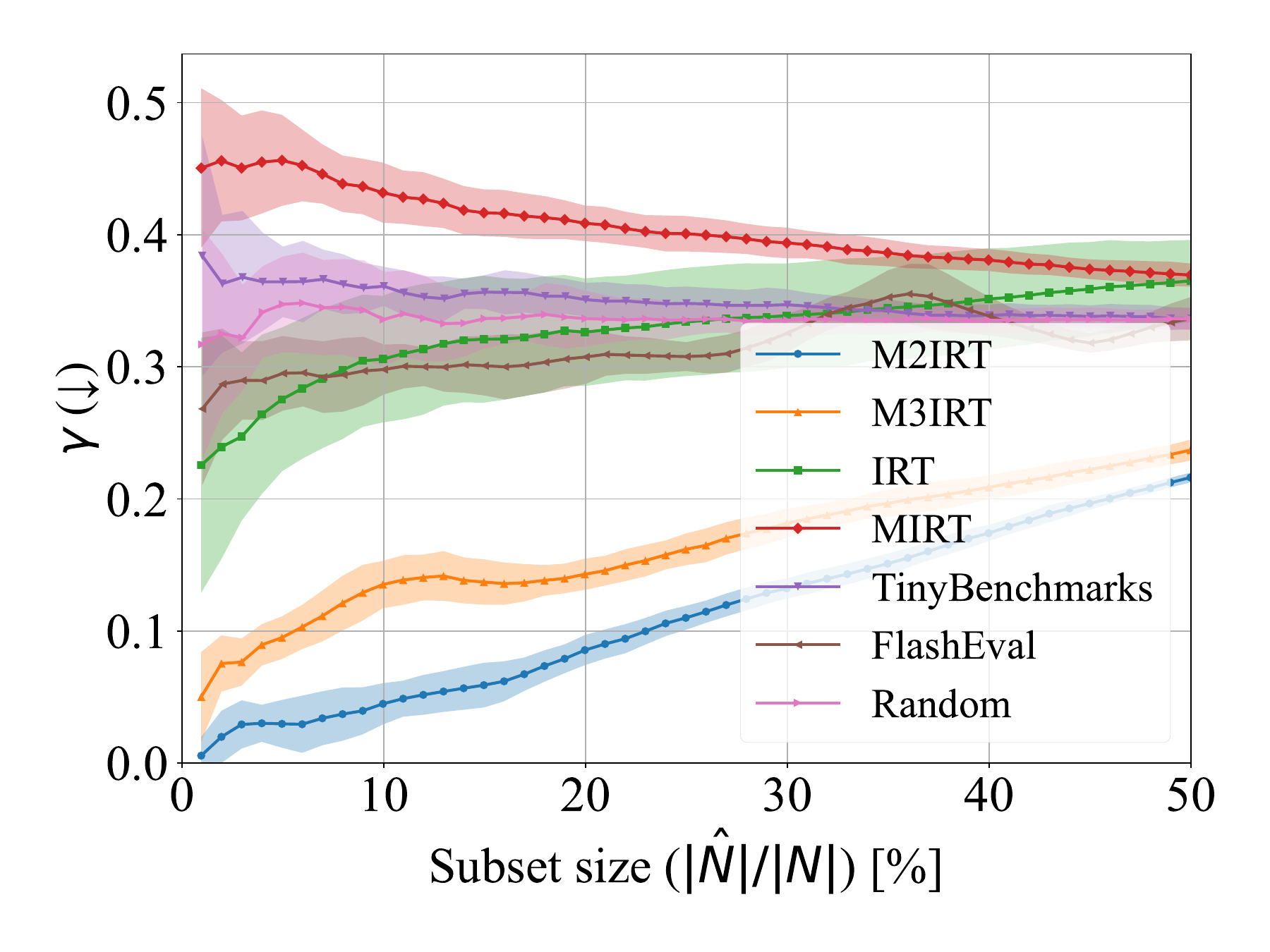}
        \caption{\textsc{SEED-Bench}}
        \label{fig:subset-contami-tate-seed}
    \end{subfigure}
    \caption{The average and standard deviation of the proportions of the low-quality questions in extracted question subsets $\gamma$ with different sizes.}
    \label{fig:subset-contami-tate}
\end{figure*}

We investigate whether a method can extract a compact subset of questions that enables us to evaluate the performance of unseen VLMs.
We randomly select a VLM from a collection of VLMs and construct a subset of the responses of remaining VLMs.
For a method, we select a subset of questions, estimate the performance of the VLM  from its responses to the subset, and obtain an estimated ranking of VLMs.
We compare the difference between rankings on the original benchmark for all models. 
We also investigate how much the artificial low-quality questions are included in the subset.

We use two measures to assess the quality of a subset $\hat{N} \subseteq N$ selected by a method.
First, we assess how much a method avoid the low-quality questions in the estimation of model rankings.
We compute the Spearman's rank correlation between model rankings on the original benchmark and the extracted subset.
Second, we evaluate how a method can distinguish between the original and low-quality questions.
We measure the proportion of low-quality questions in the extracted subset as 
\[
\gamma = \frac{|\{q \in \hat{N} \mid q\ \text{is a low-quality question}\}|}{|\hat{N}|}.
\]
We varied the subset size from 1\% to 50\% of the whole benchmark in 1\% increments.
We employed CAT with \mmirt{} using the maximum Fisher information in Sec.~\ref{sec:cat} and \mmmirt{} using D-Optimality.
We obtained the average and standard deviation from twenty four independent experiments.

\Cref{fig:subset-rank-comp} shows the Spearman's rank correlations between the model rankings on the original benchmark and on different sizes of subsets.
\Cref{fig:subset-contami-tate} shows the proportion $\gamma$ with varying size of subsets.

As shown in \Cref{fig:subset-rank-comp}, our methods accurately estimate model rankings from contaminated benchmarks, even with small subsets.
In \textsc{MMMU}, \mmirt{} achieves a rank correlation of 0.9 using only 3\% of the benchmark subset, and \mmmirt{} suprizingly achieves a rank correlation of 0.8 using the only 1\% subset.
FlashEval, which is SOTA but does not account for the presence of low-quality questions, performs similarly to Random.
In \textsc{MathVista}, \mmmirt{} achieves a rank correlation of 0.84 with a subset fraction of only 2\%, requiring 30\% to achieve a rank correlation of 0.9. 
In \textsc{Seed-Bench}, \mmirt{} achieves a rank correlation of 0.9 using only 3\% of the benchmark subset, while \mmmirt{} achieves the same rank correlation using only 1\% of the benchmark subset.

From \Cref{fig:subset-contami-tate}, we confirmed that the proportion of artificial low-quality questions included in the subset selected by the proposed method is significantly smaller compared to existing methods.
In \textsc{MMMU}, even with an extraction subset size of 50\%, all proposed methods keep the proportion of low-quality questions notably low at 24\%.
In contrast, the baseline methods choose substantially more low-quality questions than ours, which skew the estimated model rankings.
When extracting 30\% of \textsc{MathVista}, the rank correlation between \mmmirt{} and Random is about the same, but $\gamma$ is smaller for \mmmirt{}. 
We observed similar trends in the results of \textsc{Seed-Bench}.

\subsection{Robustness against Low-quality Questions}
\label{subsec:robust-auc}
\begin{figure*}
    \centering
    \begin{subfigure}[b]{0.32\textwidth}
        \centering
        \includegraphics[width=\textwidth]{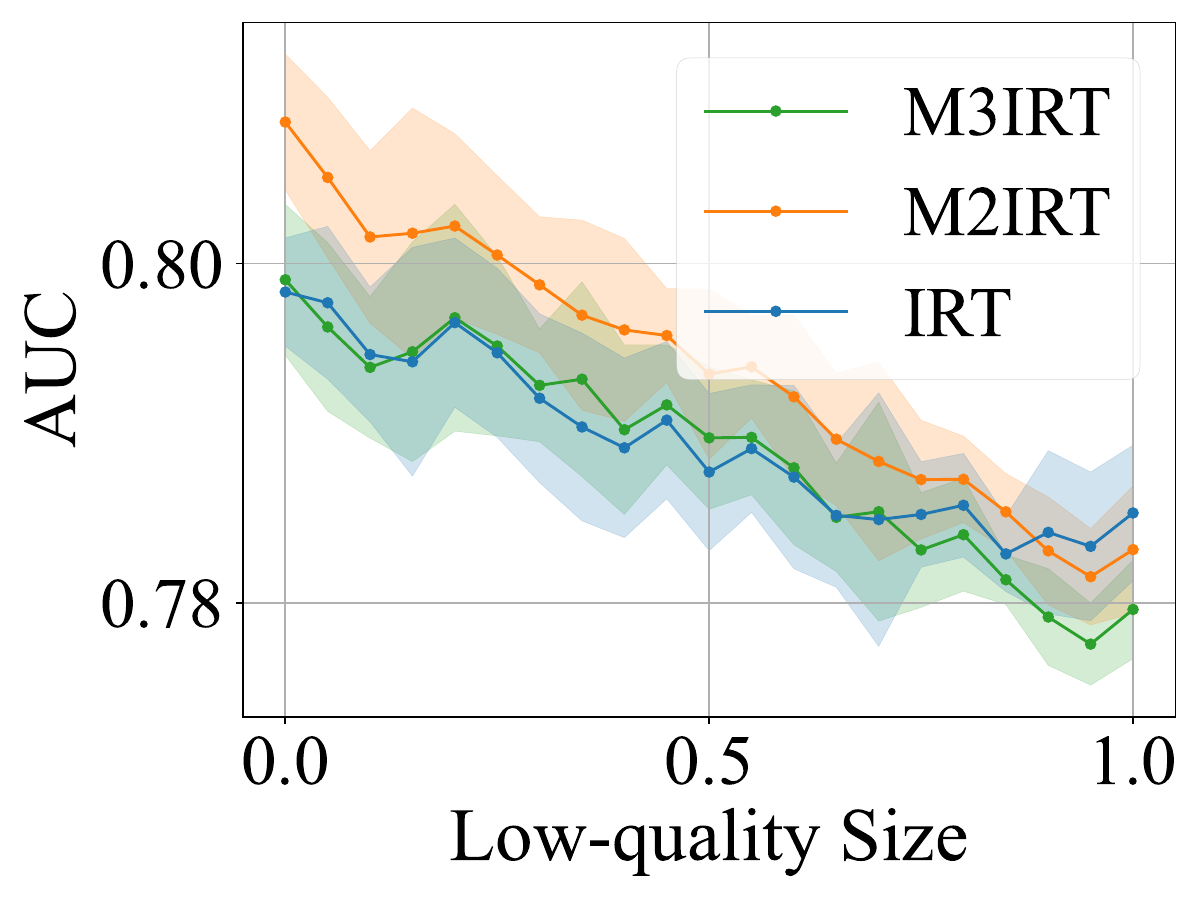}
        \caption{\textsc{MMMU}}
        \label{fig:auc-mmmu}
    \end{subfigure}
    ~
    \begin{subfigure}[b]{0.32\textwidth}
        \centering
        \includegraphics[width=\textwidth]{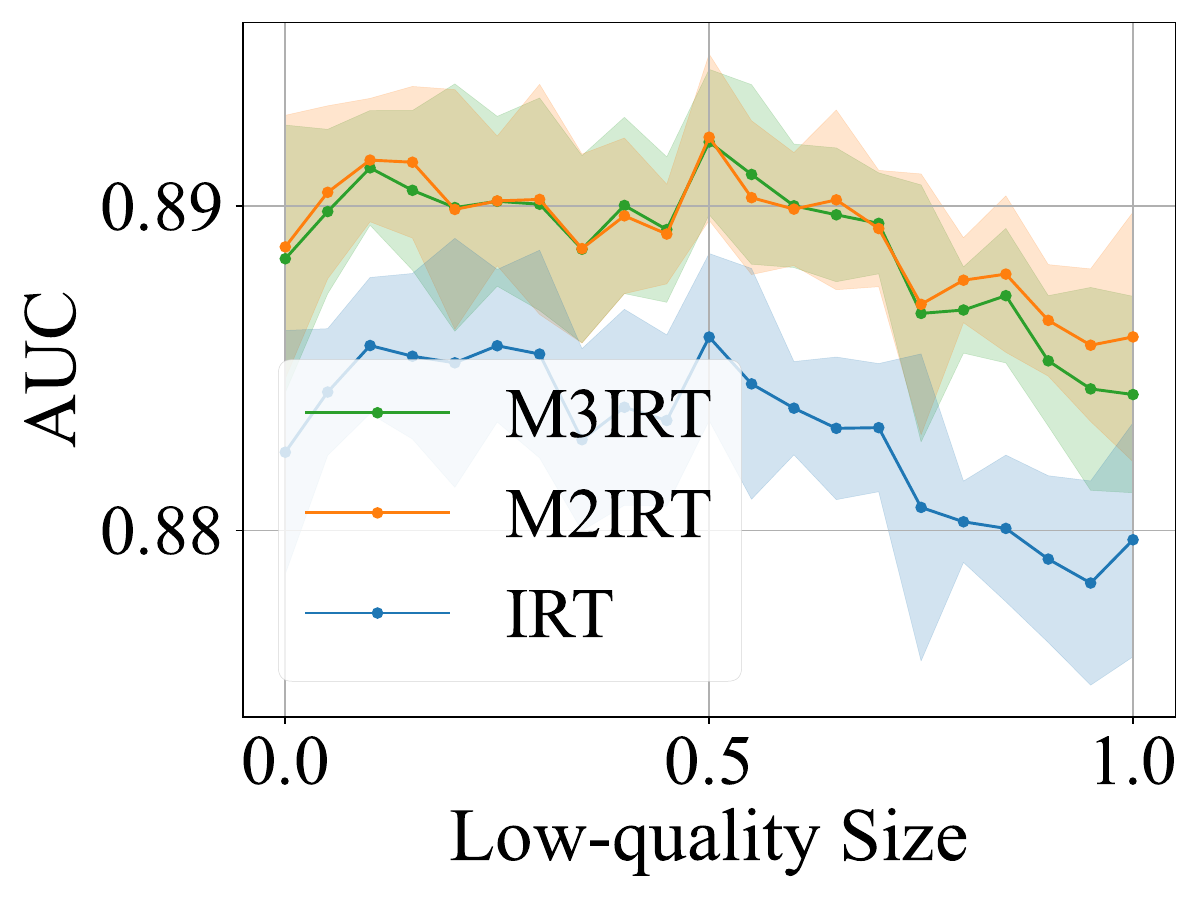}
        \caption{\textsc{MathVista}}
        \label{fig:auc-mathvista}
    \end{subfigure}
    ~
    \begin{subfigure}[b]{0.32\textwidth}
        \centering
        \includegraphics[width=\textwidth]{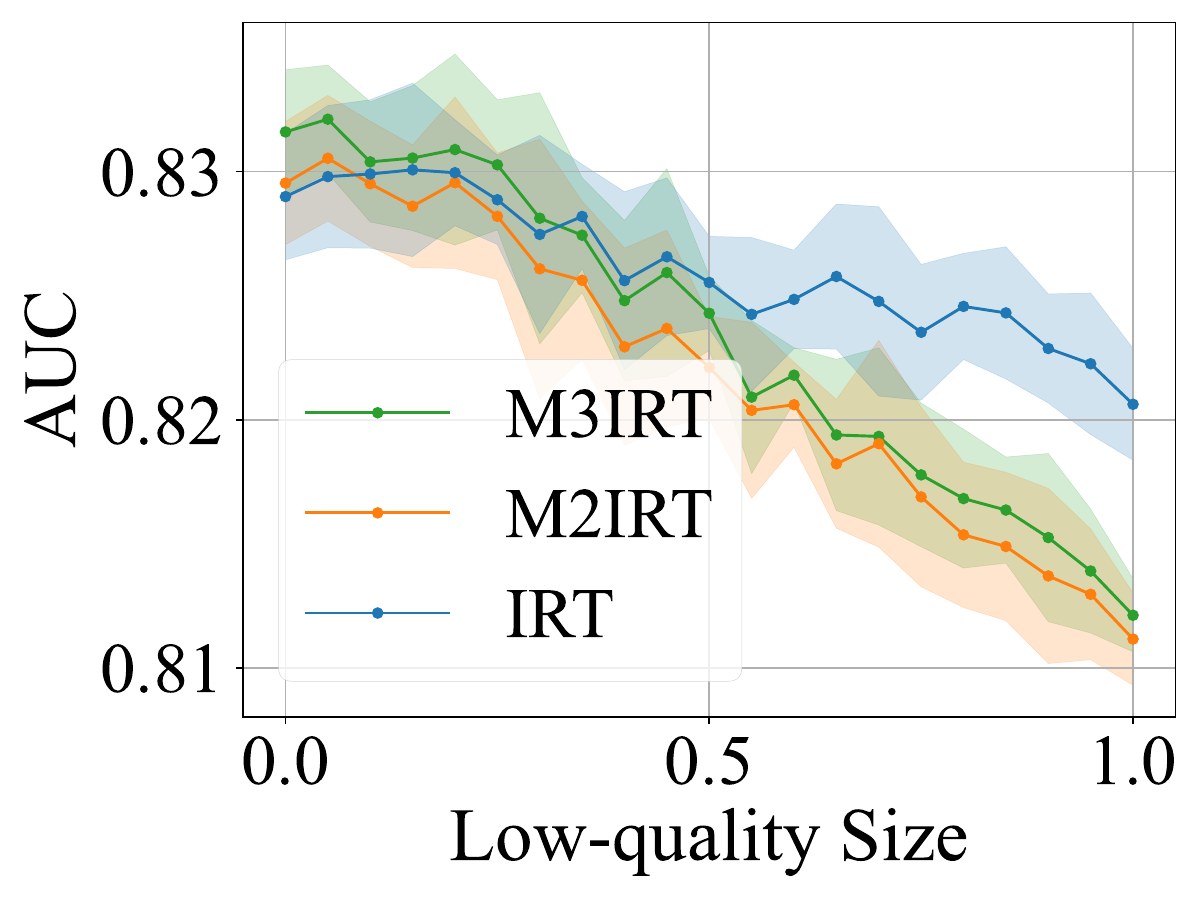}
        \caption{\textsc{SEED-Bench}}
        \label{fig:rand-auc-seed}
    \end{subfigure}
    \caption{ROC-AUC on predicting answers of the noisy benchmarks containing the different size of the low-quality questions inserted into the original benchmark.}
    \label{fig:auc}
\end{figure*}


We have evaluated the performance of the proposed method using a subset of questions. Here, we assess its performance as a latent variable method for predicting missing responses from observed ones. 
First, from the set of questions $N$, we randomly select 100 or 10\% questions each for validation and testing, using the remainder as training data. 
Next, we perform parameter estimation using the training data for both the proposed method and IRT. 
Finally, we evaluate the prediction performance on the test data using the estimated parameters with ROC-AUC. We measured ROC-AUC by varying the proportion of low-quality problems introduced in Sec. \ref{subsubsec:dataset}. We used IRT as a baseline in this experiment.
We obtained the average and standard deviation from ten independent experiments.

We show the results in \Cref{fig:auc}.
Our proposed methods achieved performance comparable to the standard IRT on ROC-AUC.
\mmirt{} was slightly better than IRT on \textsc{MMMU}, and comparable to IRT on \textsc{MATHVISTA} and \textsc{SEED-Bench}.
\mmmirt{} was slightly lower than IRT on \textsc{MATHVISTA} but the difference is small.
Even when low-quality questions are mixed in, the proposed method and IRT achieve ROC-AUC values around 0.8, suggesting that they effectively capture both the abilites of VLMs and the characteristics of the questions.

\section{Conclusion}
\label{sec:conclusion}
We addressed the challenge of assessing cross‑modal reasoning characteristics in MLLMs and multimodal benchmarks while reducing evaluation cost. We introduced \mmmirt{} and its variant \mmirt{}, which decompose both model ability and item difficulty of IRT into image‑only, text‑only, and cross‑modal components. This decomposition enables the identification of highly cross-modal items that require cross‑modal reasoning and supports lightweight assessment with far fewer items.

Across three benchmarks and 24 VLMs, we qualitatively evaluated that \mmmirt{} can estimate the degree to which an item requires cross‑modal reasoning, and assigns higher cross‑modal difficulty to genuinely cross‑modal items than to single‑modality shortcut.
Moreover, analyses with synthetically contaminated benchmarks confirmed that \mmmirt{} and \mmirt{} yields evaluations aligned with the original benchmarks, demonstrating robustness to low‑quality contamination.

\paragraph{Limitations and future work.}
Our study focuses on multiple‑choice, which is a typical form of closed‑ended questions. Extending the framework to open‑ended settings with open-ended questions is a natural next step, enabling the discovery of items that demand stronger cross‑modal reasoning and the evaluation of MLLMs under generative outputs. Beyond vision–language, applying the approach to additional modalities (e.g., audio, actions) and developing question‑generation methods that control cross‑modal difficulty are promising directions.

\section*{Ethics statement}
This work adheres to the ICLR Code of Ethics.
All datasets used in this paper (MMMU, MathVista, SEED-Bench) are publicly released benchmarks, and we strictly followed their respective licenses.
\subsubsection*{Acknowledgments}
\SU{
This work was supported by JST FOREST Program Grant Number JPMJFR232S, JST CREST Grant Number JPMJCR21D1, JST BOOST Grant Number JPMJBY24E8.
}
\bibliography{iclr2026_conference}

@inproceedings{liu2024mmbench,
  title={Mmbench: Is your multi-modal model an all-around player?},
  author={Liu, Yuan and Duan, Haodong and Zhang, Yuanhan and Li, Bo and Zhang, Songyang and Zhao, Wangbo and Yuan, Yike and Wang, Jiaqi and He, Conghui and Liu, Ziwei and others},
  booktitle={European conference on computer vision (ECCV)},
  year={2024},
}

@inproceedings{
yang2025dynamic,
title={Dynamic Multimodal Evaluation with Flexible Complexity by Vision-Language Bootstrapping},
author={Yue Yang and Shuibo Zhang and Kaipeng Zhang and Yi Bin and Yu Wang and Ping Luo and Wenqi Shao},
booktitle={The Thirteenth International Conference on Learning Representations (ICLR)},
year={2025},
}

@inproceedings{
jiang2025mac,
title={{MAC}: A Live Benchmark for Multimodal Large Language Models in Scientific Understanding},
author={Mohan Jiang and Jin Gao and Jiahao Zhan and Dequan Wang},
booktitle={Second Conference on Language Modeling (COLM)},
year={2025},
}

@inproceedings{
shabtay2025livexiv,
title={LiveXiv - A Multi-Modal live benchmark based on Arxiv papers content},
author={Nimrod Shabtay and Felipe Maia Polo and Sivan Doveh and Wei Lin and Muhammad Jehanzeb Mirza and Leshem Choshen and Mikhail Yurochkin and Yuekai Sun and Assaf Arbelle and Leonid Karlinsky and Raja Giryes},
booktitle={The Thirteenth International Conference on Learning Representations (ICLR)},
year={2025},
}

@inproceedings{
hao2025can,
title={Can {MLLM}s Reason in Multimodality? {EMMA}: An Enhanced MultiModal ReAsoning Benchmark},
author={Yunzhuo Hao and Jiawei Gu and Huichen Will Wang and Linjie Li and Zhengyuan Yang and Lijuan Wang and Yu Cheng},
booktitle={Forty-second International Conference on Machine Learning (ICML)},
year={2025},
}

@inproceedings{zhang-etal-2025-cchall,
    title = "{CCH}all: A Novel Benchmark for Joint Cross-Lingual and Cross-Modal Hallucinations Detection in Large Language Models",
    author = "Zhang, Yongheng  and
      Liu, Xu  and
      Zhou, Ruoxi  and
      Chen, Qiguang  and
      Fei, Hao  and
      Lu, Wenpeng  and
      Qin, Libo",
    booktitle = "Proceedings of the 63rd Annual Meeting of the Association for Computational Linguistics (ACL)",
    year = "2025",
}

@InProceedings{Li_2024_CVPR_seed,
    author    = {Li, Bohao and Ge, Yuying and Ge, Yixiao and Wang, Guangzhi and Wang, Rui and Zhang, Ruimao and Shan, Ying},
    title     = {SEED-Bench: Benchmarking Multimodal Large Language Models},
    booktitle = {Proceedings of the IEEE/CVF Conference on Computer Vision and Pattern Recognition (CVPR)},
    year      = {2024},
}

@inproceedings{NEURIPS2024_2f8ee6a3,
    author = {Lin Chen and Jinsong Li and Xiaoyi Dong and Pan Zhang and Yuhang Zang and Zehui Chen and Haodong Duan and Jiaqi Wang and Yu Qiao and Dahua Lin and Feng Zhao},
    title = {Are we on the right way for evaluating large vision-language models?},
    booktitle = {Advances in Neural Information Processing Systems (NeurIPS)},
    year = {2024},
}

@inproceedings{yue2024mmmu,
    title = "{MMMU}-Pro: A More Robust Multi-discipline Multimodal Understanding Benchmark",
    author = "Yue, Xiang  and
      Zheng, Tianyu  and
      Ni, Yuansheng  and
      Wang, Yubo  and
      Zhang, Kai  and
      Tong, Shengbang  and
      Sun, Yuxuan  and
      Yu, Botao  and
      Zhang, Ge  and
      Sun, Huan  and
      Su, Yu  and
      Chen, Wenhu  and
      Neubig, Graham",
    booktitle = "Proceedings of the 63rd Annual Meeting of the Association for Computational Linguistics (ACL)",
    year = "2025",
}

@inproceedings{autorageval2024,
    author = {Guinet, Gauthier and Omidvar-Tehrani, Behrooz and Deoras, Anoop and Callot, Laurent},
    title = {Automated evaluation of retrieval-augmented language models with task-specific exam generation},
    year = {2024},
    booktitle = {Proceedings of the 41st International Conference on Machine Learning (ICML)},
}

@inproceedings{Yue_2024_CVPR_MMMU,
    author = {Xiang Yue and Yuansheng Ni and Kai Zhang and Tianyu Zheng and Ruoqi Liu and Ge Zhang and Samuel Stevens and Dongfu Jiang and Weiming Ren and Yuxuan Sun and Cong Wei and Botao Yu and Ruibin Yuan and Renliang Sun and Ming Yin and Boyuan Zheng and Zhenzhu Yang and Yibo Liu and Wenhao Huang and Huan Sun and Yu Su and Wenhu Chen},
    title = {MMMU: A massive multi-discipline multimodal understanding and reasoning benchmark for expert AGI},
    booktitle = {Proceedings of the IEEE/CVF Conference on Computer Vision and Pattern Recognition (CVPR)},
    year = {2024},
}

@inproceedings{lu2024mathvista,
    author = {Pan Lu and Hritik Bansal and Tony Xia and Jiacheng Liu and Chunyuan Li and Hannaneh Hajishirzi and Hao Cheng and Kai-Wei Chang and Michel Galley and Jianfeng Gao},
    title = {MathVista: Evaluating mathematical reasoning of foundation models in visual contexts},
    booktitle = {International Conference on Learning Representations (ICLR)},
    year = {2024}
}

@inproceedings{chen2023vqa,
    author = {Chongyan Chen and Samreen Anjum and Danna Gurari},
    title = {Vqa therapy: Exploring answer differences by visually grounding answers},
    booktitle = {Proceedings of the IEEE International Conference on Computer Vision (ICCV)},
    year = {2023}
}

@inproceedings{gurari2018vizwiz,
    author = {Danna Gurari and Qing Li and Abigale J Stangl and Anhong Guo and Chi Lin and Kristen Grauman and Jiebo Luo and Jeffrey P Bigham},
    title = {Vizwiz grand challenge: Answering visual questions from blind people},
    booktitle = {Proceedings of the IEEE Conference on Computer Vision and Pattern Recognition (CVPR)},
    year = {2018},
}

@inproceedings{balanced_vqa_v2,
    author = {Yash Goyal and Tejas Khot and Douglas Summers-Stay and Dhruv Batra and Devi Parikh},
    title = {Making the V in VQA matter: Elevating the role of image understanding in Visual Question Answering},
    booktitle = {Proceedings of the IEEE Conference on Computer Vision and Pattern Recognition (CVPR)},
    year = {2017}
}

@inproceedings{Zhao_2024_CVPR_FlashEval,
    author = {Lin Zhao and Tianchen Zhao and Zinan Lin and Xuefei Ning and Guohao Dai and Huazhong Yang and Yu Wang},
    title = {FlashEval: Towards fast and accurate evaluation of text-to-image diffusion generative models},
    booktitle = {Proceedings of the IEEE/CVF Conference on Computer Vision and Pattern Recognition (CVPR)},
    year = {2024},
}

@inproceedings{lalor-etal-2016-building-irt-nlp,
    author = {John P Lalor and Hao Wu and Hong Yu},
    title = {Building an evaluation scale using item response theory},
    booktitle = {Proceedings of the 2016 Conference on Empirical Methods in Natural Language Processing(EMNLP)},
    year = {2016},
    pages = {648--657},
}

@misc{Qwen2.5-VL,
    title={Qwen2.5-VL Technical Report},
    author={Bai, Shuai and Chen, Keqin and Liu, Xuejing and Wang, Jialin and Ge, Wenbin and Song, Sibo and Dang, Kai and Wang, Peng and Wang, Shijie and Tang, Jun and Zhong, Humen and Zhu, Yuanzhi and Yang, Mingkun and Li, Zhaohai and Wan, Jianqiang and Wang, Pengfei and Ding, Wei and Fu, Zheren and Xu, Yiheng and Ye, Jiabo and Zhang, Xi and Xie, Tianbao and Cheng, Zesen and Zhang, Hang and Yang, Zhibo and Xu, Haiyang and Lin, Junyang},
    journal={arXiv preprint arXiv:2502.13923},
    year={2025}
}

@article{cat,
    author = {Han, Kyung Chris T},
    year = {2018},
    month = {03},
    pages = {7},
    title = {Components of item selection algorithm in computerized adaptive testing},
    volume = {15},
    journal = {Journal of Educational Evaluation for Health Professions},
}

@inproceedings{hirai2023applying,
    title = "Applying Item Response Theory to Task-oriented Dialogue Systems for Accurately Determining User`s Task Success Ability",
    author = "Hirai, Ryu and
      Guo, Ao and Higashinaka, Ryuichiro",
    booktitle = "Proceedings of the 24th Annual Meeting of the Special Interest Group on Discourse and Dialogue(SIGDIAL)",
    year = "2023",
    pages = "421--427",
}

@inproceedings{liu2023what,
    author = {Liu, Yang and Medlar, Alan and Glowacka, Dorota},
    title = {What We Evaluate When We Evaluate Recommender Systems: Understanding Recommender Systems’ Performance using Item Response Theory},
    year = {2023},
    booktitle = {Proceedings of the 17th ACM Conference on Recommender Systems (RecSys)},
}

@misc{paszke2019pytorch,
    title={Pytorch: An imperative style, high-performance deep learning library},
    author={Paszke, A},
    journal={arXiv preprint arXiv:1912.01703},
    year={2019}
}

@inproceedings{kipnis2025metabenchsparsebenchmark,
    author = {Alex Kipnis and Konstantinos Voudouris and Luca M Schulze Buschoff and Eric Schulz},
    title = {metabench -- A sparse benchmark of reasoning and knowledge in large language models},
    booktitle = {International Conference on Learning Representations(ICLR)},
    year = {2025}
}

@misc{zhuang2025positionaievaluationlearn,
    author = {Yan Zhuang and Qi Liu and Yuting Ning and Weizhe Huang and Rui Lv and Zhenya Huang and Guanhao Zhao and Zheng Zhang and Qingyang Mao and Shijin Wang and Enhong Chen},
    title = {Efficiently measuring the cognitive ability of LLMs: An adaptive testing perspective},
    journal = {arXiv preprint arXiv:2306.10512},
    year = {2023}
}

@misc{polo2024tinybenchmarksevaluatingllmsfewer,
    author = {Felipe Maia Polo and Lucas Weber and Leshem Choshen and Yuekai Sun and Gongjun Xu and Mikhail Yurochkin},
    title = {tinyBenchmarks: Evaluating LLMs with fewer examples},
    journal = {arXiv preprint arXiv:2402.14992},
    year = {2024}
}

@misc{xu2024dataefficientevaluationlarge,
    author = {Cong Xu and Gayathri Saranathan and Mahammad Parwez Alam and Arpit Shah and James Lim and Soon Yee Wong and Martin Foltin and Suparna Bhattacharya},
    title = {Data efficient evaluation of large language models and text-to-image models via adaptive sampling},
    journal = {arXiv preprint arXiv:2406.15527},
    year = {2024}
}

@misc{huang2024activetestinglargelanguage,
    author = {Yuheng Huang and Jiayang Song and Qiang Hu and Felix Juefei-Xu and Lei Ma},
    title = {Active testing of large language model via multi-stage sampling},
    journal = {arXiv preprint arXiv:2408.03573},
    year = {2024}
}

@misc{li2024activeevaluationacquisitionefficient,
    author = {Yang Li and Jie Ma and Miguel Ballesteros and Yassine Benajiba and Graham Horwood},
    title = {Active evaluation acquisition for efficient LLM benchmarking},
    journal = {arXiv preprint arXiv:2410.05952},
    year = {2024}
}

@misc{yuan2025onesizefitsalltailoredbenchmarksefficient,
    author = {Peiwen Yuan and Yueqi Zhang and Shaoxiong Feng and Yiwei Li and Xinglin Wang and Jiayi Shi and Chuyi Tan and Boyuan Pan and Yao Hu and Kan Li},
    title = {Beyond one-size-fits-all: Tailored benchmarks for efficient evaluation},
    journal = {arXiv preprint arXiv:2502.13576},
    year = {2025}
}

@inproceedings{
saranathan2024dele,
title={{DELE}: Data Efficient {LLM} Evaluation},
author={Gayathri Saranathan and Mahammad Parwez Alam and James Lim and Suparna Bhattacharya and Soon Yee Wong and Martin Foltin and Cong Xu},
booktitle={ICLR 2024 Workshop on Navigating and Addressing Data Problems for Foundation Models(ICLR Workshop)},
year={2024}
}

@misc{ye2023predictablelargelanguagemodel,
   title        = {How Predictable Are Large Language Model Capabilities? A Case Study on BIG-bench},
   author       = {Qinyuan Ye and Harvey Yiyun Fu and Xiang Ren and Robin Jia},
   journal      = {arXiv preprint arXiv:2305.14947},
   year         = {2023}
}

@inproceedings{vivek-etal-2024-anchor,
    author = {Rajan Vivek and Kawin Ethayarajh and Diyi Yang and Douwe Kiela},
    title = {Anchor points: Benchmarking models with much fewer examples},
    booktitle = {Proceedings of the 18th European Chapter of the Association for Computational Linguistics (EACL)},
    year = {2024},
    pages = {1576--1601}
}

@article{jiang2025raisingbarinvestigatingvalues,
    author = {Han Jiang and Xiaoyuan Yi and Zhihua Wei and Ziang Xiao and Shu Wang and Xing Xie},
    title = {Raising the bar: Investigating the values of large language models via generative evolving testing},
    journal = {arXiv preprint arXiv:2406.14230},
    year = {2025}
}

@misc{zhuang2023fromstaticbenchmarkstoadaptivetesting,
    author = {Yan Zhuang and Qi Liu and Yuting Ning and Weizhe Huang and Zachary A Pardos and Patrick C Kyllonen and Jiyun Zu and Qingyang Mao and Rui Lv and Zhenya Huang and Enhong Chen},
    title = {From static benchmarks to adaptive testing: Psychometrics in AI evaluation},
    journal = {arXiv preprint arXiv:2306.10512},
    year = {2023}
}

@book{Lord1980Applications,
    author = {Frederic M Lord},
    title = {Applications of item response theory to practical testing problems},
    year = {1980},
    publisher = {Lawrence Erlbaum Associates}
}

@article{10445007,
    author = {Jingyi Zhang and Jiaxing Huang and Sheng Jin and Shijian Lu},
    title = {Vision-language models for vision tasks: A survey},
    journal = {IEEE Transactions on Pattern Analysis and Machine Intelligence},
    year = {2024},
    volume = {46},
    number = {8},
    pages = {5625--5644}
}

@article{mirt-cat-mulder,
    author = {Joris Mulder and Wim Linden},
    title = {Multidimensional adaptive testing with optimal design criteria for item selection},
    journal = {Psychometrika},
    year = {2009},
    volume = {74},
    pages = {273--296}
}

@misc{meta2024llama,
    author = {Meta},
    title = {Llama 3.2: Revolutionizing edge AI and vision with open, customizable models},
    year = {2024},
    howpublished = {https://ai.meta.com/blog/llama-3-2-connect-2024-vision-edge-mobile-devices/}
}

@article{agrawal2024pixtral,
    author = {Pravesh Agrawal and Szymon Antoniak and Emma Bou Hanna and Baptiste Bout and Devendra Chaplot and Jessica Chudnovsky and Diogo Costa and Baudouin De Monicault and Saurabh Garg and Theophile Gervet and Robin Lutz},
    title = {Pixtral 12B},
    journal = {arXiv preprint arXiv:2410.07073},
    year = {2024}
}

@article{minimax2025minimax01scalingfoundationmodels,
    author = {MiniMax Team},
    title = {MiniMax-01: Scaling foundation models with lightning attention},
    journal = {arXiv preprint arXiv:2501.08313},
    year = {2025}
}

@article{10.1093/nsr/nwae403,
    author = {Shukang Yin and Chaoyou Fu and Sirui Zhao and Ke Li and Xing Sun and Tong Xu and Enhong Chen},
    title = {A survey on multimodal large language models},
    journal = {National Science Review},
    year = {2024},
    volume = {11},
    number = {12},
    pages = {nwae403},
}

@inproceedings{Jiang_2023_CVPR,
    author = {Ding Jiang and Mang Ye},
    title = {Cross-modal implicit relation reasoning and aligning for text-to-image person retrieval},
    booktitle = {Proceedings of the IEEE/CVF Conference on Computer Vision and Pattern Recognition(CVPR)},
    year = {2023},
}

@article{doi:10.1177/0013164498058003001,
    author = {Xitao Fan},
    title = {Item response theory and classical test theory: An empirical comparison of their item/person statistics},
    journal = {Educational and Psychological Measurement},
    year = {1998},
    volume = {58},
    number = {3},
    pages = {357--381},
    doi = {10.1177/0013164498058003001}
}

@inproceedings{kingma2014adam,
    title={Adam: A method for stochastic optimization},
    author={Kingma, Diederik P and Ba, Jimmy},
    booktitle={International Conference on Learning Representations(ICLR)},
    year={2014}
}

@article{9c2d3927-7193-3702-9179-0a57b6a5e7e0,
    author = {David J Weiss and G Gage Kingsbury},
    title = {Application of computerized adaptive testing to educational problems},
    journal = {Journal of Educational Measurement},
    year = {1984},
    volume = {21},
    number = {4},
    pages = {361--375}
}

@misc{google2024gemini15,
    author = {Google Gemini Team},
    title = {Gemini 1.5: Unlocking multimodal understanding across millions of tokens of context},
    year = {2024},
    howpublished = {https://storage.googleapis.com/deepmind-media/gemini/gemini\_v1\_5\_report.pdf}
}

@misc{openai2024gpt4ocard,
    author = {OpenAI},
    title = {Hello gpt-4o},
    year = {2024},
    howpublished = {https://openai.com/index/hello-gpt-4o}
}

@misc{openai2024gpt4omini,
    author = {OpenAI},
    title = {GPT-4o mini: Advancing cost-efficient intelligence},
    year = {2024},
    howpublished = {https://openai.com/index/gpt-4o-mini-advancing-cost-efficient-intelligence}
}

@misc{openai2025gpt41,
    author = {OpenAI},
    title = {Introducing GPT-4.1 in the API},
    year = {2025},
    howpublished = {https://openai.com/index/gpt-4-1}
}

@misc{Intelligence2024,
    author = {Amazon Artificial General Intelligence},
    title = {The Amazon Nova family of models: Technical report and model card},
    year = {2024},
    howpublished = {https://www.amazon.science/publications/the-amazon-nova-family-of-models-technical-report-and-model-card}
}

@misc{pichai2024gemini,
    author = {Sundar Pichai},
    title = {Introducing Gemini 2.0: Our new AI model for the agentic era},
    year = {2024},
    howpublished = {https://blog.google/technology/google-deepmind/google-gemini-ai-update-december-2024/\#ceo-message}
}

@misc{anthropic2024claude3,
    author = {Anthropic},
    title = {Introducing the next generation of Claude},
    year = {2024},
    howpublished = {https://www.anthropic.com/news/claude-3-family}
}

@misc{anthropic2024claude35,
    author = {Anthropic},
    title = {Claude 3.5 Sonnet},
    year = {2024},
    howpublished = {https://www.anthropic.com/news/claude-3-5-sonnet}
}

@online{anthropic2025claude37,
    author       = {Anthropic},
    title        = {Claude 3.7 Sonnet and Claude Code},
    year         = {2025},
    howpublished          = {https://www.anthropic.com/news/claude-3-7-sonnet},
}

@misc{grok2,
    author = {xAI},
    title = {Grok-2 Beta Release},
    year = {2024},
    howpublished = {https://x.ai/news/grok-2}
}

@book{reckase2009multidimensional,
  title={Multidimensional Item Response Theory},
  author={Reckase, M. D.},
  year={2009},
  publisher={Springer Science \& Business Media}
}
\bibliographystyle{iclr2026_conference}

\newpage
\appendix
\section{Details of Low-quality Question Generation}\label{appendix:low-question}
For \textsc{MMMU}, we made three types of low-quality questions: (A) 300 questions consisting of the image, text, and multiple-choice selected from all different questions, (B) 300 questions where the image was replaced with that from different questions; and (C) 300 questions where the text was replaced with that from different questions.
For \textsc{MathVista}, we made 333 questions each for (A), (B), and (C).  
For \textsc{SEED-Bench}, we made 333 questions each for (A), (B), and (C).  

\section{Alignment with Human Answering Patterns}
\label{appendix:human}
\SU{To investigate whether the cross-modal difficulty $\bsynergy$ estimated by \mmmirt{}
corresponds to human answering patterns, we conducted a crowdsourcing experiment.
We randomly sampled 200 questions per benchmark.
Nine crowd workers answered each question under three input formats:
Image+Text, Image-only, and Text-only.
For each question $j$, we computed the accuracy under each format
and defined two measures of cross-modal synergy:
\begin{align}
    \text{Decline}_{\text{mean},j} &= \text{Acc}_{\text{image+text},j}
      - \frac{\text{Acc}_{\text{image},j}+\text{Acc}_{\text{text},j}}{2}, \\
    \text{Decline}_{\text{max},j} &= \text{Acc}_{\text{image+text},j}
      - \max(\text{Acc}_{\text{image},j},\;\text{Acc}_{\text{text},j}).
\end{align}
$\text{Decline}_{\text{mean}}$ measures how much the joint presentation improves
over the average single-modality performance,
while $\text{Decline}_{\text{max}}$ measures the improvement over the
best single modality.
We report Spearman's rank correlation between $\bsynergy$ and each measure.
Inter-annotator agreement is assessed by Fleiss' $\kappa$,
where values above 0.6 are generally considered acceptable.}

\SU{\Cref{tab:sheet1_data} summarizes the results.
On \textsc{SEED-Bench}, we observed weak positive correlations
between $\bsynergy$ and both $\text{Decline}_{\text{mean}}$ (0.35)
and $\text{Decline}_{\text{max}}$ (0.35),
with moderate inter-annotator agreement ($\kappa=0.44$)
and crowd worker accuracy of 0.62 (compared to 0.57 for VLMs).
On \textsc{MMMU} and \textsc{MathVista}, the correlations were weak or negative
($\text{Decline}_{\text{mean}}$: 0.077 and $-0.16$;
$\text{Decline}_{\text{max}}$: 0.090 and $-0.15$, respectively),
with low inter-annotator agreement ($\kappa=0.25$ and $0.28$).
Crowd worker accuracy on these benchmarks was notably low
($\text{Acc}_{\text{image+text}}=0.41$ and $0.23$),
whereas VLMs achieved 0.52 on both.
These results indicate that VLMs and humans exhibit
different cross-modal answering patterns:
the questions that \mmmirt{} identifies as requiring strong cross-modal reasoning
for VLMs do not necessarily pose the same cross-modal demand for human solvers.
This difference was observed across all three benchmarks, though to varying degrees,
and was larger on \textsc{MMMU} and \textsc{MathVista},
where crowd worker accuracy was notably low.}

\SU{For the human evaluation, we collected responses from nine crowd workers via Amazon Mechanical Turk (MTurk).
The tasks consisted solely of answering multiple-choice questions
drawn from the aforementioned public benchmarks.
No personally identifiable or sensitive information was collected.}

\begin{table}[h]
\centering
\caption{Relationships between VLMs and human answer patterns}
\begin{tabular}{lccc}
\toprule
& MMMU & MathVista & SEED-Bench \\
\midrule
$\text{Decline}_{max}$ & 0.032 & -0.23 & 0.18 \\
Corr. between $\text{Decline}_{max}$ and $\bsynergy$ & 0.090 & -0.15 & 0.35 \\
$\text{Decline}_{mean}$ & 0.14 & -0.11 & 0.31 \\
Corr. between $\text{Decline}_{mean}$ and $\bsynergy$ & 0.077 & -0.16 & 0.35 \\
Fleiss' $\kappa$ & 0.25 & 0.28 & 0.44 \\
Accuracy of Crowd workers & 0.41 & 0.23 & 0.62 \\
Accuracy of VLMs & 0.52 & 0.52 & 0.57 \\
\bottomrule
\end{tabular}
\label{tab:sheet1_data}
\end{table}

\section{Omitted Results of Multimodal Difficulty and Ability Decomposition}\label{appendix:b}
\begin{figure}
    \centering
     \quad
    \begin{subfigure}[b]{0.75\textwidth}
        \centering
        \includegraphics[width=\textwidth]{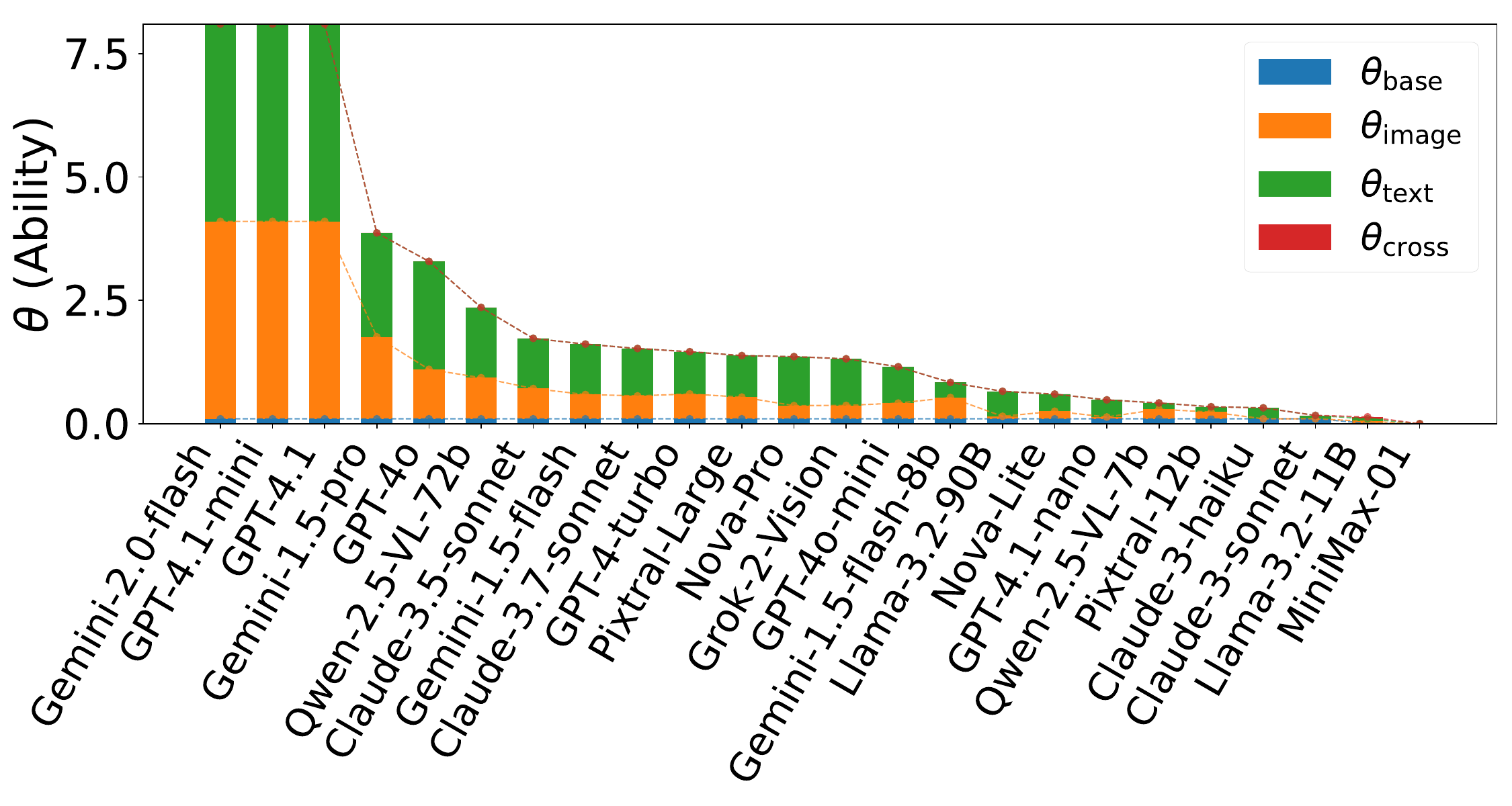}
        \caption{\textsc{MMMU}}
        \label{fig:theta-onedim-mmmu}
    \end{subfigure}
    \begin{subfigure}[b]{0.75\textwidth}
        \centering
        \includegraphics[width=\textwidth]{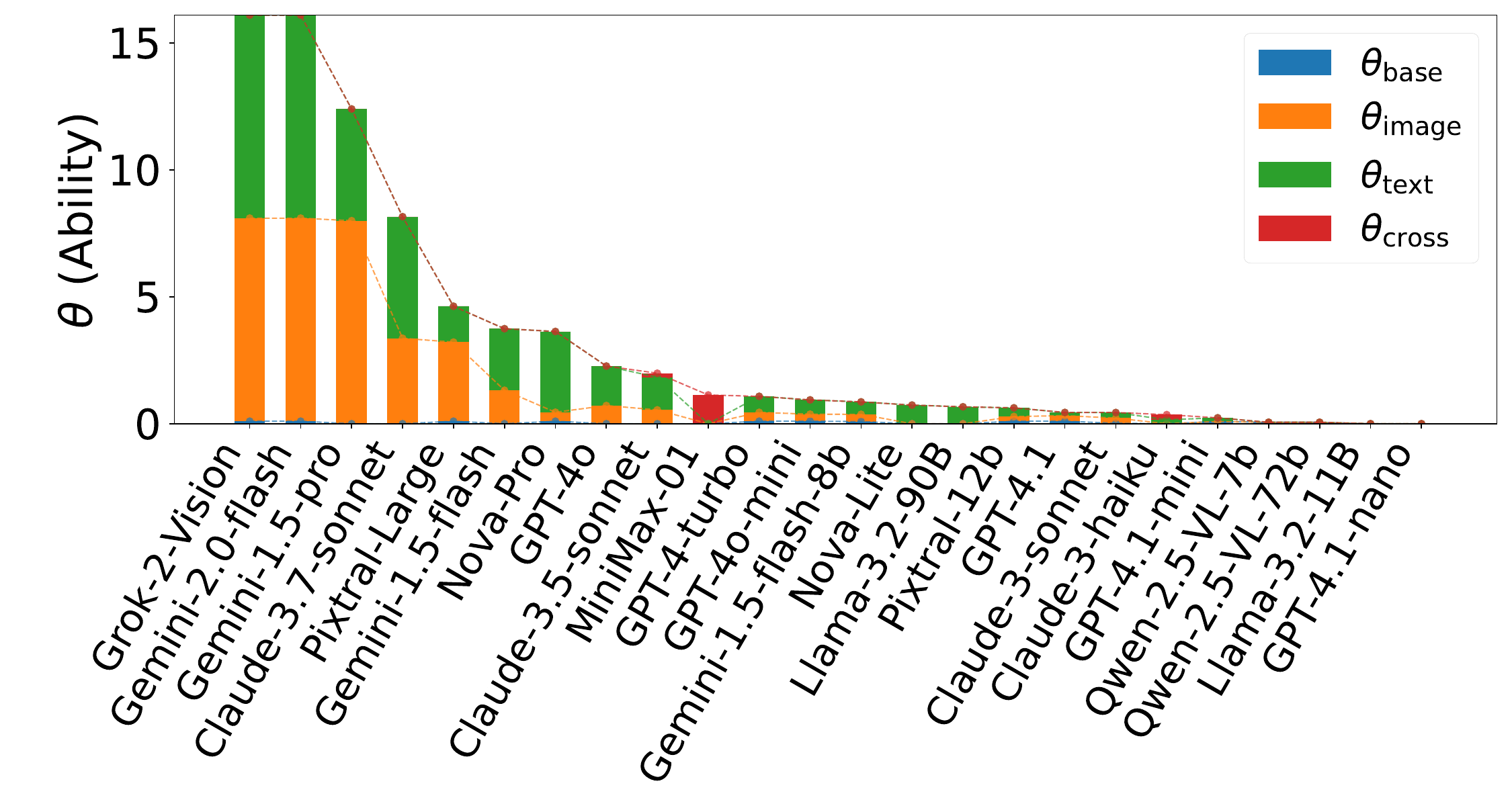}
        \caption{\textsc{MathVista}}
        \label{fig:theta-onedim-mathvista}
    \end{subfigure}
    \begin{subfigure}[b]{0.75\textwidth}
        \centering
        \includegraphics[width=\textwidth]{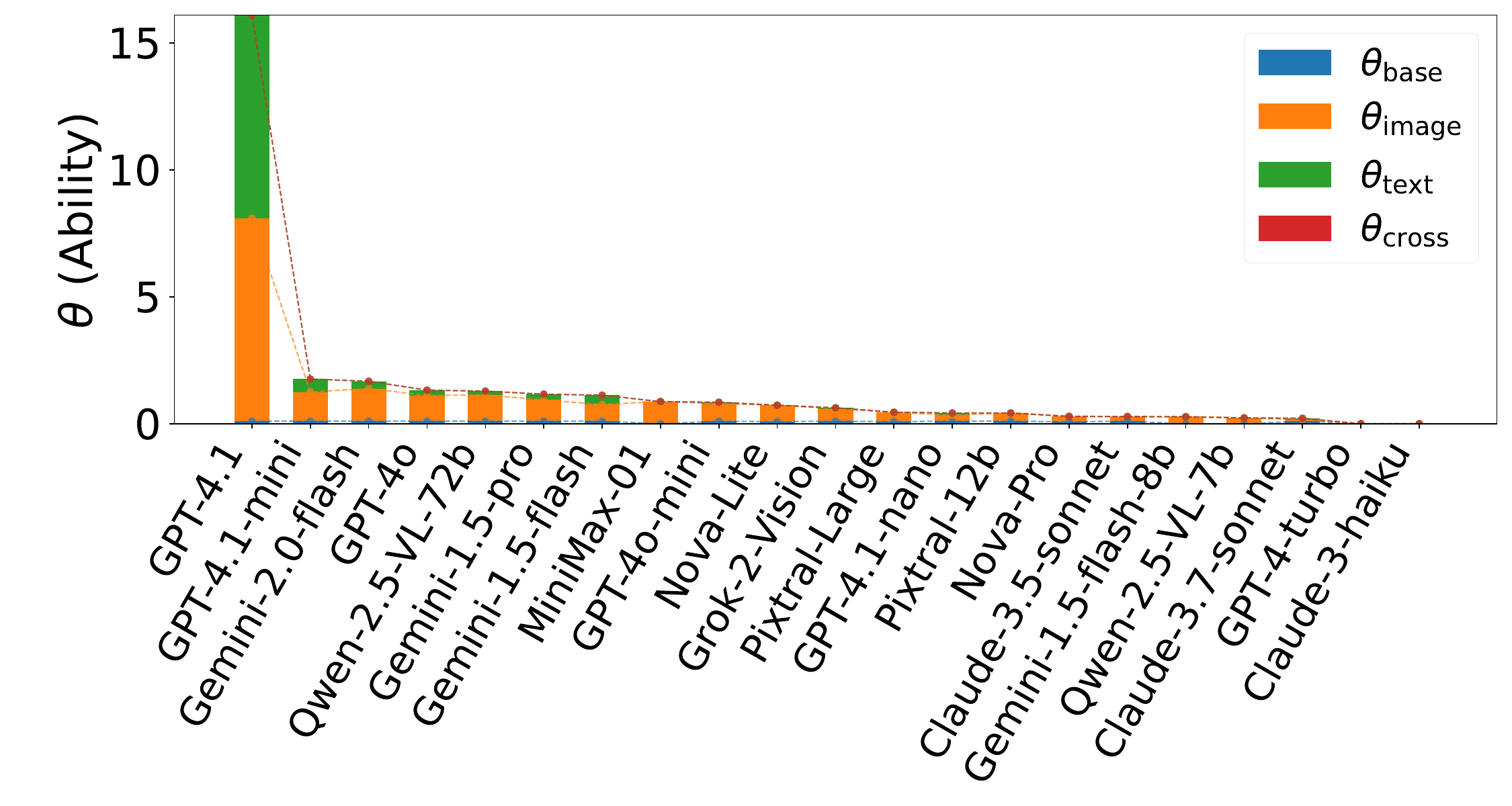}
        \caption{\textsc{SEED-Bench}}
        \label{fig:theta-onedim-seed}
    \end{subfigure}
    \caption{Distributions of $\theta$ estimated by \mmirt{} sorted in descending order.}
    \label{fig:appendix-theta-onedim}
\end{figure}

\begin{figure}
    \centering
     \quad
    \begin{subfigure}[b]{0.75\textwidth}
        \centering
        \includegraphics[width=\textwidth]{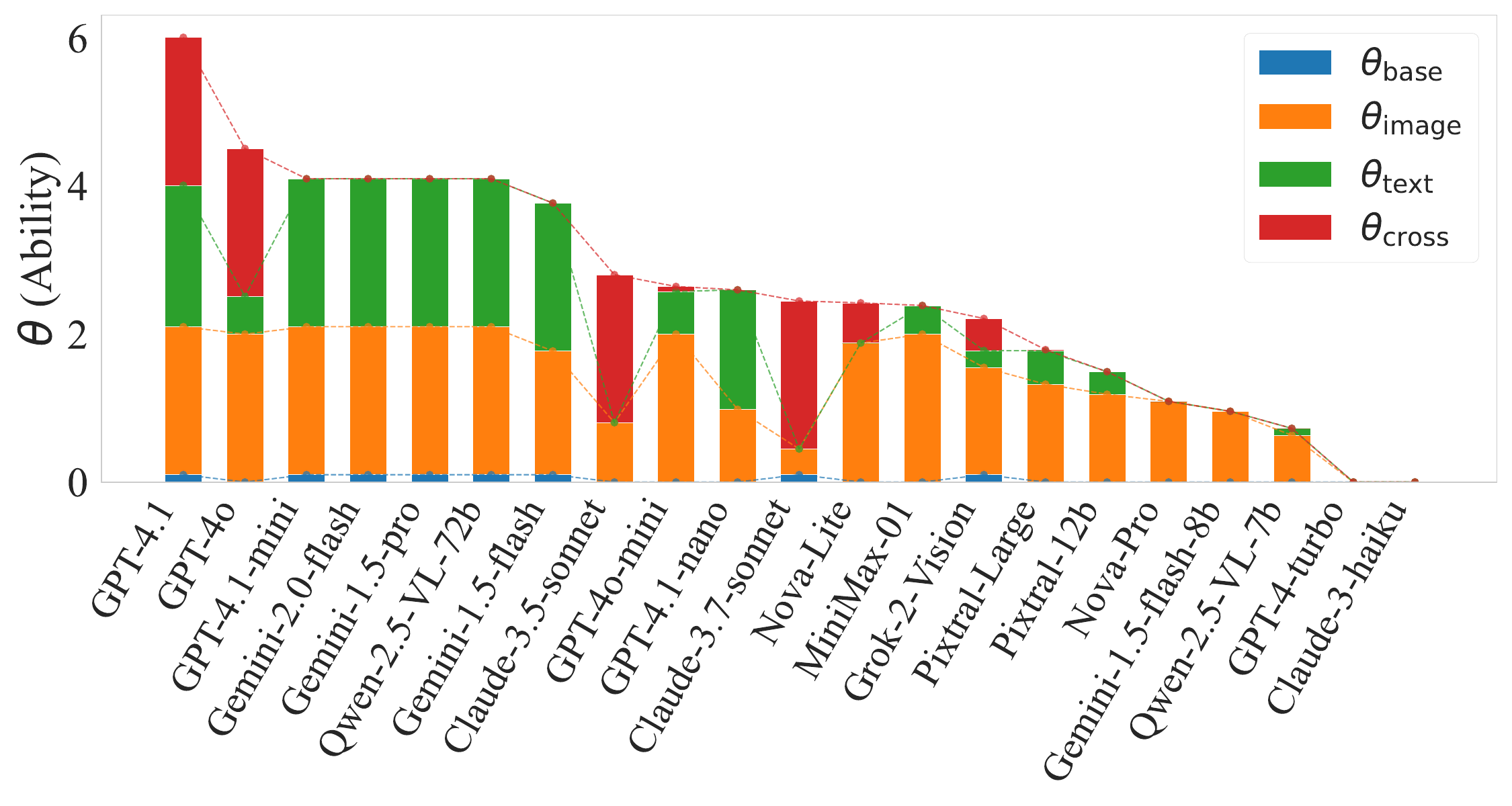}
        \caption{\textsc{SEED-Bench}}
        \label{fig:theta-inappropriate-seed}
    \end{subfigure}
    \caption{Distributions of $\theta$ estimated by \mmmirt{} sorted in descending order.}
    \label{fig:appendix-theta-seed}
\end{figure}

\Cref{fig:appendix-theta-onedim} shows the decomposed reasoning abilities of VLMs estimated by \mmirt{}.
\Cref{fig:theta-onedim-mmmu}, \cref{fig:theta-onedim-mathvista}, and \cref{fig:theta-onedim-seed} show the decomposed reasoning abilities of VLMs for \textsc{MMMU}, \textsc{mathVista}, and \textsc{SEED-Bench}.
\Cref{fig:appendix-theta-seed} shows the decomposed reasoning abilities of VLMs for \textsc{SEED-Bench}.
In \textsc{SEED-Bench}, most VLMs have high $\timage$.
This result corresponds to the fact that \textsc{SEED-Bench} contains problems which require images strongly to solve.

\subsection{Questions with High or Low Cross-modal Difficulty}
\begin{figure}[H]
    \centering
    \begin{subfigure}[b]{0.32\textwidth}
        \centering
        \includegraphics[width=\textwidth]{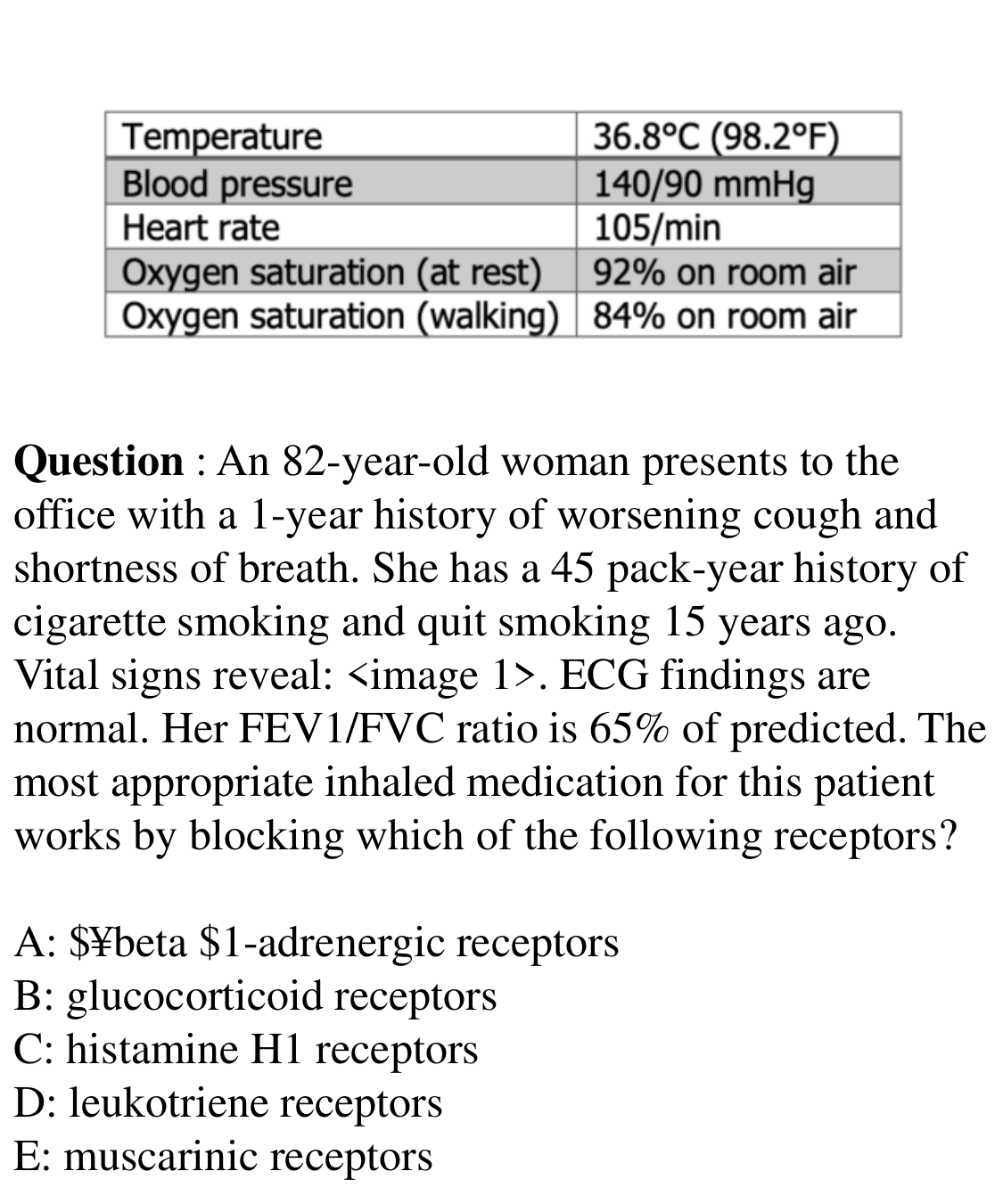}
        \caption{validation Pharmacy 2}
        \label{fig:high-synergy-mmmu1}
    \end{subfigure}
    \begin{subfigure}[b]{0.32\textwidth}
        \centering
        \includegraphics[width=\textwidth]{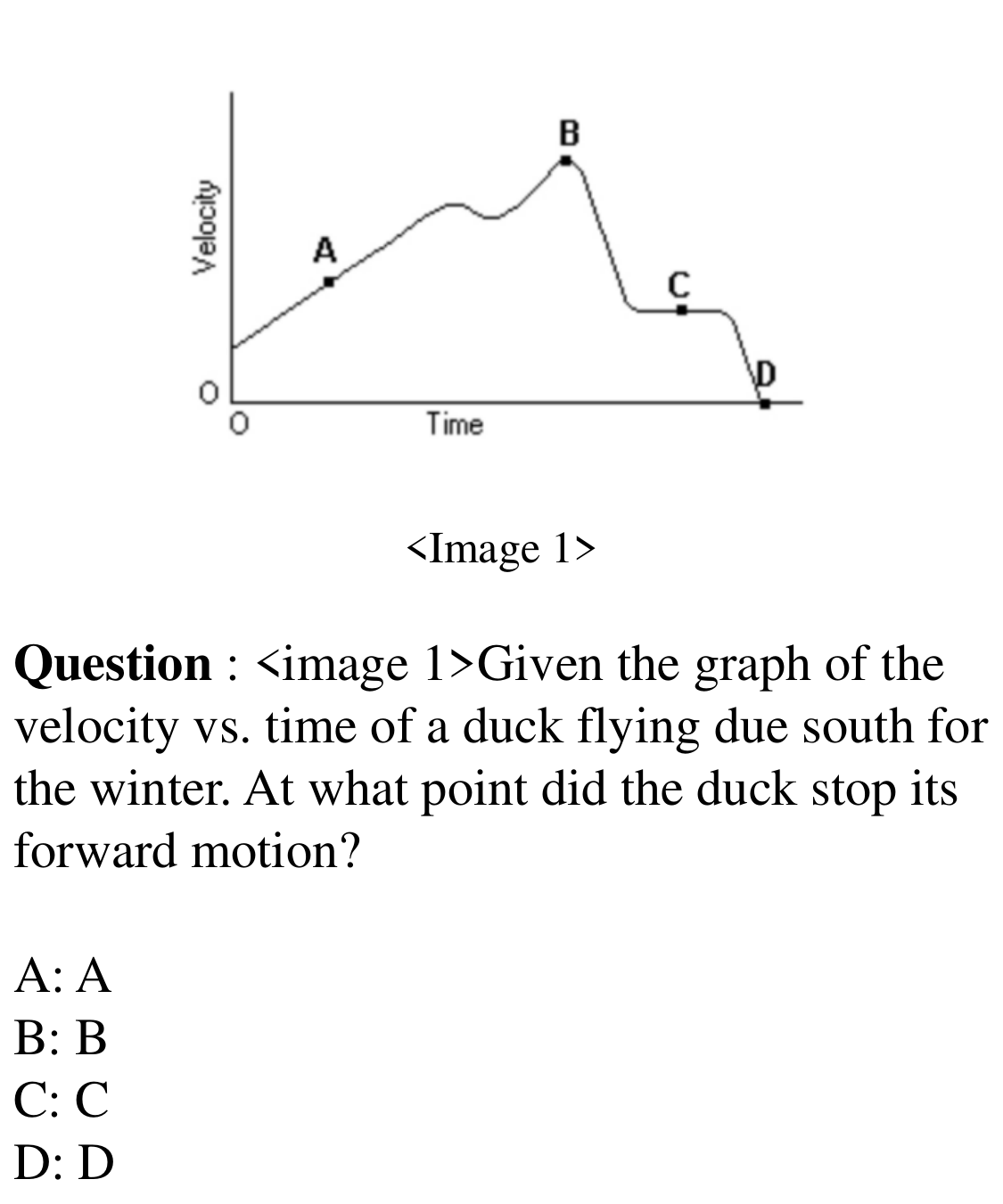}
        \caption{validation Physics 26}
        \label{fig:high-synergy-mmmu2}
    \end{subfigure}
    \begin{subfigure}[b]{0.32\textwidth}
        \centering
        \includegraphics[width=\textwidth]{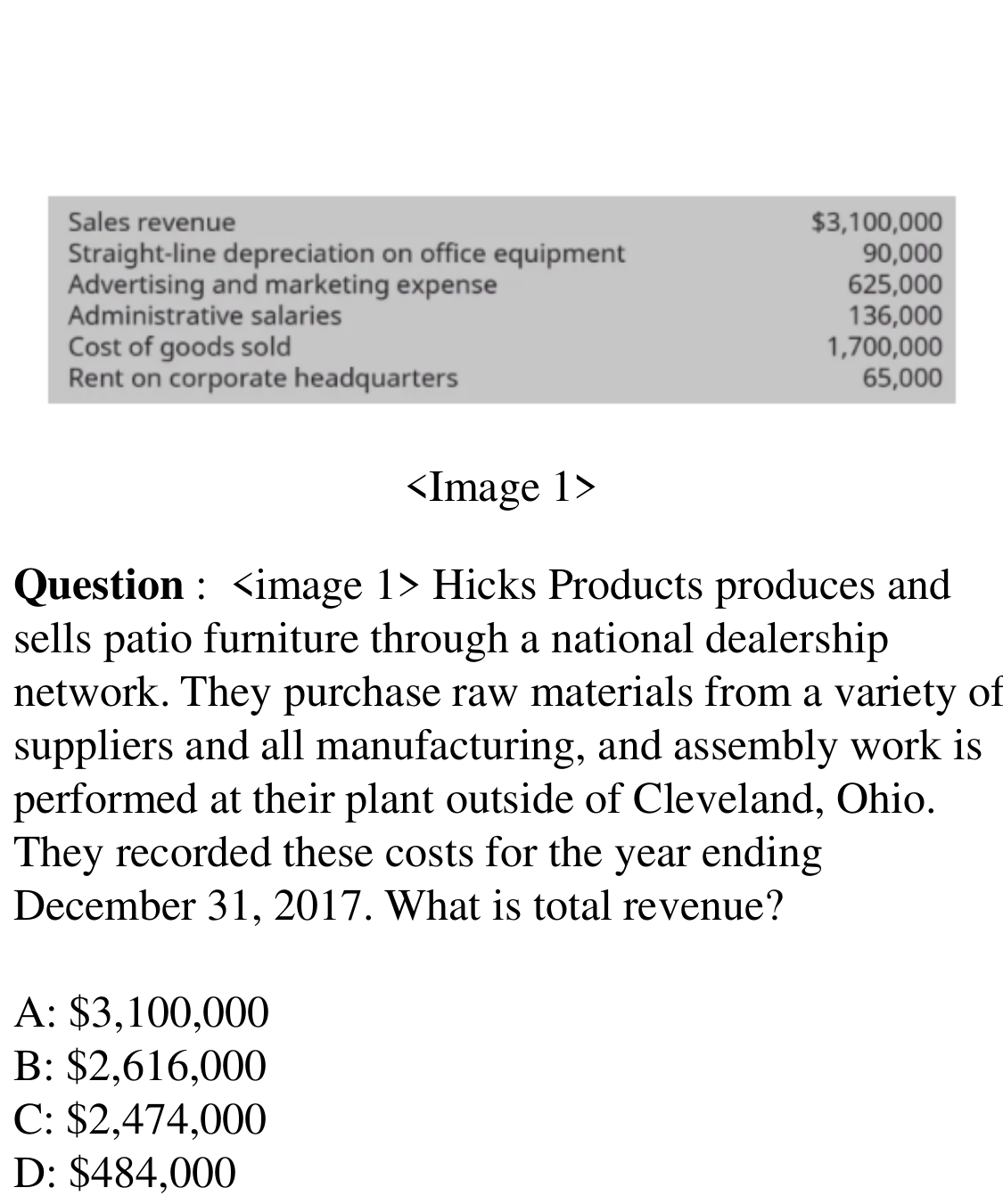}
        \caption{validation Accounting 12}
        \label{fig:high-synergy-mmmu3}
    \end{subfigure}
    \caption{\textsc{MMMU}: Questions with the high cross-modal difficulties $\bsynergy$}
    \label{fig:high-synergy-mmmu}
\end{figure}

\begin{figure}[H]
    \centering
    \begin{subfigure}[b]{0.32\textwidth}
        \centering
        \includegraphics[width=\textwidth]{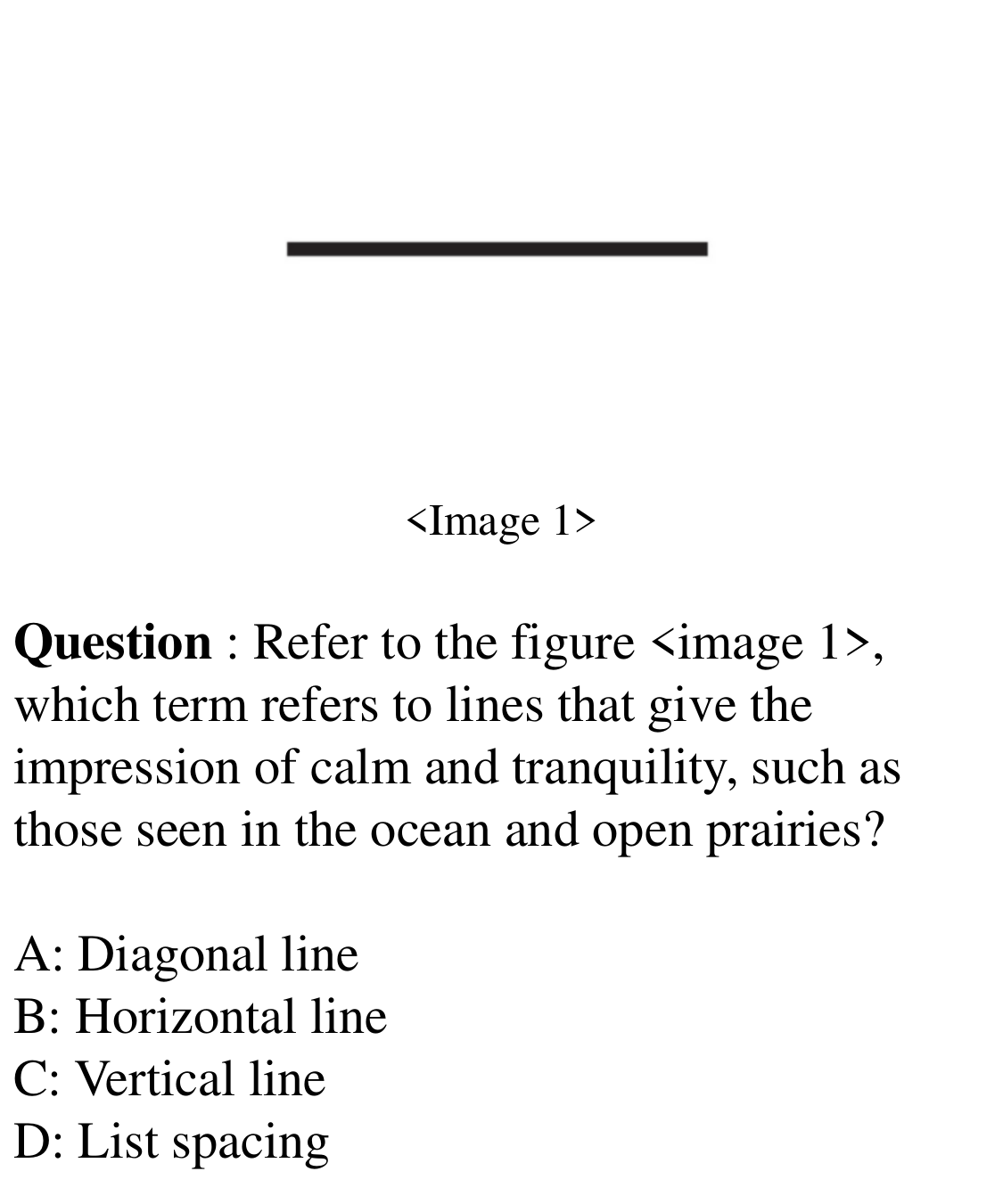}
        \caption{validation Literature 6}
        \label{fig:low-synergy-mmmu2}
    \end{subfigure}
    \begin{subfigure}[b]{0.32\textwidth}
        \centering
        \includegraphics[width=\textwidth]{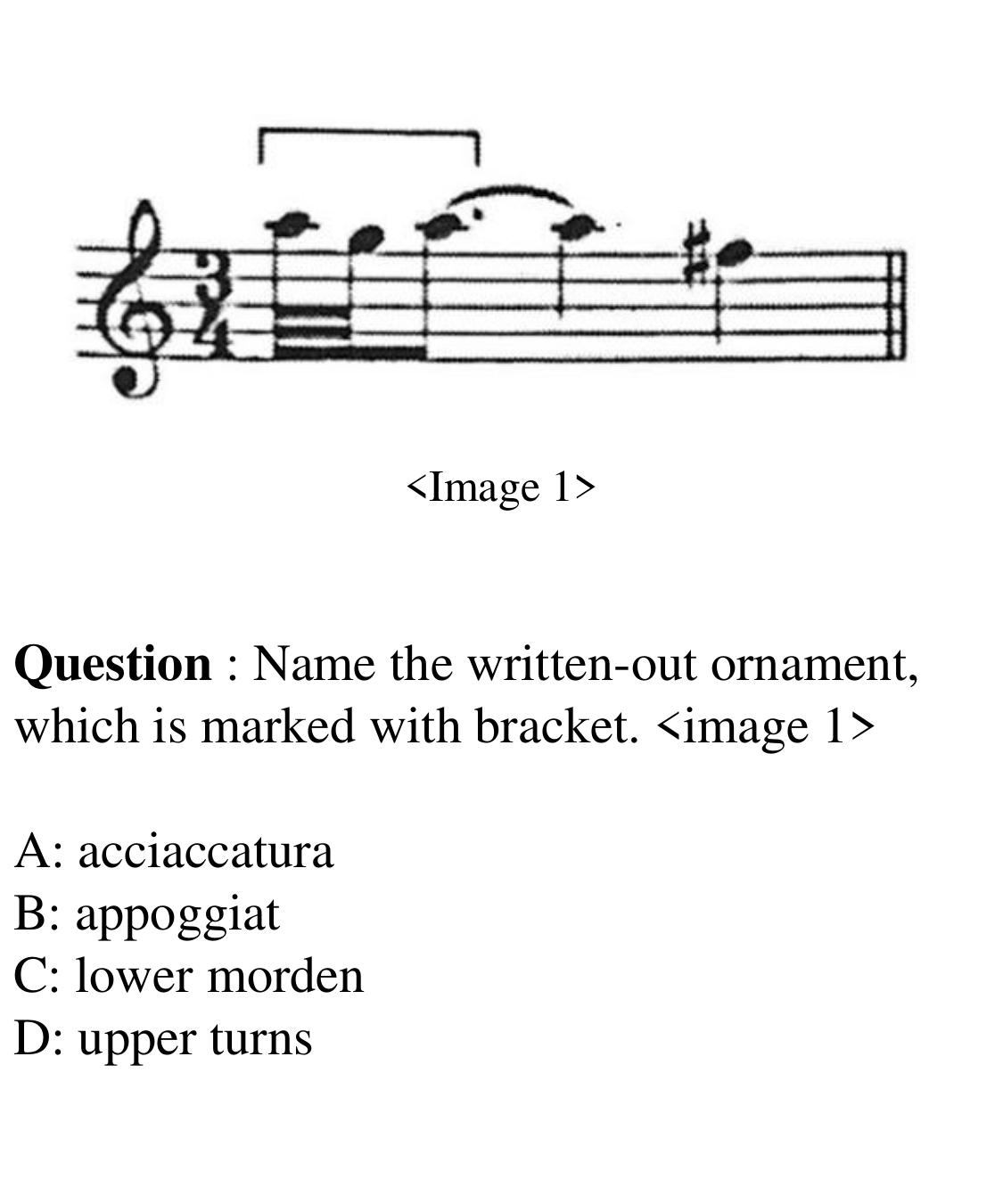}
        \caption{validation Music 11}
        \label{fig:low-synergy-mmmu3}
    \end{subfigure}
    \begin{subfigure}[b]{0.32\textwidth}
        \centering
        \includegraphics[width=\textwidth]{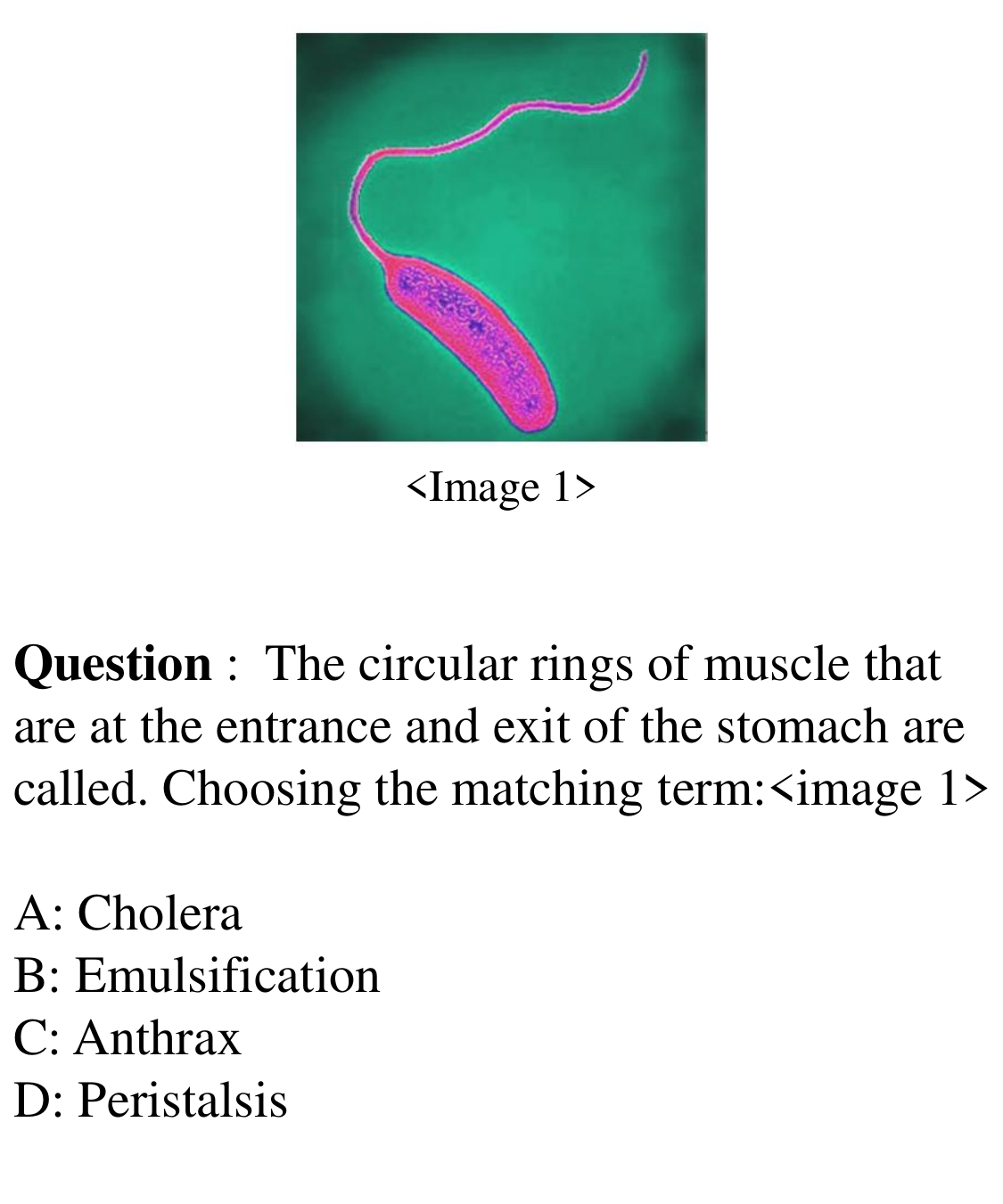}
        \caption{validation Agriculture 12}
        \label{fig:low-synergy-mmmu4}
    \end{subfigure}
    \caption{\textsc{MMMU}: Questions with the low cross-modal difficulties $\bsynergy$.}
    \label{fig:low-synergy-mmmu}
\end{figure}

\begin{figure}[H]
    \centering
    \begin{subfigure}[b]{0.32\textwidth}
        \centering
        \includegraphics[width=\textwidth]{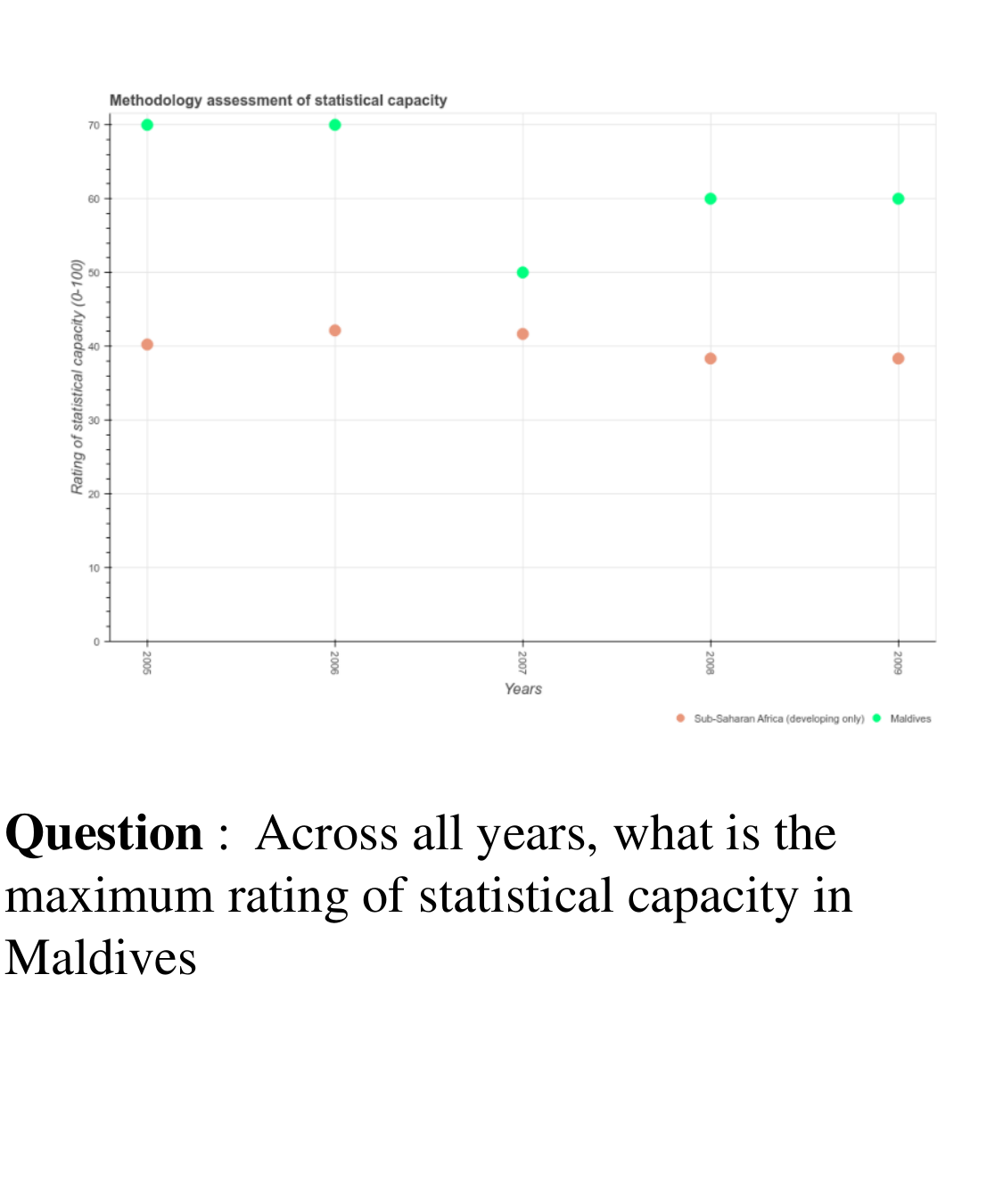}
        \caption{784}
        \label{fig:high-synergy-mathvista1}
    \end{subfigure}
    \begin{subfigure}[b]{0.32\textwidth}
        \centering
        \includegraphics[width=\textwidth]{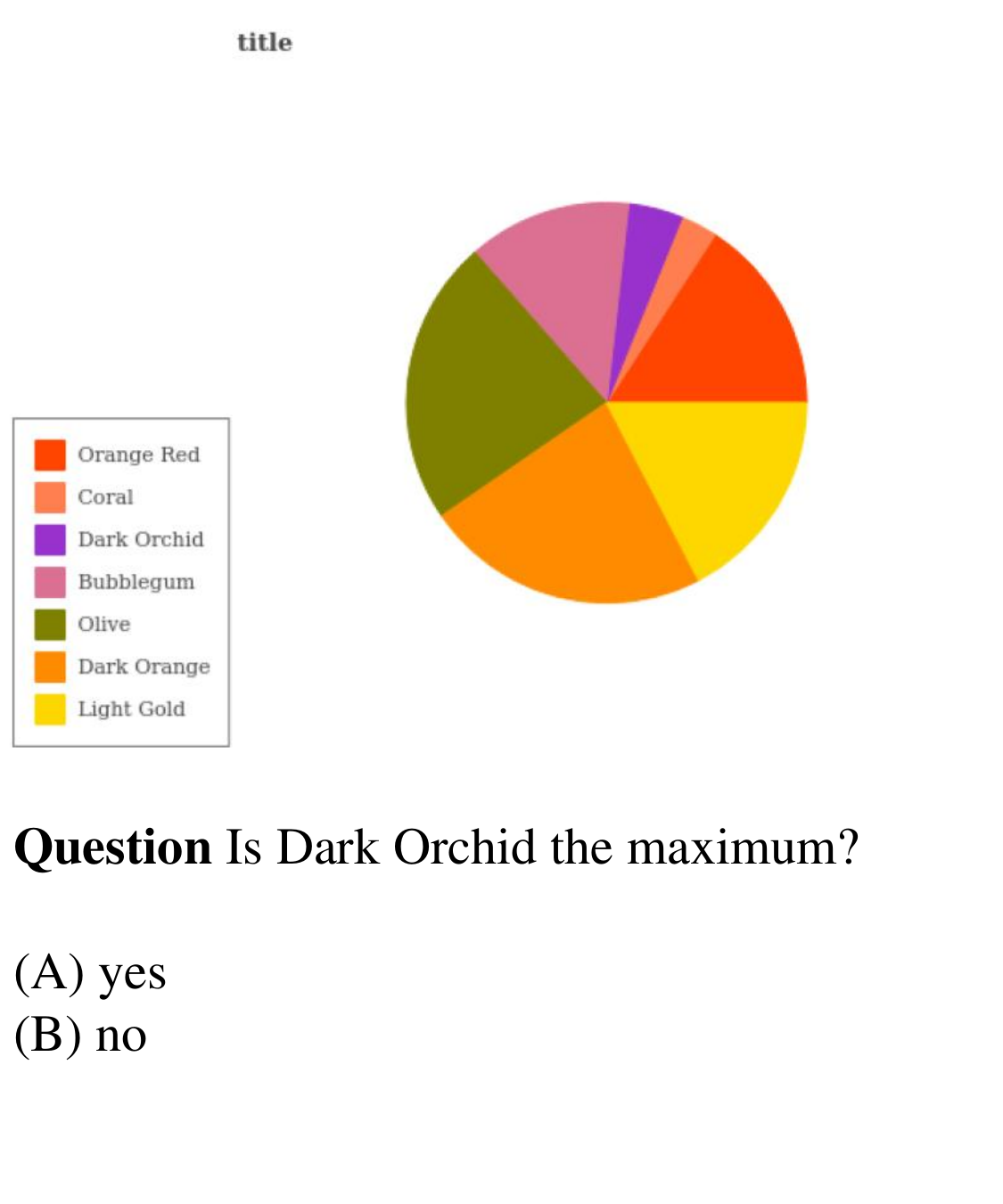}
        \caption{618}
        \label{fig:high-synergy-mathvista2}
    \end{subfigure}
    \begin{subfigure}[b]{0.32\textwidth}
        \centering
        \includegraphics[width=\textwidth]{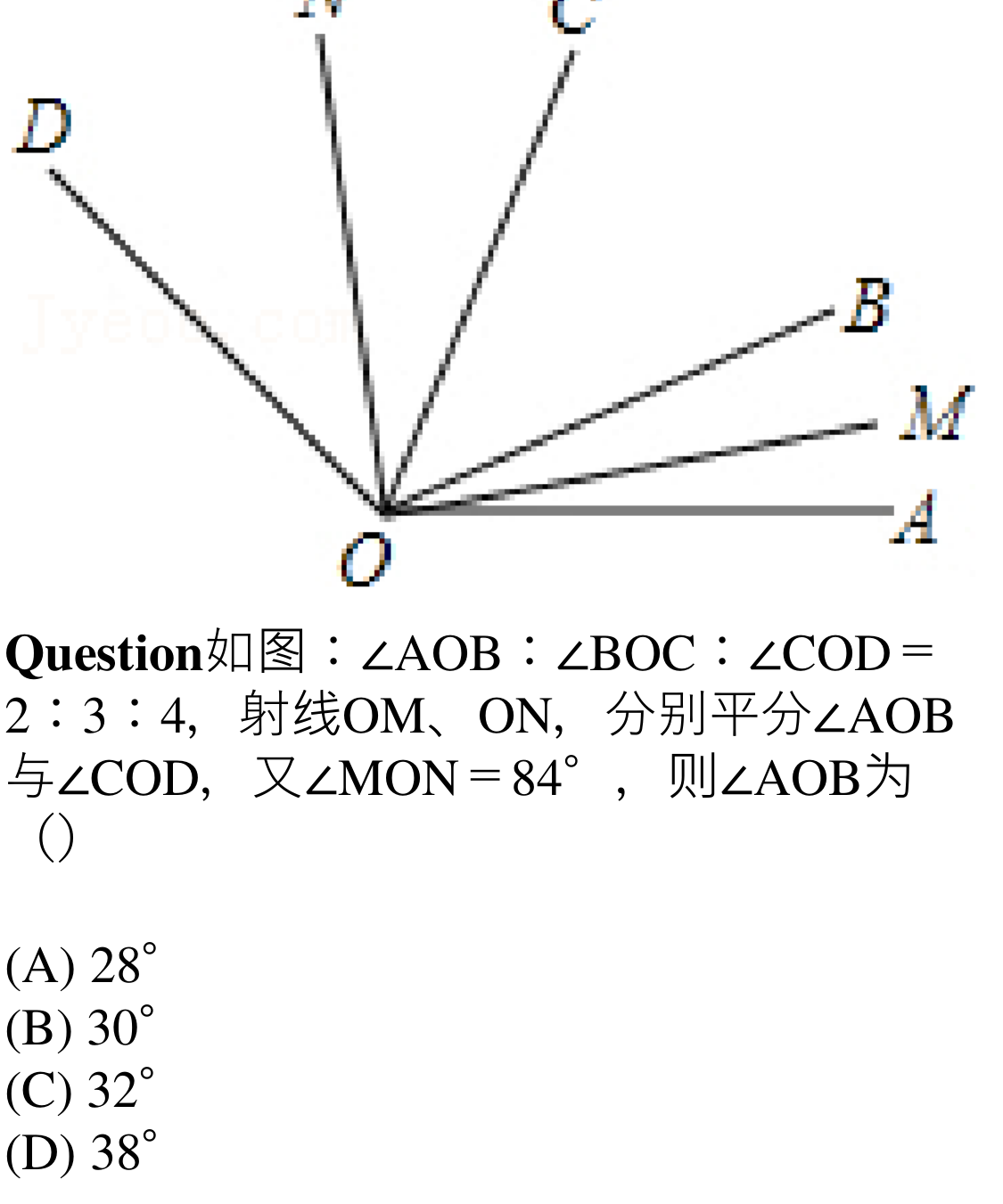}
        \caption{998}
        \label{fig:high-synergy-mathvista3}
    \end{subfigure}
    \caption{\textsc{MathVista}: Questions with the high cross-modal difficulties $\bsynergy$. Each caption is the ID of the problem in \textsc{MathVista}}
    \label{fig:high-synergy-mathvista}
\end{figure}

\begin{figure}[H]
    \centering
    \begin{subfigure}[b]{0.32\textwidth}
        \centering
        \includegraphics[width=\textwidth]{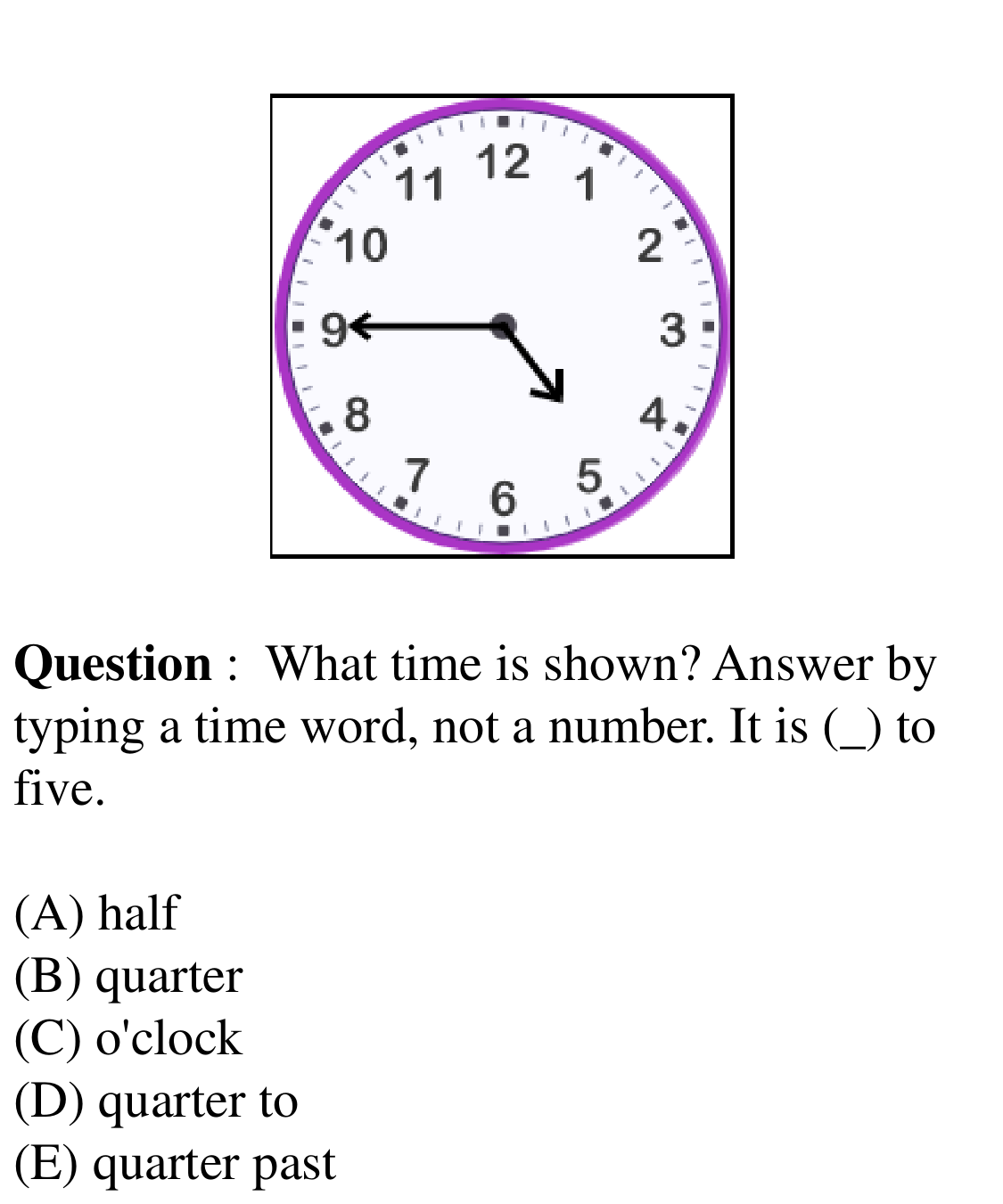}
        \caption{531}
        \label{fig:low-synergy-mathvista1}
    \end{subfigure}
    \begin{subfigure}[b]{0.32\textwidth}
        \centering
        \includegraphics[width=\textwidth]{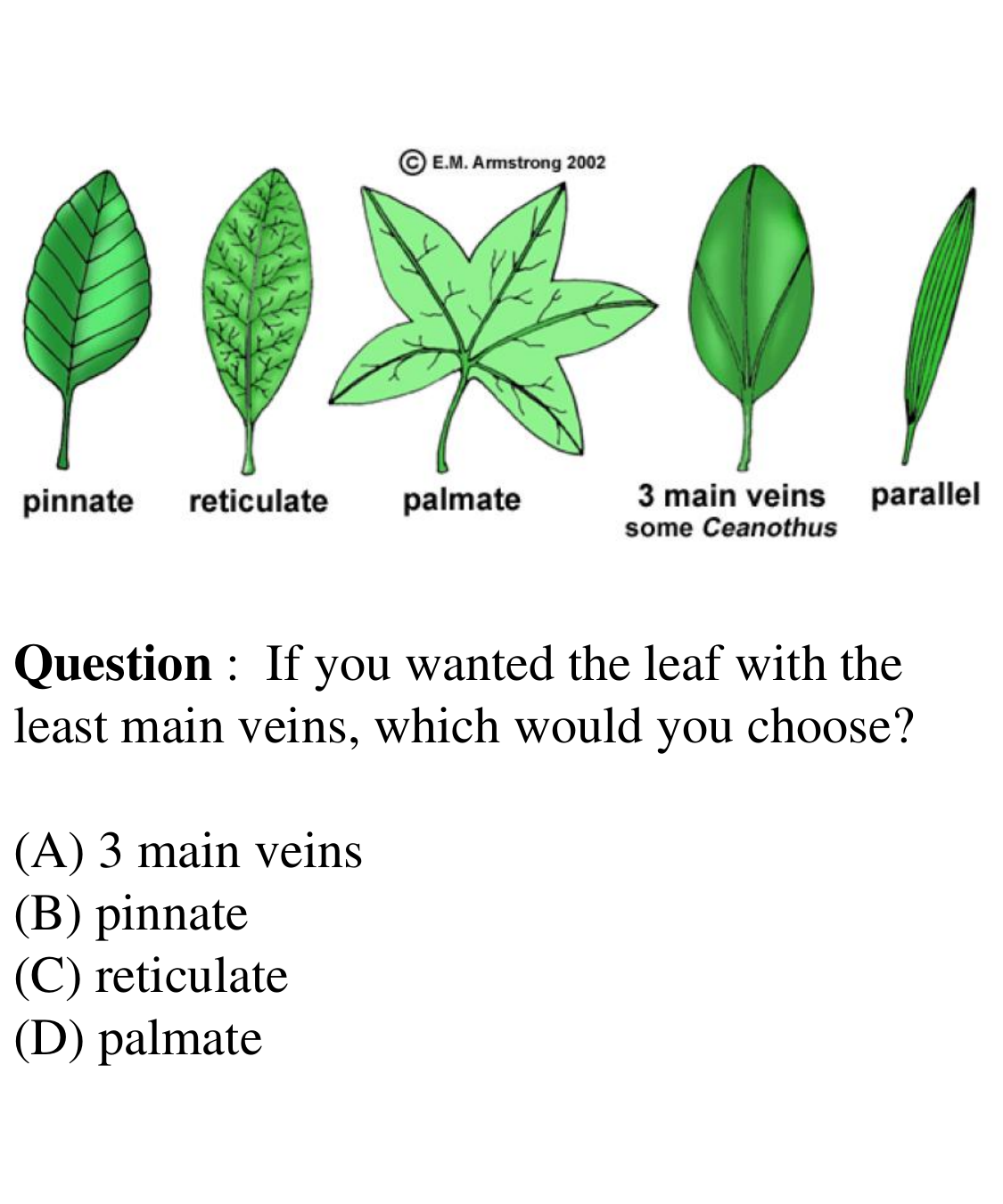}
        \caption{514}
        \label{fig:low-synergy-mathvista2}
    \end{subfigure}
    \begin{subfigure}[b]{0.32\textwidth}
        \centering
        \includegraphics[width=\textwidth]{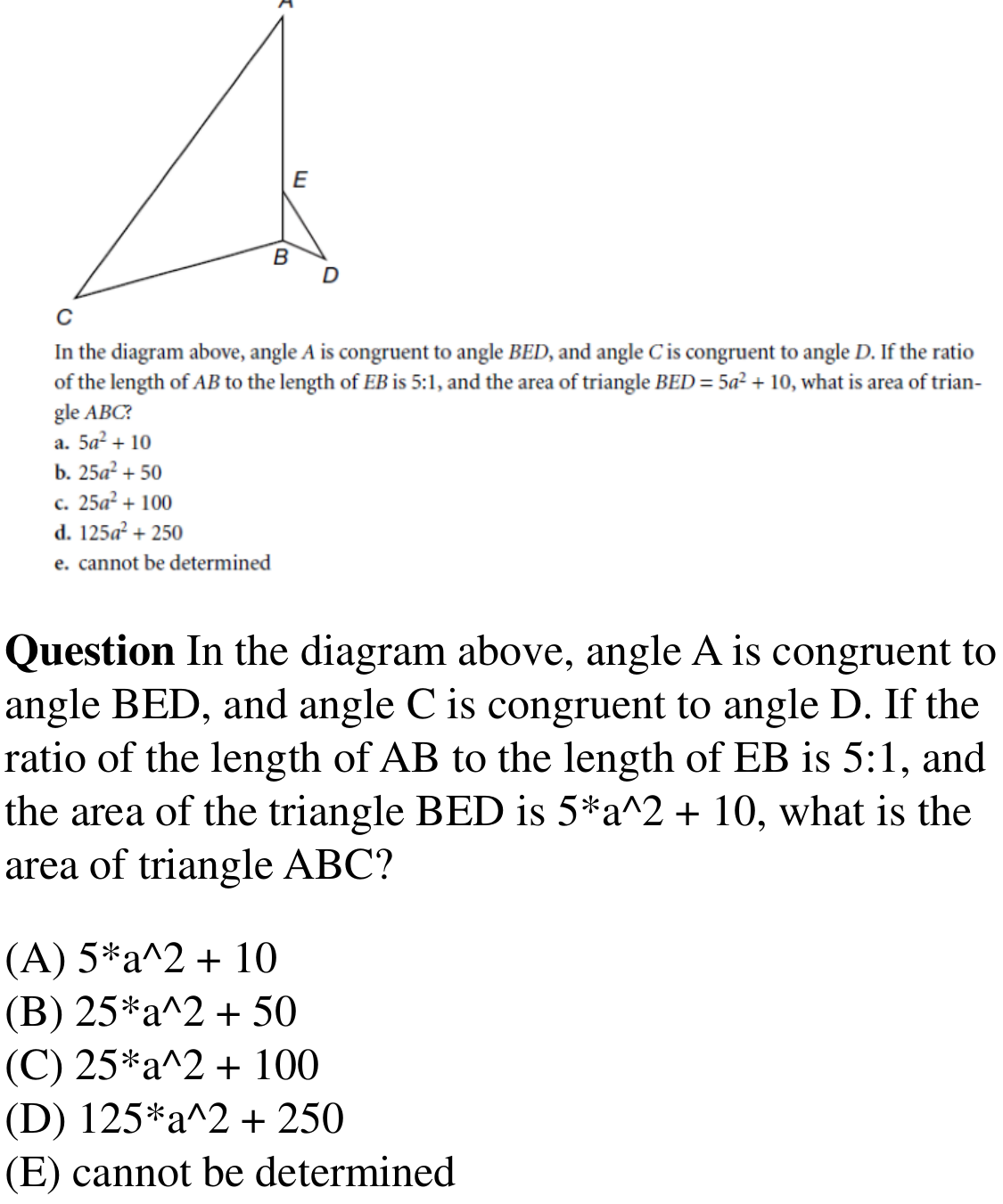}
        \caption{315}
        \label{fig:low-synergy-mathvista3}
    \end{subfigure}
    \caption{\textsc{MathVista}: Questions with the low cross-modal difficulties $\bsynergy$. Each caption is the ID of the problem in \textsc{MathVista}}
    \label{fig:low-synergy-mathvista}
\end{figure}

\begin{figure}[H]
    \centering
    \begin{subfigure}[b]{0.32\textwidth}
        \centering
        \includegraphics[width=\textwidth]{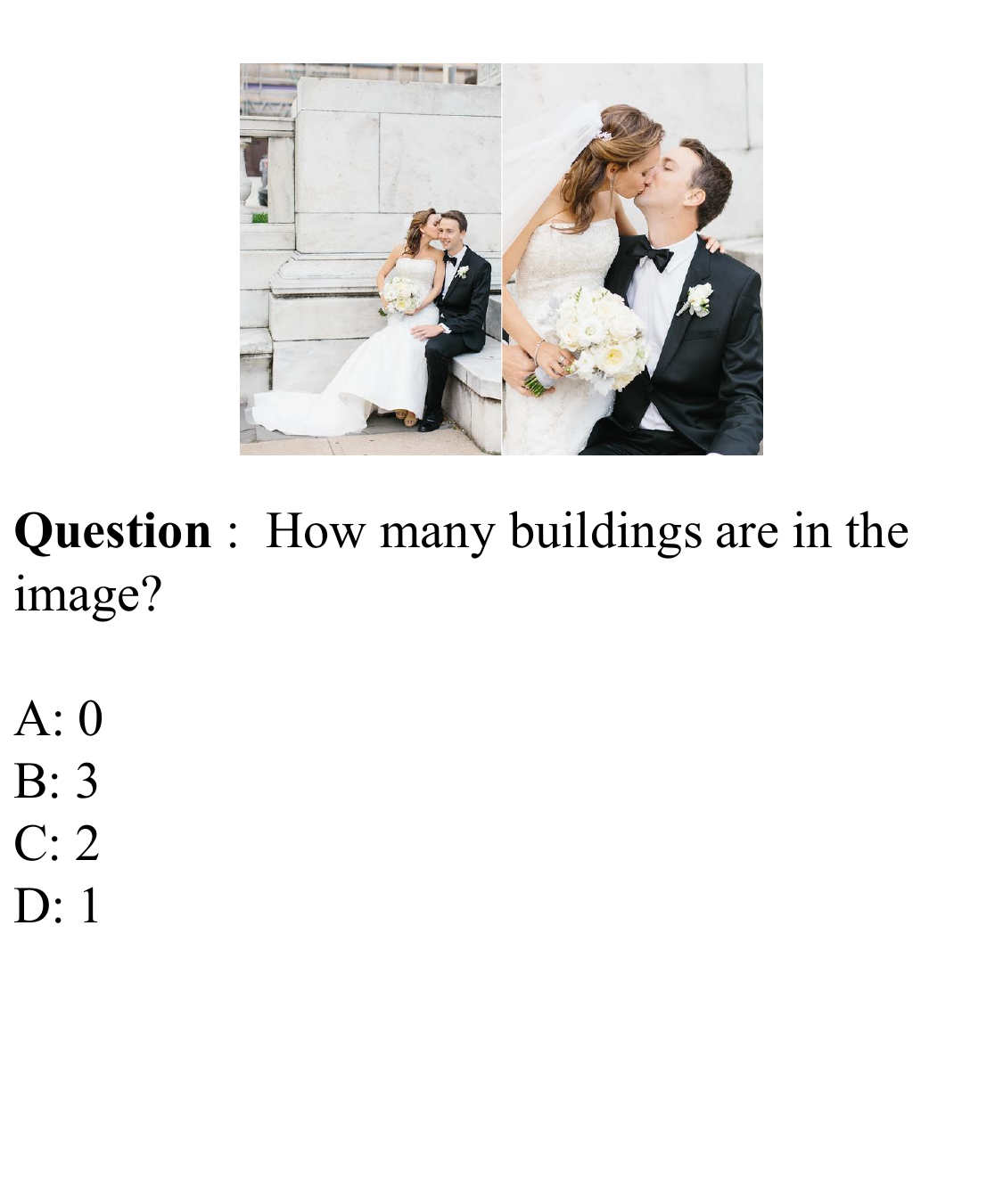}
        \caption{234357012}
        \label{fig:high-synergy-seed1}
    \end{subfigure}
    \begin{subfigure}[b]{0.32\textwidth}
        \centering
        \includegraphics[width=\textwidth]{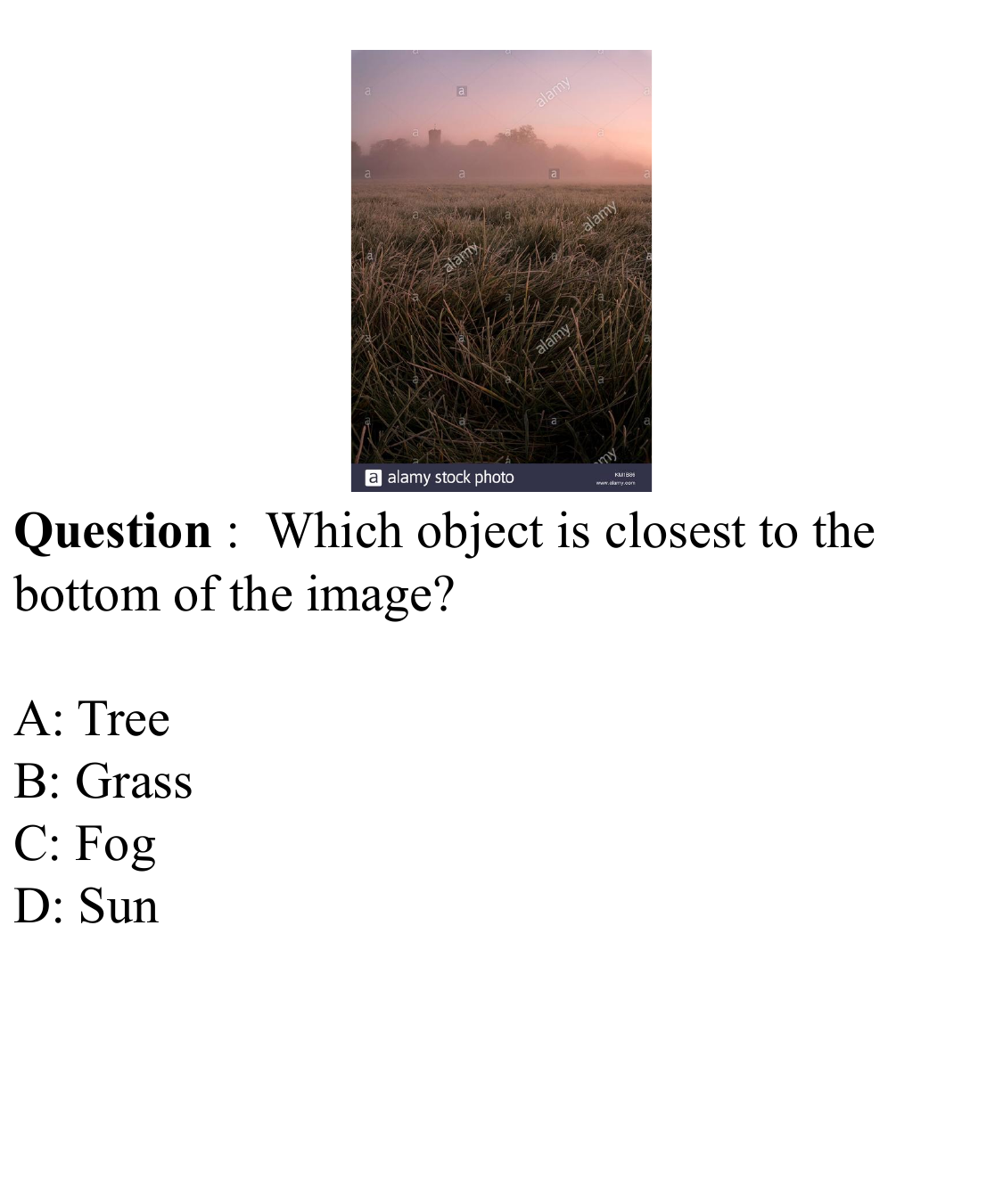}
        \caption{VizWiz train 00018501}
        \label{fig:high-synergy-seed2}
    \end{subfigure}
    \begin{subfigure}[b]{0.32\textwidth}
        \centering
        \includegraphics[width=\textwidth]{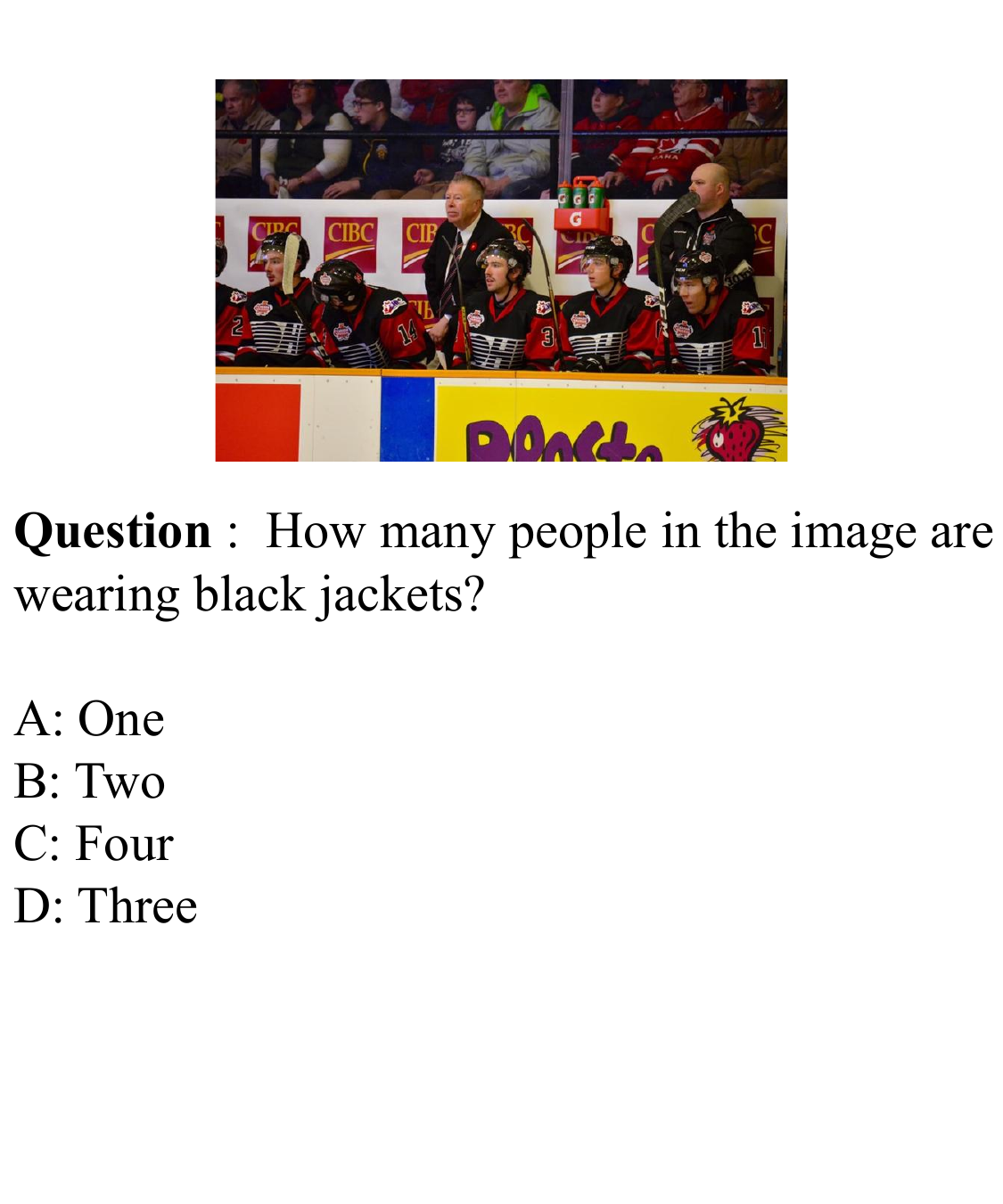}
        \caption{1907002}
        \label{fig:high-synergy-seed3}
    \end{subfigure}
    \caption{\textsc{SEED-Bench}: Questions with the high cross-modal difficulties $\bsynergy$.}
    \label{fig:high-synergy-seed}
\end{figure}

\begin{figure}[H]
    \centering
    \begin{subfigure}[b]{0.32\textwidth}
        \centering
        \includegraphics[width=\textwidth]{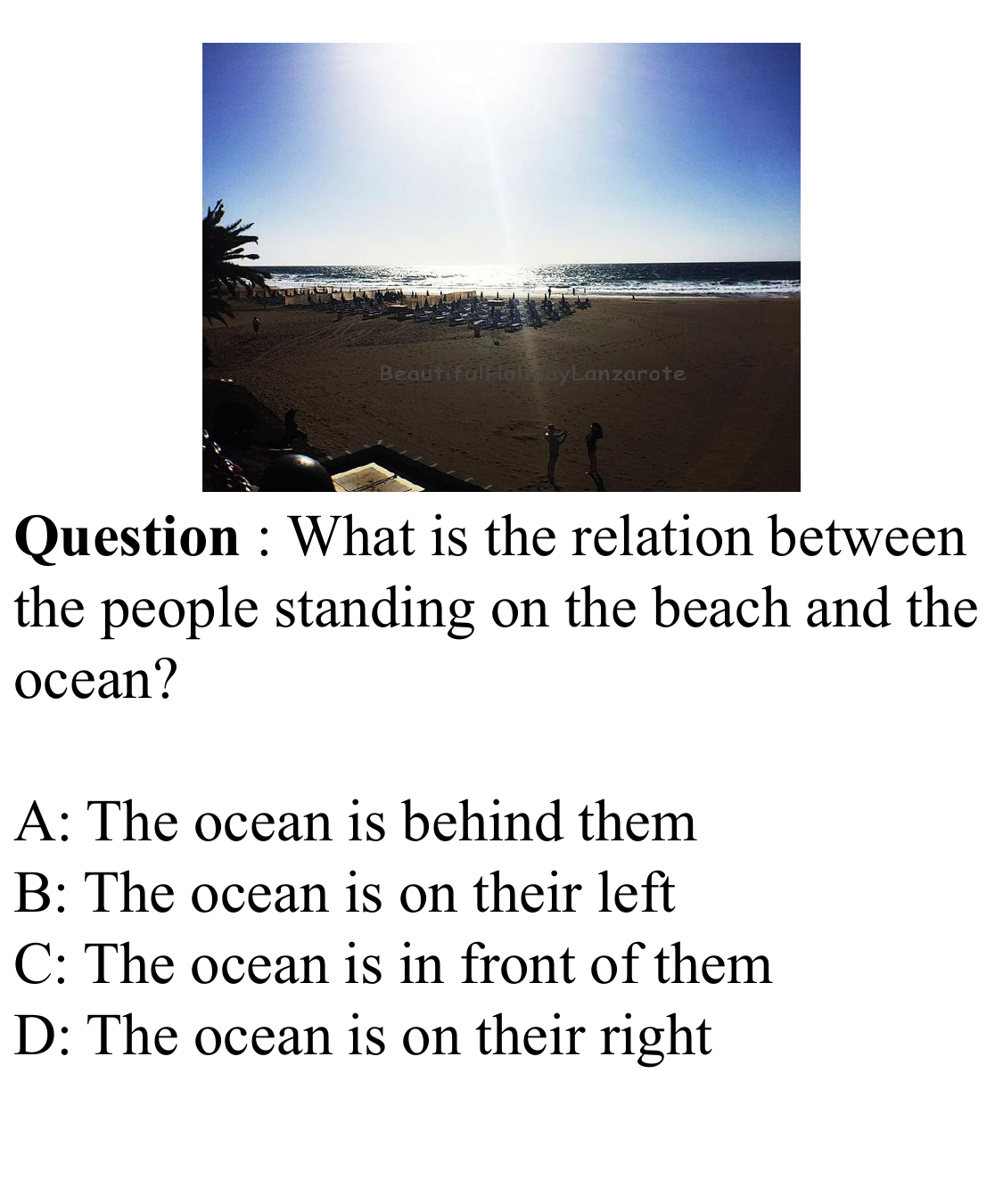}
        \caption{VizWiz train 00003082}
        \label{fig:low-synergy-seed1}
    \end{subfigure}
    \begin{subfigure}[b]{0.32\textwidth}
        \centering
        \includegraphics[width=\textwidth]{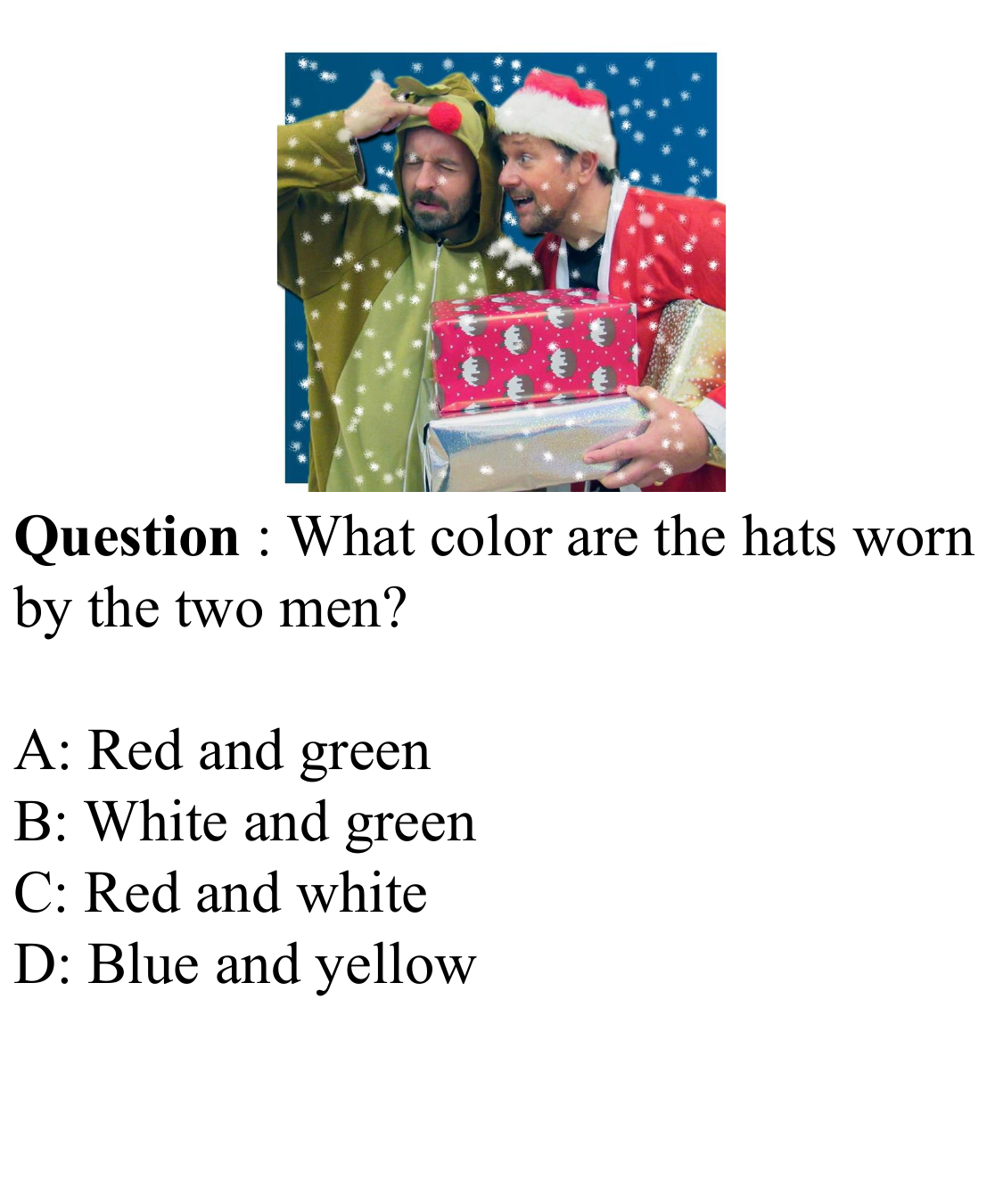}
        \caption{354261000}
        \label{fig:low-synergy-seed2}
    \end{subfigure}
    \begin{subfigure}[b]{0.32\textwidth}
        \centering
        \includegraphics[width=\textwidth]{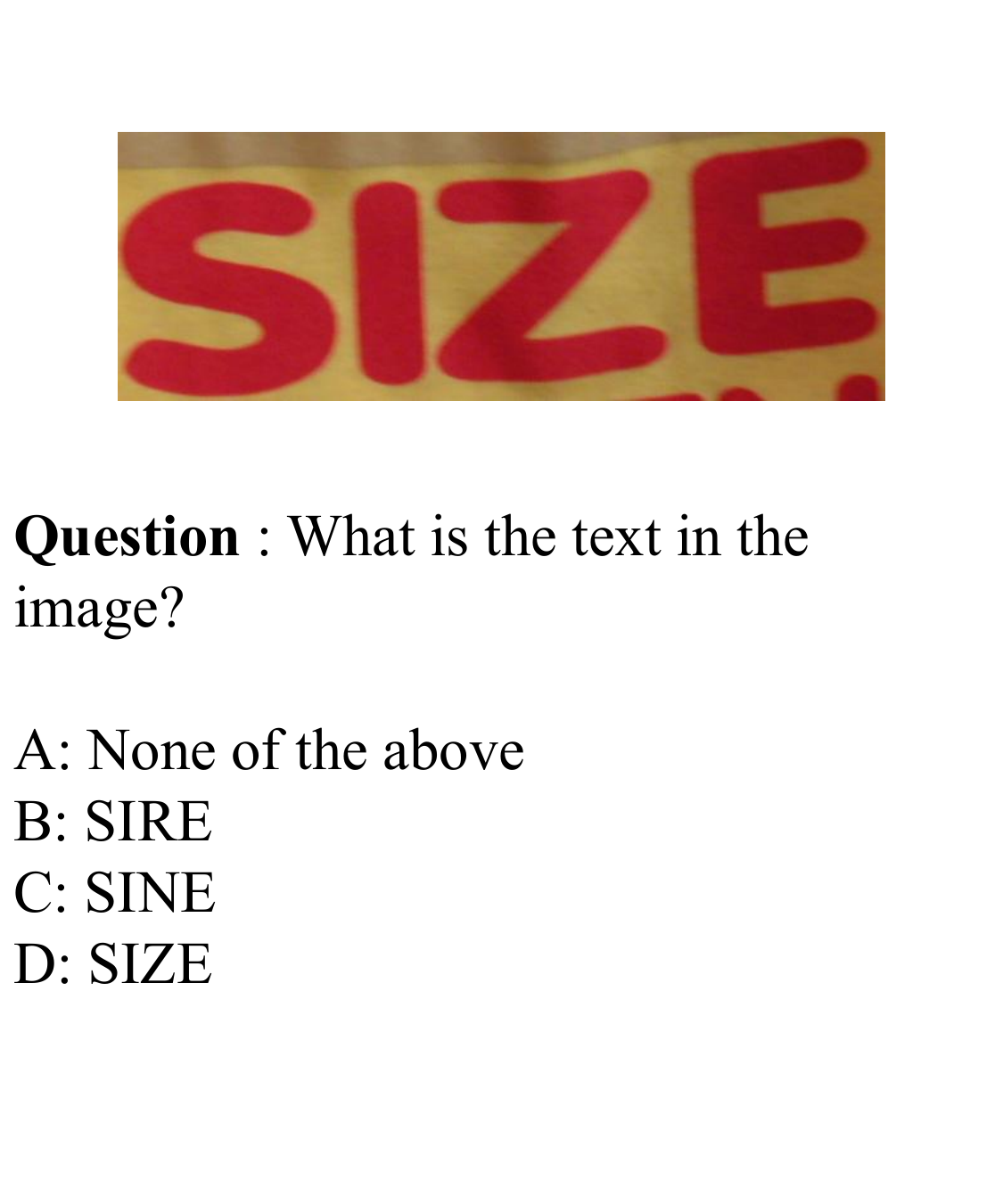}
        \caption{279787000}
        \label{fig:low-synergy-seed3}
    \end{subfigure}
    \caption{\textsc{VQAAT}: Questions with the low cross-modal difficulties $\bsynergy$}
    \label{fig:low-synergy-seed}
\end{figure}

We show the questions detected by \mmirt{} that require the high or low cross-modal reasoning ability from \textsc{MMMU} in \cref{fig:low-synergy-mmmu} and \cref{fig:high-synergy-mmmu}, from \textsc{MATHVISTA} in \cref{fig:low-synergy-mathvista} and \cref{fig:high-synergy-mathvista}, and from \textsc{SEED-Bench} in \cref{fig:low-synergy-seed} and \cref{fig:high-synergy-seed}, respectively.

As shown in \cref{fig:high-synergy-mmmu}, \cref{fig:high-synergy-mathvista}, and \cref{fig:high-synergy-seed}, questions which require the high cross-modal reasoning ability, whereas questions in \cref{fig:low-synergy-mmmu}, \cref{fig:low-synergy-mathvista}, and \cref{fig:low-synergy-seed} can be solved by using a single-modality only.
For example, the question shown in \cref{fig:low-synergy-mmmu4} presents an image of cholera bacteria, where the correct answer (A) can be identified solely from the image and answer choices, even without the text.
The question shown in \cref{fig:low-synergy-mathvista2} can be solved correctly without the image if one knows the number of veins for each plant.
For the question shown in \cref{fig:low-synergy-seed3} , since the question in \cref{fig:low-synergy-seed3}, this problem can be solved correctly simply by answering the characters shown in the image.
On the other hand, the question shown in \cref{fig:high-synergy-mmmu2} requires both the image, which provides velocity information, and the text, which specifies the particular conditions to identify within the figure. Consequently, the problem cannot be solved correctly if either the image or text component is missing.
The question shown in \cref{fig:high-synergy-mathvista3}.
The question in \cref{fig:high-synergy-seed3}, requires both the image, which provides there is one person who wears black clothes, and the text, which specifies specifying what to count within the figure. 
Consequently, the problem cannot be solved correctly if either the image or text component is missing.
Thus, the $\bsynergy$ successfully distinguishes between questions suitable for evaluating cross-modal ability where essential information is distributed across both image and text, requiring an examination of both to obtain the necessary information and those that do not effectively evaluate cross-modal ability.
\subsection{Detailed Result}
\Cref{tab:theta_mmmu}, \cref{tab:theta_mathvista}, and \cref{tab:theta_seedbench} show the values of $\theta$ predicted in \cref{fig:theta-inappropriate} and \cref{fig:appendix-theta-seed}.
\begin{table}[H]
\caption{Estimated $\theta$ on \textsc{MMMU}}
\label{tab:theta_mmmu}
\centering
\scalebox{0.9}{
\begin{tabular}{lrrrrr}
\toprule
model & $\theta_{\rm base}$ & $\theta_{\rm image}$ & $\theta_{\rm text}$ & $\theta_{\rm cross}$ & total \\
\midrule
Gemini-1.5-flash-8b & 0.00 & 1.2 & 0.00 & 4.0 & 5.2 \\
Claude-3.7-sonnet & 0.03 & 0.78 & 4.0 & 0.00 & 4.8 \\
Claude-3.5-sonnet & 0.06 & 0.68 & 4.0 & 0.00 & 4.7 \\
Gemini-2.0-flash & 0.09 & 1.4 & 0.68 & 2.1 & 4.3 \\
Gemini-1.5-pro & 0.10 & 1.0 & 0.29 & 2.8 & 4.2 \\
GPT-4.1-mini & 0.09 & 1.8 & 0.70 & 1.4 & 3.9 \\
Pixtrl-large & 0.03 & 0.90 & 0.37 & 2.5 & 3.8 \\
GPT-4.1 & 0.10 & 1.4 & 0.57 & 1.7 & 3.8 \\
GPT-4o & 0.10 & 0.66 & 0.71 & 2.2 & 3.7 \\
Gemini-1.5-flash & 0.06 & 0.75 & 0.38 & 2.4 & 3.6 \\
Qwen2.5-VL-72B & 0.07 & 1.2 & 0.51 & 1.7 & 3.5 \\
Pixtral-12b & 0.00 & 0.81 & 0.00 & 2.6 & 3.4 \\
Nova-Pro & 0.08 & 0.62 & 0.68 & 1.7 & 3.0 \\
Llama-3.2-90B & 0.09 & 0.08 & 0.20 & 2.7 & 3.0 \\
GPT-4o-mini & 0.07 & 0.43 & 0.41 & 2.0 & 3.0 \\
GPT-4-turbo & 0.10 & 0.54 & 0.94 & 1.1 & 2.7 \\
Grok-2-Vision & 0.10 & 0.38 & 1.0 & 1.0 & 2.5 \\
Llama-3.2-11B & 0.01 & 0.65 & 0.00 & 1.6 & 2.3 \\
Qwen2.5-VL-7B & 0.03 & 1.0 & 0.32 & 0.79 & 2.2 \\
Nova-Lite & 0.10 & 0.22 & 0.24 & 1.6 & 2.1 \\
GPT-4.1-nano & 0.10 & 0.13 & 0.19 & 1.4 & 1.9 \\
Claude-3-haiku & 0.10 & 0.00 & 0.63 & 0.09 & 0.82 \\
Claude-3-sonnet & 0.10 & 0.07 & 0.50 & 0.00 & 0.67 \\
MiniMax-01 & 0.00 & 0.00 & 0.00 & 0.00 & 0.00 \\
\bottomrule
\end{tabular}
}
\end{table}

\begin{table}
\caption{Estimated $\theta$ on \textsc{Mathvista}}
\label{tab:theta_mathvista}
\centering
\scalebox{0.9}{
\begin{tabular}{lrrrrr}
\toprule
model & $\theta_{\rm base}$ & $\theta_{\rm image}$ & $\theta_{\rm text}$ & $\theta_{\rm cross}$ & total \\
\midrule
Gemini-2.0-flash & 0.00 & 0.93 & 0.99 & 4.0 & 5.9 \\
Claude-3.7-sonnet & 0.00 & 0.63 & 2.0 & 3.1 & 5.7 \\
Nova-Pro & 0.10 & 0.36 & 4.0 & 0.56 & 5.0 \\
Grok-2-Vision & 0.10 & 0.89 & 3.4 & 0.57 & 5.0 \\
Gemini-1.5-pro & 0.00 & 0.97 & 1.4 & 2.6 & 4.9 \\
Gemini-1.5-flash & 0.00 & 0.60 & 0.84 & 2.7 & 4.1 \\
Nova-Lite & 0.10 & 0.00 & 4.0 & 0.00 & 4.1 \\
Qwen-2.5-VL-72b & 0.10 & 0.00 & 0.00 & 4.0 & 4.1 \\
Pixtral-Large & 0.10 & 0.70 & 2.4 & 0.25 & 3.4 \\
GPT-4o & 0.00 & 0.32 & 2.5 & 0.57 & 3.4 \\
GPT-4o-mini & 0.00 & 0.00 & 2.8 & 0.41 & 3.2 \\
Llama-3.2-90B & 0.10 & 0.00 & 3.0 & 0.00 & 3.1 \\
Claude-3.5-sonnet & 0.00 & 0.38 & 1.7 & 0.61 & 2.7 \\
GPT-4-turbo & 0.00 & 0.46 & 1.1 & 0.51 & 2.1 \\
Gemini-1.5-flash-8b & 0.00 & 0.36 & 0.60 & 0.78 & 1.7 \\
Pixtral-12b & 0.00 & 0.41 & 1.1 & 0.00 & 1.5 \\
GPT-4.1 & 0.10 & 0.30 & 0.00 & 1.1 & 1.5 \\
Qwen-2.5-VL-7b & 0.10 & 0.00 & 0.00 & 1.2 & 1.3 \\
GPT-4.1-mini & 0.10 & 0.00 & 0.00 & 1.2 & 1.3 \\
Claude-3-sonnet & 0.00 & 0.00 & 0.98 & 0.00 & 0.98 \\
MiniMax-01 & 0.00 & 0.64 & 0.02 & 0.31 & 0.96 \\
Claude-3-haiku & 0.00 & 0.00 & 0.71 & 0.00 & 0.71 \\
Llama-3.2-11B & 0.10 & 0.00 & 0.40 & 0.00 & 0.50 \\
GPT-4.1-nano & 0.10 & 0.00 & 0.00 & 0.15 & 0.25 \\
\bottomrule
\end{tabular}
}
\end{table}

\begin{table}
\caption{Estimated $\theta$ on \textsc{SeedBench}}
\label{tab:theta_seedbench}
\centering
\scalebox{0.9}{
\begin{tabular}{lrrrrr}
\toprule
model & $\theta_{\rm base}$ & $\theta_{\rm image}$ & $\theta_{\rm text}$ & $\theta_{\rm cross}$ & total \\
\midrule
GPT-4.1 & 0.10 & 2.0 & 1.9 & 2.0 & 6.0 \\
GPT-4o & 0.00 & 2.0 & 0.51 & 2.0 & 4.5 \\
GPT-4.1-mini & 0.10 & 2.0 & 2.0 & 0.00 & 4.1 \\
Gemini-2.0-flash & 0.10 & 2.0 & 2.0 & 0.00 & 4.1 \\
Gemini-1.5-pro & 0.10 & 2.0 & 2.0 & 0.00 & 4.1 \\
Qwen-2.5-VL-72b & 0.10 & 2.0 & 2.0 & 0.00 & 4.1 \\
Gemini-1.5-flash & 0.10 & 1.7 & 2.0 & 0.00 & 3.8 \\
Claude-3.5-sonnet & 0.00 & 0.80 & 0.00 & 2.0 & 2.8 \\
GPT-4o-mini & 0.00 & 2.0 & 0.58 & 0.07 & 2.6 \\
GPT-4.1-nano & 0.00 & 0.98 & 1.6 & 0.00 & 2.6 \\
Claude-3.7-sonnet & 0.10 & 0.35 & 0.00 & 2.0 & 2.4 \\
Nova-Lite & 0.00 & 1.9 & 0.00 & 0.54 & 2.4 \\
MiniMax-01 & 0.00 & 2.0 & 0.39 & 0.00 & 2.4 \\
Grok-2-Vision & 0.10 & 1.4 & 0.23 & 0.43 & 2.2 \\
Pixtral-Large & 0.00 & 1.3 & 0.45 & 0.02 & 1.8 \\
Pixtral-12b & 0.00 & 1.2 & 0.30 & 0.00 & 1.5 \\
Nova-Pro & 0.00 & 1.1 & 0.00 & 0.00 & 1.1 \\
Gemini-1.5-flash-8b & 0.00 & 0.96 & 0.00 & 0.00 & 0.96 \\
Qwen-2.5-VL-7b & 0.00 & 0.63 & 0.09 & 0.00 & 0.73 \\
GPT-4-turbo & 0.00 & 0.00 & 0.00 & 0.00 & 0.00 \\
Claude-3-haiku & 0.00 & 0.00 & 0.00 & 0.00 & 0.00 \\
\bottomrule
\end{tabular}
}
\end{table}
\section{Omitted Results of Multimodal Benchmark Refinement}
\label{appendix:benchmarkrefinement}
\begin{figure}[tb]
    \centering
    \includegraphics[width=0.8\linewidth]{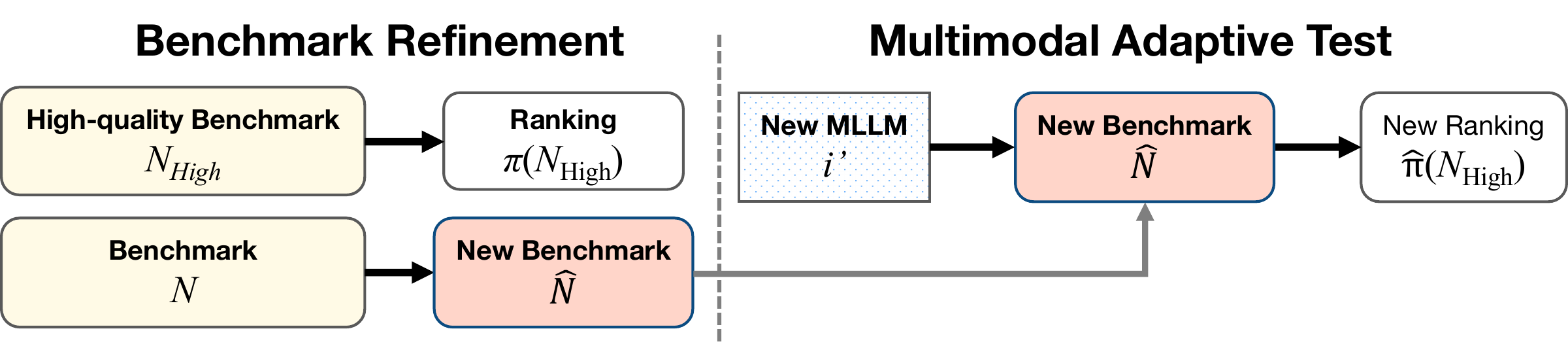}
    \caption{An illustration of our benchmark refinement and testing processes.}
    \label{fig:problem_seting}
\end{figure}
\begin{figure}[tb]
    \centering
    \begin{subfigure}[b]{0.32\textwidth}
    \includegraphics[width=\linewidth]{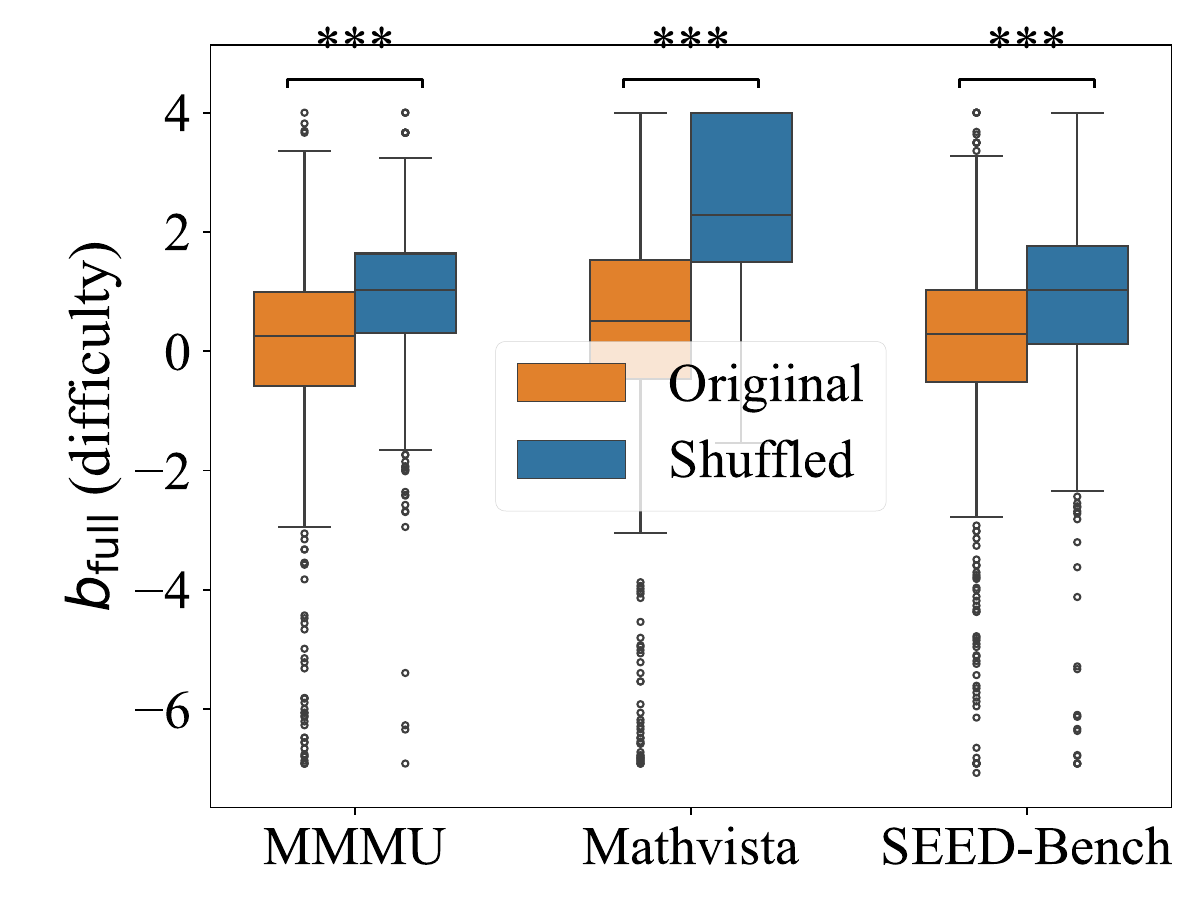}
    \caption{Difficulty}
    \label{fig:difficulty-mmirt}
    \end{subfigure}
    \begin{subfigure}[b]{0.32\textwidth}
    \includegraphics[width=\linewidth]{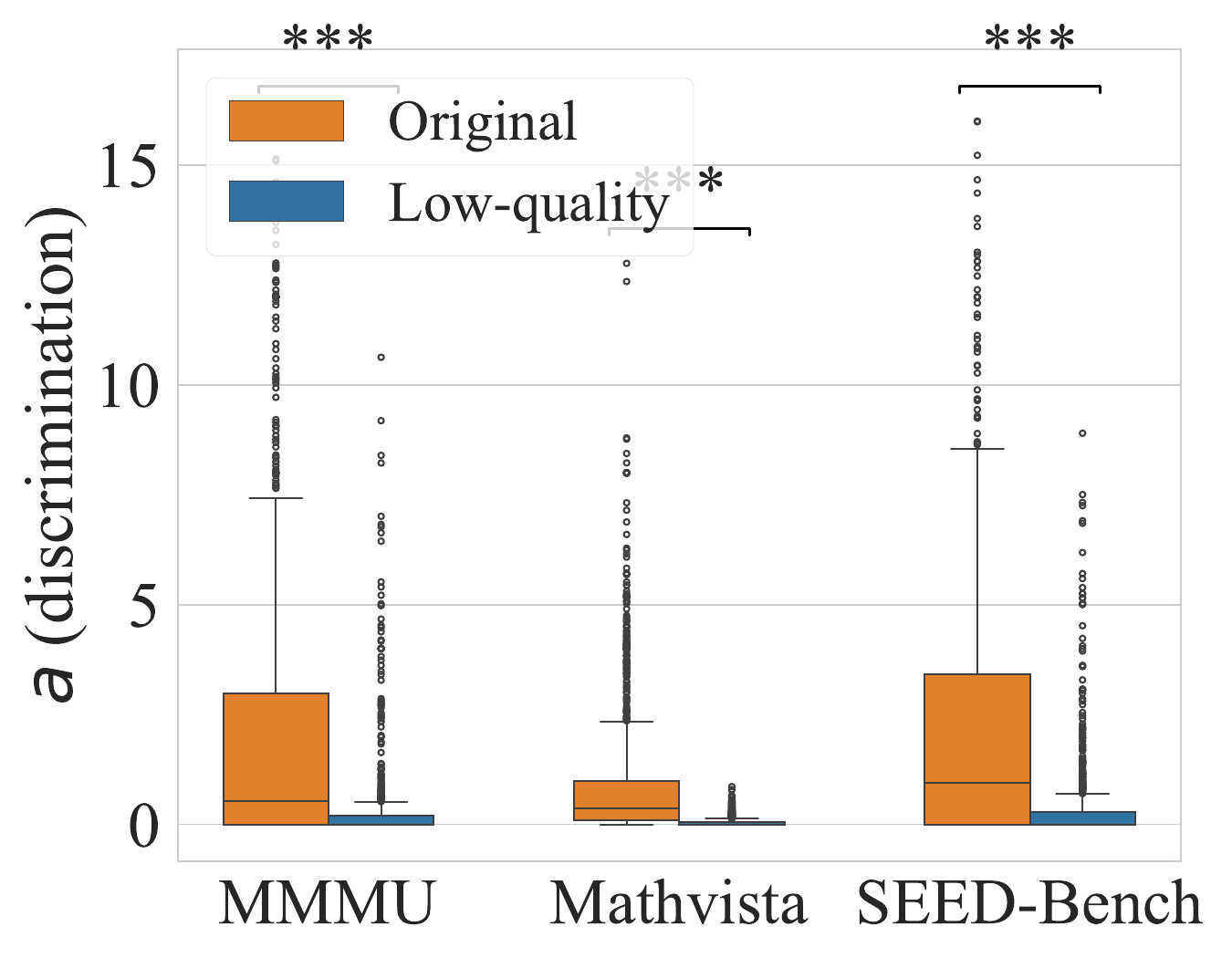}
    \caption{Discrimination}
    \label{fig:discrimination-mmirt}
    \end{subfigure}
    \begin{subfigure}[b]{0.32\textwidth}
    \includegraphics[width=\linewidth]{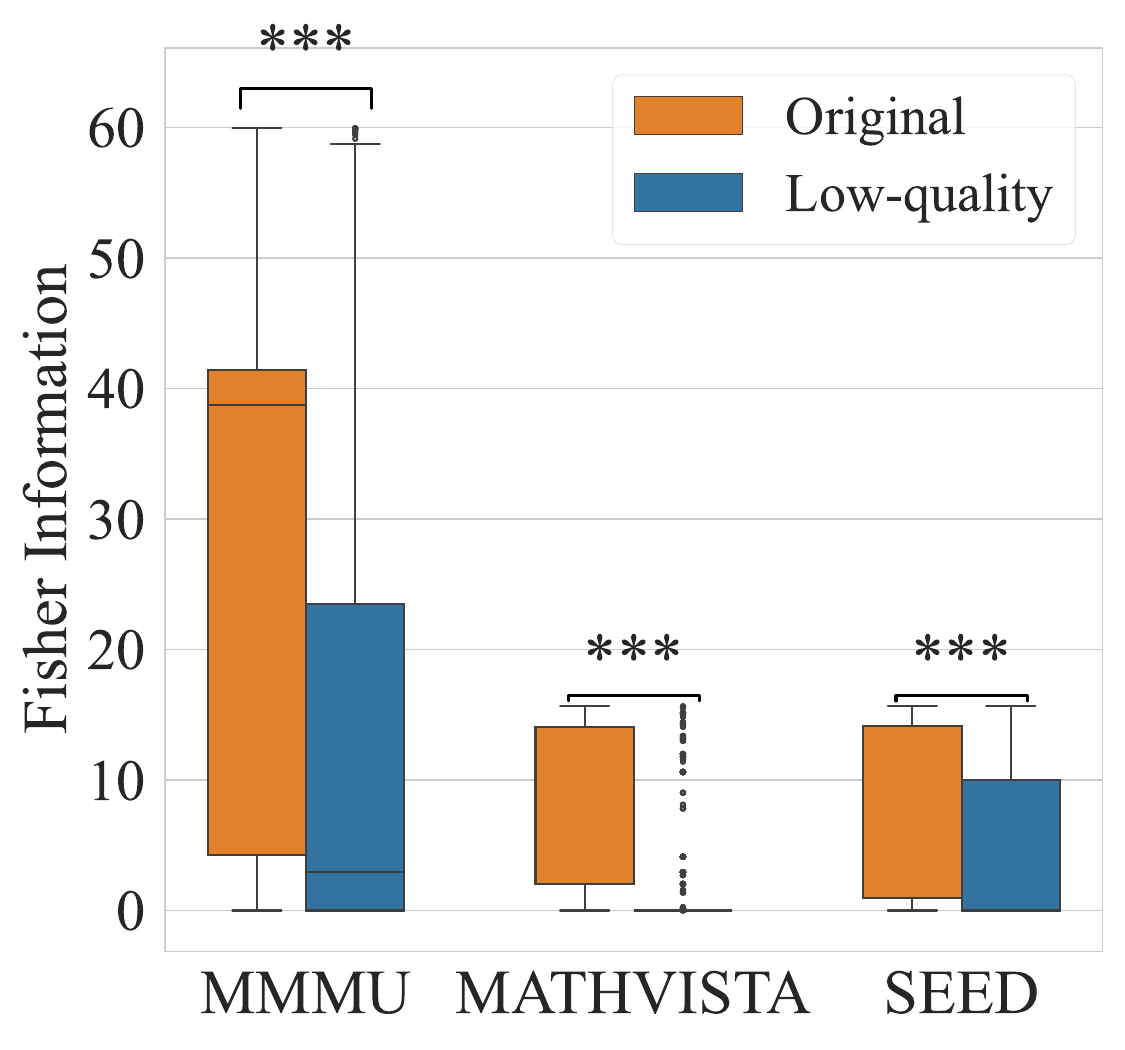}
    \caption{Fisher information}
    \label{fig:fisher-mmirt}
    \end{subfigure}
    \caption{Comparisons of parameters estimated by \mmirt{} between the original and artificial questions.}
    \label{fig:fisher}
\end{figure}

\begin{figure}[tb]
    \centering
    \begin{subfigure}[b]{0.32\textwidth}
    \includegraphics[width=\linewidth]{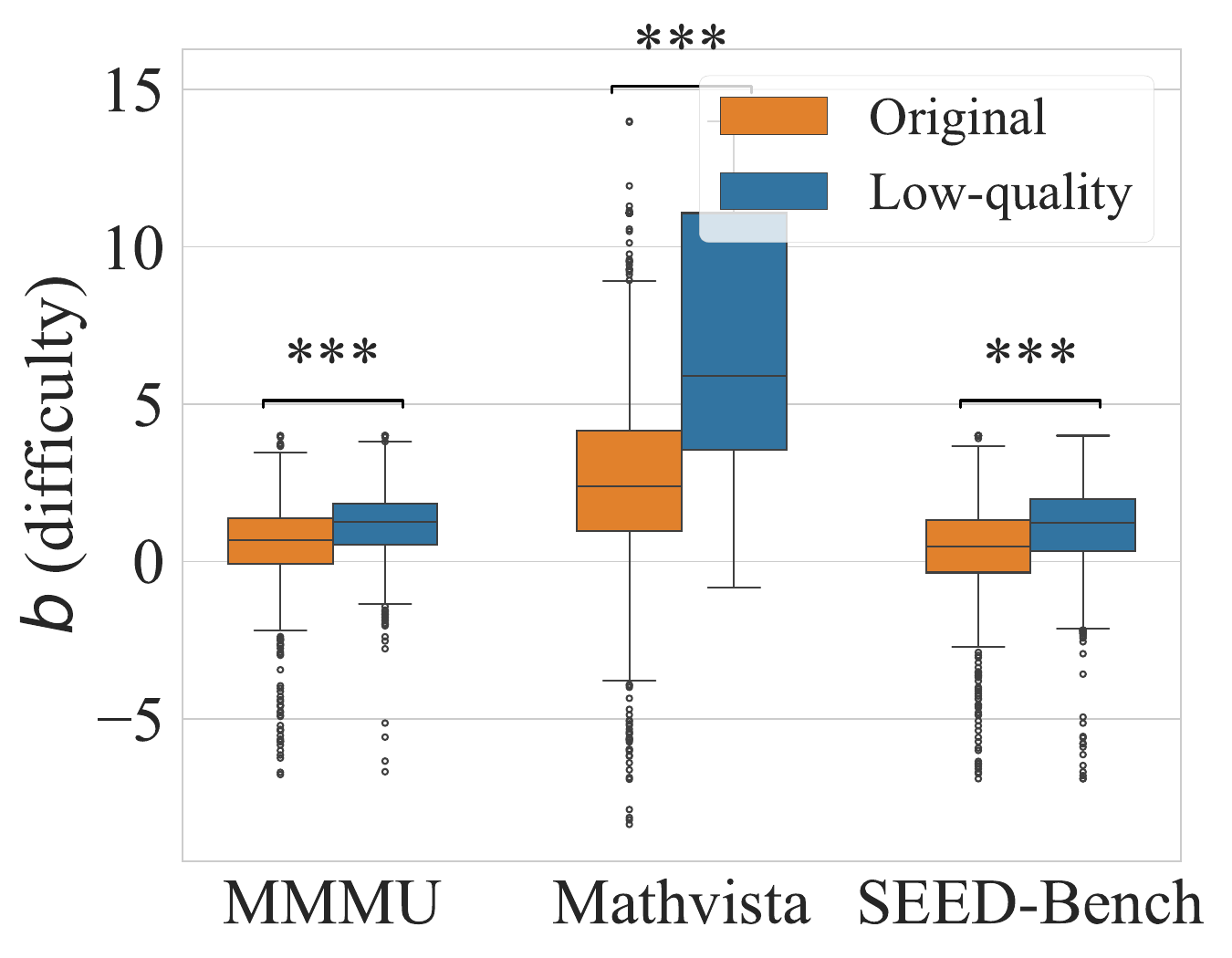}
    \caption{Difficulty}
    \label{fig:difficulty-mmmirt}
    \end{subfigure}
    \begin{subfigure}[b]{0.32\textwidth}
    \includegraphics[width=\linewidth]{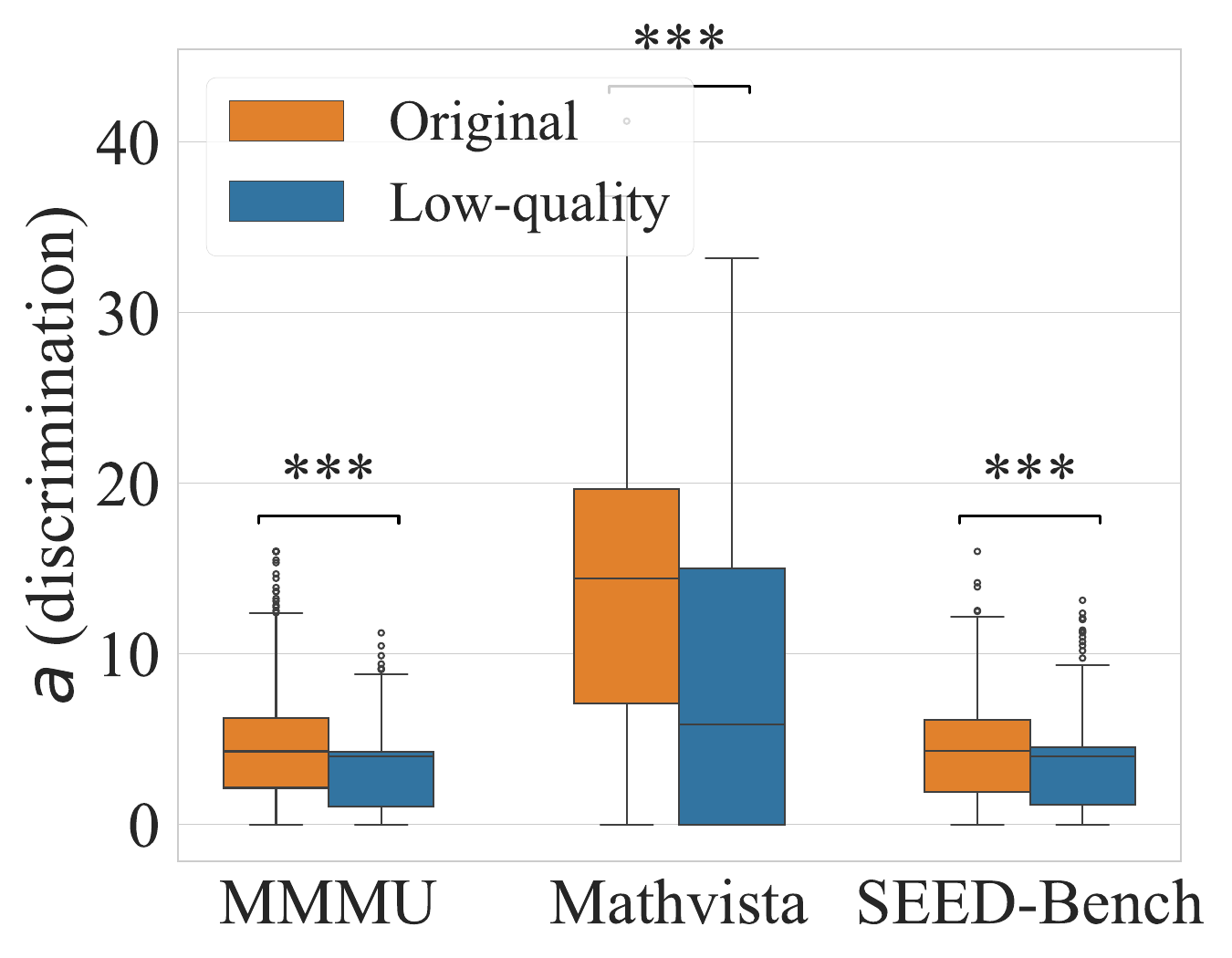}
    \caption{Discrimination}
    \label{fig:discrimination-mmmirt}
    \end{subfigure}
    \begin{subfigure}[b]{0.32\textwidth}
    \includegraphics[width=\linewidth]{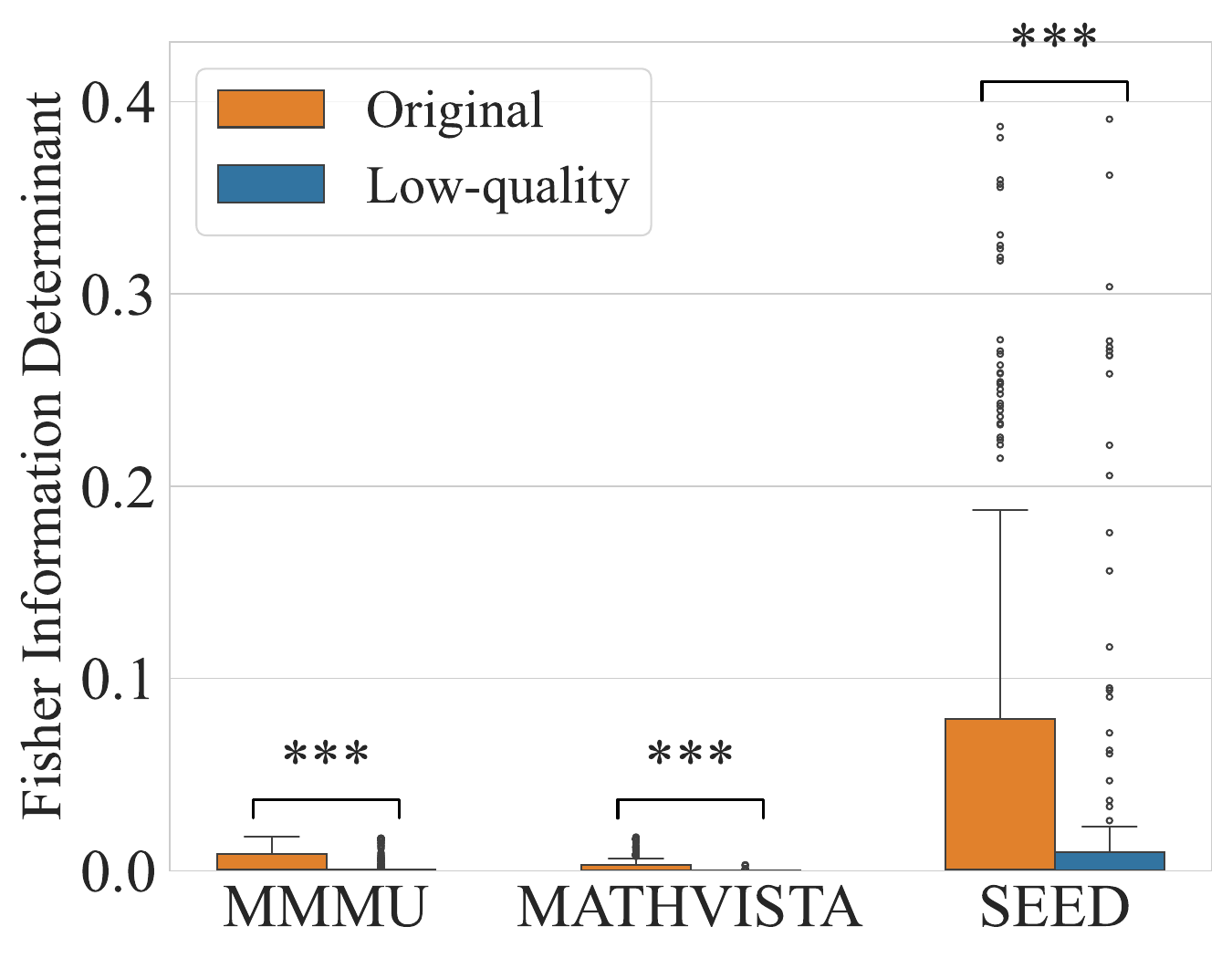}
    \caption{Fisher information}
    \label{fig:fisher-mmmirt}
    \end{subfigure}
    \caption{Comparisons of parameters estimated by \mmmirt{} between the original and artificial questions.}
    \label{fig:param-stats-mmmirt}
\end{figure}
First, we illustrate our problem setting for benchmark refinement in \cref{fig:problem_seting}.
To investigate how the estimated parameters of the original questions and low-quality questions vary, we show the distribution of the estimated difficulty, discrimination, and the Fisher information of the original questions and the low-quality questions in \cref{fig:fisher} and \cref{fig:param-stats-mmmirt}.
We also investigated whether there is a significant difference between the two distributions with the Mann-Whitney U test. Asterisks mark the Mann-Whitney U test results comparing the original questions with the low-quality questions.
\footnote{Significance follows: $^{*}p < 0.05$, $^{**}p < 0.01$, $^{***}p < 0.001$.}
We confirmed significant differences between the groups of original and low-quality questions.

We additionally examined the Wasserstein distance between original and low-quality questions, and found that the Wasserstein distance for \textsc{MMMU}, \textsc{MathVista}, and \textsc{SEED-Bench} were $0.20$, $0.14$, and $0.051$, respectively.
\subsection{Detailed Results}
\SU{Detailed results of the experiments depicted in \cref{fig:subset-rank-comp} are reported in \cref{tab:spearman_detail}.
Detailed results of the experiments depicted in \cref{fig:subset-contami-tate} are reported in \cref{tab:gamma_detail}.}
\begin{table}[htbp]
\centering
\caption{The average and standard deviation of Spearman's rank correlations between model rankings on the original benchmark and those estimated on extracted question subsets with 10\%, 20\%, 30\%, 40\%, and 50\% of whole dataset.}
\label{tab:spearman_detail}
\scalebox{0.8}{
\begin{tabular}{llccccc}
\toprule
\textbf{Benchmark} & \textbf{Method} & \textbf{10\%} & \textbf{20\%} & \textbf{30\%} & \textbf{40\%} & \textbf{50\%} \\
\midrule
\multirow{7}{*}{MMMU} 
 & \mmirt{}       & 0.96 $\pm$ 0.061 & 0.96 $\pm$ 0.057 & 0.96 $\pm$ 0.047 & 0.97 $\pm$ 0.042 & 0.96 $\pm$ 0.035 \\
 & \mmmirt{}       & 0.89 $\pm$ 0.046 & 0.92 $\pm$ 0.030 & 0.92 $\pm$ 0.033 & 0.93 $\pm$ 0.023 & 0.93 $\pm$ 0.023 \\
 & IRT             & 0.50 $\pm$ 0.21 & 0.36 $\pm$ 0.25 & 0.38 $\pm$ 0.20 & 0.35 $\pm$ 0.23 & 0.12 $\pm$ 0.28 \\
 & MIRT            & 0.16 $\pm$ 0.089 & 0.33 $\pm$ 0.091 & 0.47 $\pm$ 0.11 & 0.61 $\pm$ 0.076 & 0.69 $\pm$ 0.055 \\
 & TinyBenchmarks  & 0.43 $\pm$ 0.13 & 0.50 $\pm$ 0.12 & 0.56 $\pm$ 0.11 & 0.61 $\pm$ 0.11 & 0.67 $\pm$ 0.082 \\
 & FlashEval       & 0.79 $\pm$ 0.029 & 0.79 $\pm$ 0.025 & 0.79 $\pm$ 0.024 & 0.80 $\pm$ 0.017 & 0.80 $\pm$ 0.015 \\
 & Random          & 0.77 $\pm$ 0.062 & 0.80 $\pm$ 0.027 & 0.82 $\pm$ 0.028 & 0.82 $\pm$ 0.029 & 0.81 $\pm$ 0.024 \\
\midrule
\multirow{7}{*}{MathVista} 
 & \mmirt{}       & 0.81 $\pm$ 0.036 & 0.84 $\pm$ 0.032 & 0.88 $\pm$ 0.042 & 0.91 $\pm$ 0.034 & 0.93 $\pm$ 0.017 \\
 & \mmmirt{}       & 0.92 $\pm$ 0.028 & 0.94 $\pm$ 0.018 & 0.93 $\pm$ 0.022 & 0.93 $\pm$ 0.018 & 0.93 $\pm$ 0.010 \\
 & IRT             & 0.81 $\pm$ 0.047 & 0.88 $\pm$ 0.040 & 0.91 $\pm$ 0.023 & 0.93 $\pm$ 0.017 & 0.94 $\pm$ 0.014 \\
 & MIRT            & 0.58 $\pm$ 0.049 & 0.67 $\pm$ 0.038 & 0.72 $\pm$ 0.029 & 0.76 $\pm$ 0.026 & 0.81 $\pm$ 0.022 \\
 & TinyBenchmarks  & 0.79 $\pm$ 0.038 & 0.84 $\pm$ 0.020 & 0.86 $\pm$ 0.018 & 0.88 $\pm$ 0.017 & 0.88 $\pm$ 0.011 \\
 & FlashEval       & 0.89 $\pm$ 0.020 & 0.91 $\pm$ 0.015 & 0.91 $\pm$ 0.010 & 0.91 $\pm$ 0.014 & 0.91 $\pm$ 0.008 \\
 & Random          & 0.89 $\pm$ 0.040 & 0.91 $\pm$ 0.029 & 0.92 $\pm$ 0.018 & 0.93 $\pm$ 0.015 & 0.93 $\pm$ 0.013 \\
\midrule
\multirow{7}{*}{SEED-Bench} 
 & \mmirt{}       & 0.94 $\pm$ 0.008 & 0.96 $\pm$ 0.006 & 0.97 $\pm$ 0.0080 & 0.97 $\pm$ 0.005 & 0.95 $\pm$ 0.005 \\
 & \mmmirt{}       & 0.94 $\pm$ 0.019 & 0.95 $\pm$ 0.013 & 0.95 $\pm$ 0.011 & 0.95 $\pm$ 0.018 & 0.95 $\pm$ 0.017 \\
 & IRT             & 0.86 $\pm$ 0.13 & 0.83 $\pm$ 0.18 & 0.84 $\pm$ 0.15 & 0.81 $\pm$ 0.18 & 0.79 $\pm$ 0.17 \\
 & MIRT            & 0.42 $\pm$ 0.120 & 0.59 $\pm$ 0.074 & 0.71 $\pm$ 0.052 & 0.77 $\pm$ 0.045 & 0.82 $\pm$ 0.035 \\
 & TinyBenchmarks  & 0.69 $\pm$ 0.057 & 0.75 $\pm$ 0.050 & 0.77 $\pm$ 0.056 & 0.80 $\pm$ 0.049 & 0.81 $\pm$ 0.044 \\
 & FlashEval       & 0.89 $\pm$ 0.021 & 0.90 $\pm$ 0.015 & 0.90 $\pm$ 0.013 & 0.90 $\pm$ 0.010 & 0.90 $\pm$ 0.009 \\
 & Random          & 0.86 $\pm$ 0.046 & 0.87 $\pm$ 0.032 & 0.89 $\pm$ 0.027 & 0.91 $\pm$ 0.025 & 0.91 $\pm$ 0.011 \\
\bottomrule
\end{tabular}
}
\end{table}

\begin{table}[htbp]
\centering
\caption{The average and standard deviation of the proportions of the low-quality questions in extracted question subsets $\gamma$ with 10\%, 20\%, 30\%, 40\%, and 50\%  of whole dataset. "TB" means TinyBenchmarks, and "FE" means FlashEval.}
\label{tab:gamma_detail}
\scalebox{0.8}{
\begin{tabular}{llccccc}
\toprule
\textbf{Benchmark} & \textbf{Method} & \textbf{10\%} & \textbf{20\%} & \textbf{30\%} & \textbf{40\%} & \textbf{50\%} \\
\midrule
\multirow{7}{*}{MMMU} 
 & \mmirt{}       & 0.038 $\pm$ 0.015 & 0.090 $\pm$ 0.014 & 0.12 $\pm$ 0.014 & 0.16 $\pm$ 0.014 & 0.19 $\pm$ 0.013 \\
 & \mmmirt{}       & 0.091 $\pm$ 0.017 & 0.11 $\pm$ 0.013 & 0.14 $\pm$ 0.012 & 0.16 $\pm$ 0.011 & 0.19 $\pm$ 0.0070 \\
 & IRT             & 0.32 $\pm$ 0.040 & 0.34 $\pm$ 0.035 & 0.36 $\pm$ 0.029 & 0.37 $\pm$ 0.023 & 0.38 $\pm$ 0.020 \\
 & MIRT            & 0.42 $\pm$ 0.020 & 0.40 $\pm$ 0.016 & 0.38 $\pm$ 0.012 & 0.37 $\pm$ 0.012 & 0.36 $\pm$ 0.011 \\
 & TB  & 0.38 $\pm$ 0.023 & 0.39 $\pm$ 0.021 & 0.38 $\pm$ 0.022 & 0.37 $\pm$ 0.021 & 0.36 $\pm$ 0.020 \\
 & FE       & 0.28 $\pm$ 0.014 & 0.29 $\pm$ 0.017 & 0.29 $\pm$ 0.021 & 0.27 $\pm$ 0.017 & 0.29 $\pm$ 0.010 \\
 & Random          & 0.31 $\pm$ 0.031 & 0.30 $\pm$ 0.015 & 0.30 $\pm$ 0.010 & 0.30 $\pm$ 0.0081 & 0.30 $\pm$ 0.0080 \\
\midrule
\multirow{7}{*}{MathVista} 
 & \mmirt{}       & 0.011 $\pm$ 0.0060 & 0.037 $\pm$ 0.014 & 0.068 $\pm$ 0.021 & 0.11 $\pm$ 0.021 & 0.15 $\pm$ 0.016 \\
 & \mmmirt{}       & 0.054 $\pm$ 0.013 & 0.11 $\pm$ 0.014 & 0.15 $\pm$ 0.019 & 0.17 $\pm$ 0.013 & 0.20 $\pm$ 0.0078 \\
 & IRT             & 0.27 $\pm$ 0.055 & 0.25 $\pm$ 0.037 & 0.24 $\pm$ 0.028 & 0.23 $\pm$ 0.020 & 0.22 $\pm$ 0.018 \\
 & MIRT            & 0.50 $\pm$ 0.024 & 0.43 $\pm$ 0.023 & 0.39 $\pm$ 0.018 & 0.36 $\pm$ 0.013 & 0.33 $\pm$ 0.0083 \\
 & TB  & 0.33 $\pm$ 0.027 & 0.31 $\pm$ 0.019 & 0.30 $\pm$ 0.015 & 0.29 $\pm$ 0.010 & 0.28 $\pm$ 0.0065 \\
 & FE       & 0.27 $\pm$ 0.016 & 0.28 $\pm$ 0.018 & 0.30 $\pm$ 0.016 & 0.31 $\pm$ 0.015 & 0.30 $\pm$ 0.014 \\
 & Random          & 0.29 $\pm$ 0.031 & 0.29 $\pm$ 0.017 & 0.29 $\pm$ 0.0086 & 0.29 $\pm$ 0.0074 & 0.29 $\pm$ 0.0070 \\
\midrule
\multirow{7}{*}{SEED-Bench} 
 & \mmirt{}       & 0.045 $\pm$ 0.016 & 0.086 $\pm$ 0.011 & 0.13 $\pm$ 0.0074 & 0.17 $\pm$ 0.0062 & 0.22 $\pm$ 0.0037 \\
 & \mmmirt{}       & 0.14 $\pm$ 0.018 & 0.14 $\pm$ 0.012 & 0.18 $\pm$ 0.011 & 0.21 $\pm$ 0.0091 & 0.24 $\pm$ 0.0080 \\
 & IRT             & 0.31 $\pm$ 0.048 & 0.33 $\pm$ 0.041 & 0.34 $\pm$ 0.040 & 0.35 $\pm$ 0.038 & 0.37 $\pm$ 0.031 \\
 & MIRT            & 0.43 $\pm$ 0.023 & 0.41 $\pm$ 0.013 & 0.39 $\pm$ 0.012 & 0.38 $\pm$ 0.0083 & 0.37 $\pm$ 0.0087 \\
 & TB  & 0.36 $\pm$ 0.015 & 0.35 $\pm$ 0.013 & 0.35 $\pm$ 0.012 & 0.34 $\pm$ 0.0082 & 0.34 $\pm$ 0.0081 \\
 & FE       & 0.30 $\pm$ 0.019 & 0.31 $\pm$ 0.020 & 0.33 $\pm$ 0.014 & 0.34 $\pm$ 0.015 & 0.34 $\pm$ 0.016 \\
 & Random          & 0.34 $\pm$ 0.036 & 0.34 $\pm$ 0.021 & 0.34 $\pm$ 0.011 & 0.34 $\pm$ 0.0084 & 0.34 $\pm$ 0.0084 \\
\bottomrule
\end{tabular}
}
\end{table}

\subsection{VQAAT}
\textsc{\textbf{VQA-AnswerTherapy} (VQAAT)}~\citep{chen2023vqa} consists of VizWiz Dataset~\citep{gurari2018vizwiz}, which is visual questions asked by visually impaired people, and VQA v2.0~\citep{balanced_vqa_v2}. We randomly sample $1000$ questions from the train and validation sets of Single Answer Grounding Challenge.
This dataset presents images, questions, and multiple annotators' responses to those questions to the VLM, asking whether the annotators' answers are based on the same part of the image.
Therefore, this dataset consists solely of binary-choice questions.

We conducted an additional experiment on VQAAT under the same condition as \cref{subsec:subset}.
\Cref{fig:subset-rank-comp} shows the Spearman's rank correlations between the model rankings on the original benchmark and on an extracted subset and proportions of the low-quality questions.
In contrast to experiment in \cref{subsec:subset}, on \textsc{VQAAT}, \mmirt{} and \mmmirt{} are worse than baselines in terms of the Spearman's rank correlation.
Interestingly, while proposed methods extracts fewer low-quality questions than the baselines, this filtering does not translate to improved accuracy in model ranking.
We hypothesize that the discrepancy arises because \textsc{VQAAT} itself contains numerous low-quality questions, significantly influencing its "ground truth" ranking.
As our method filters such low-quality questions, the resulting ranking deviates from the original benchmark.

\begin{figure*}
    \centering
    \begin{subfigure}[b]{0.47\textwidth}
        \centering
        \includegraphics[width=\textwidth]{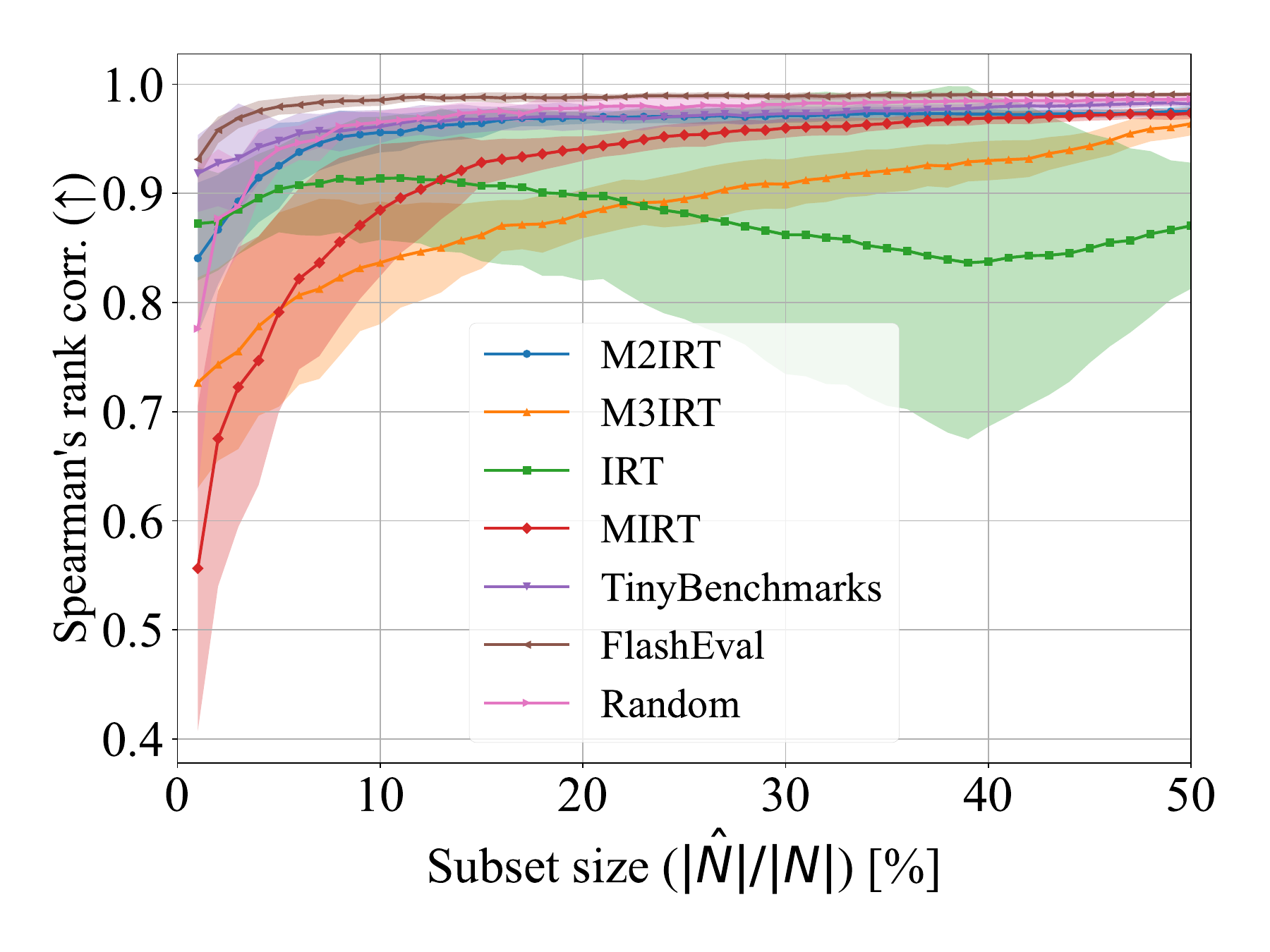}
        \caption{Spearman's rank correlations on \textsc{VQAAT}}
        \label{fig:subset-rank-comp-VQAAT}
    \end{subfigure}
    \begin{subfigure}[b]{0.47\textwidth}
        \centering
        \includegraphics[width=\textwidth]{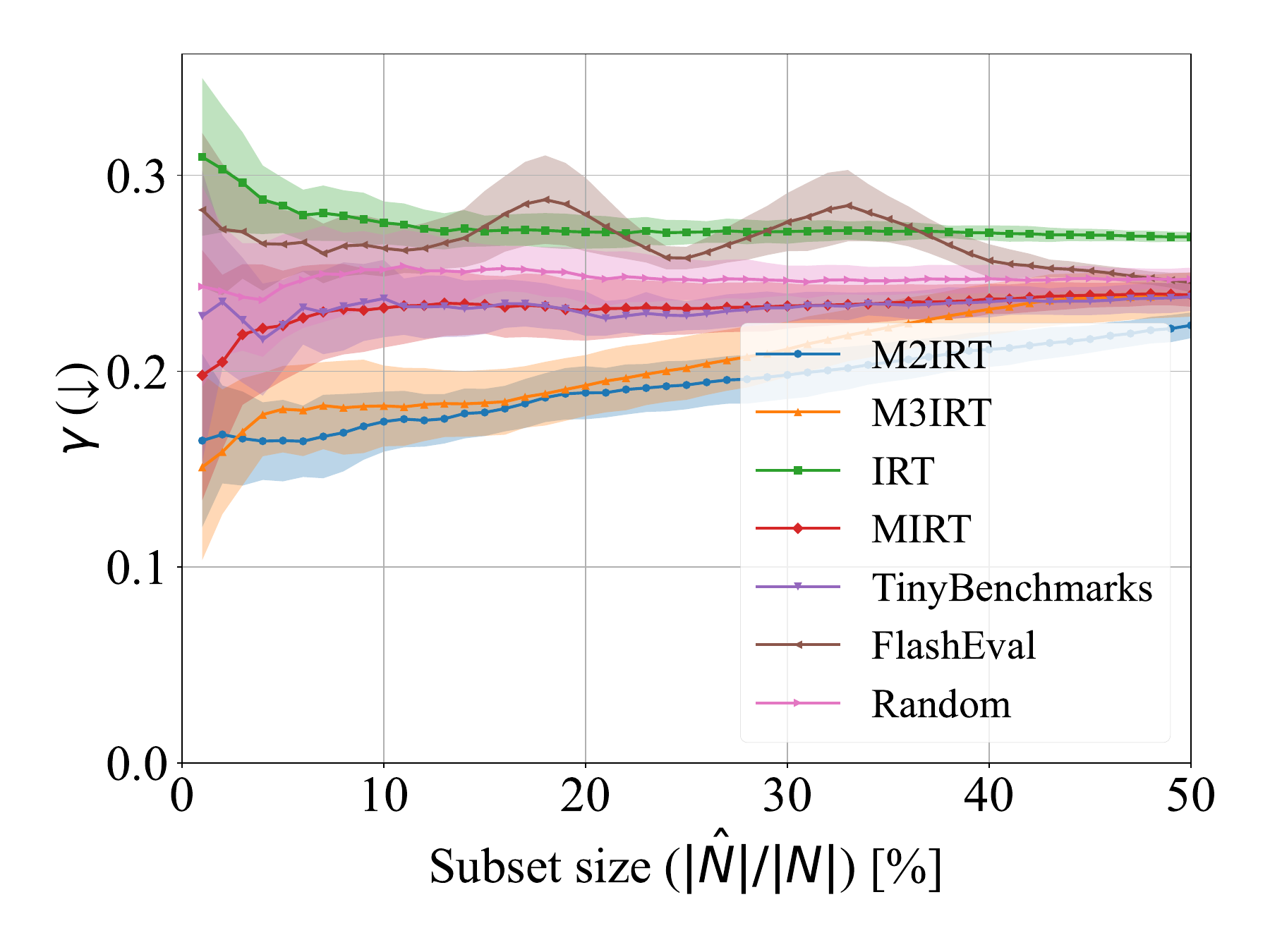}
        \caption{The proportions of the low-quality questions $\gamma$ on \textsc{VQAAT}}
        \label{fig:subset-contami-tate-VQAAT}
    \end{subfigure}
    \caption{The average and standard deviation of Spearman's rank correlations on extracted question subsets of VQAAT and the proportions of the low-quality questions $\gamma$ in extracted question subsets  with different sizes.}
    \label{fig:vqaat-subset}
\end{figure*}
\subsection{For Sparse Response Matrix}
\begin{figure*}
    \centering
    \begin{subfigure}[b]{0.32\textwidth}
        \centering
        \includegraphics[width=\textwidth]{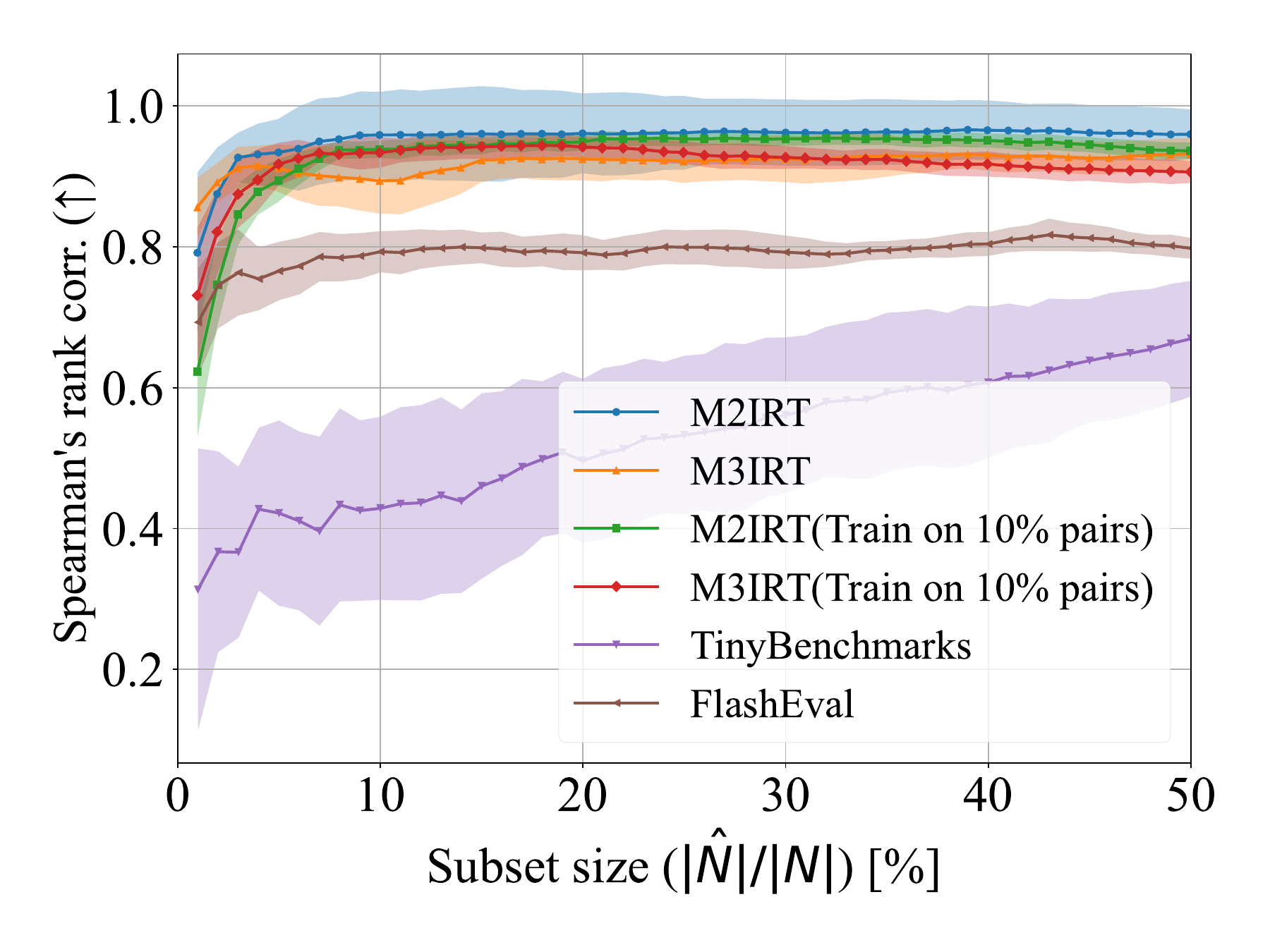}
        \caption{\textsc{MMMU}}
        \label{fig:subset-rank-comp-mmmu-sparse}
    \end{subfigure}
    \begin{subfigure}[b]{0.32\textwidth}
        \centering
        \includegraphics[width=\textwidth]{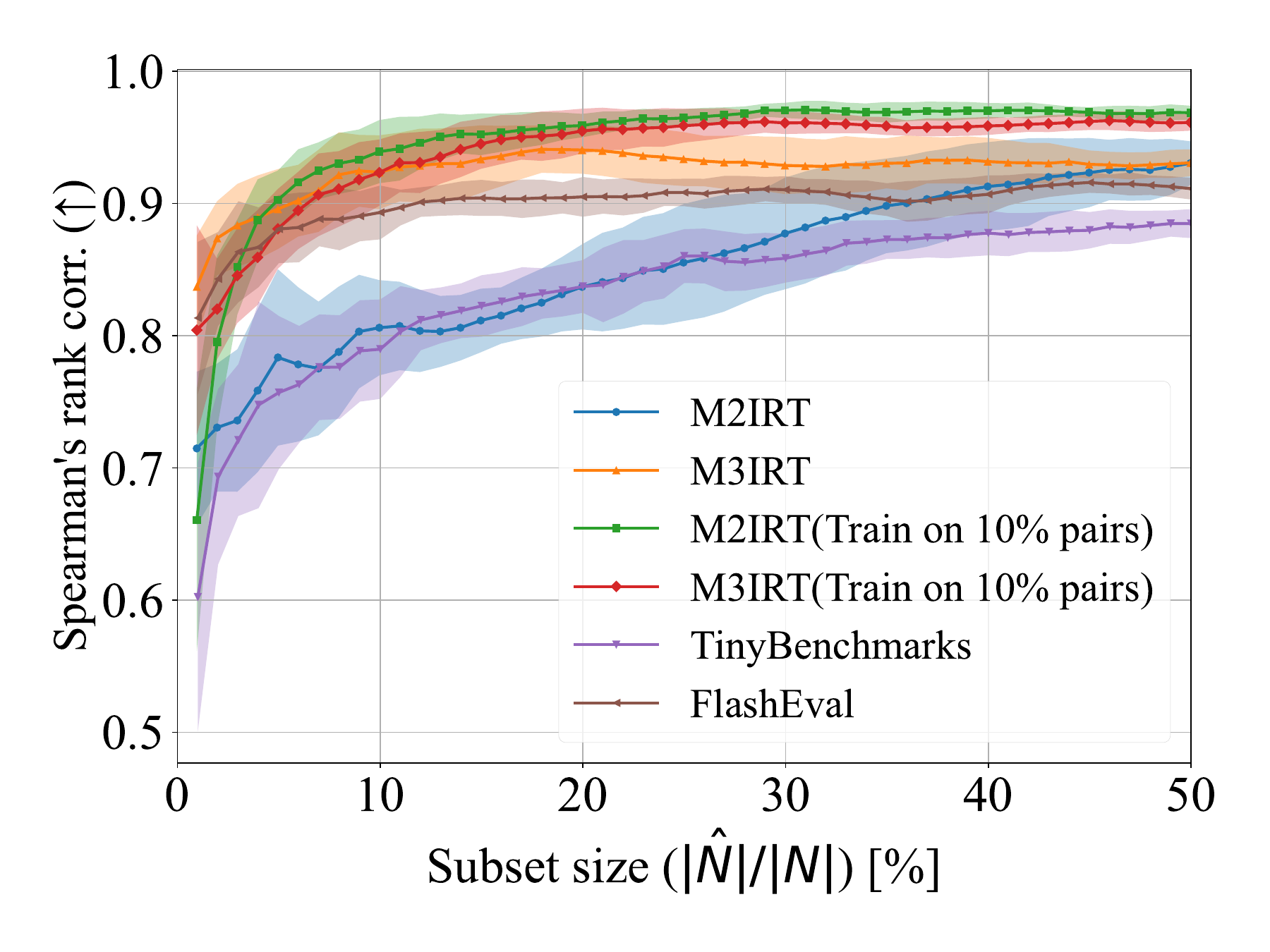}
        \caption{\textsc{MathVista}}
        \label{fig:subset-rank-comp-mathvista-sparse}
    \end{subfigure}
    \begin{subfigure}[b]{0.32\textwidth}
        \centering
        \includegraphics[width=\textwidth]{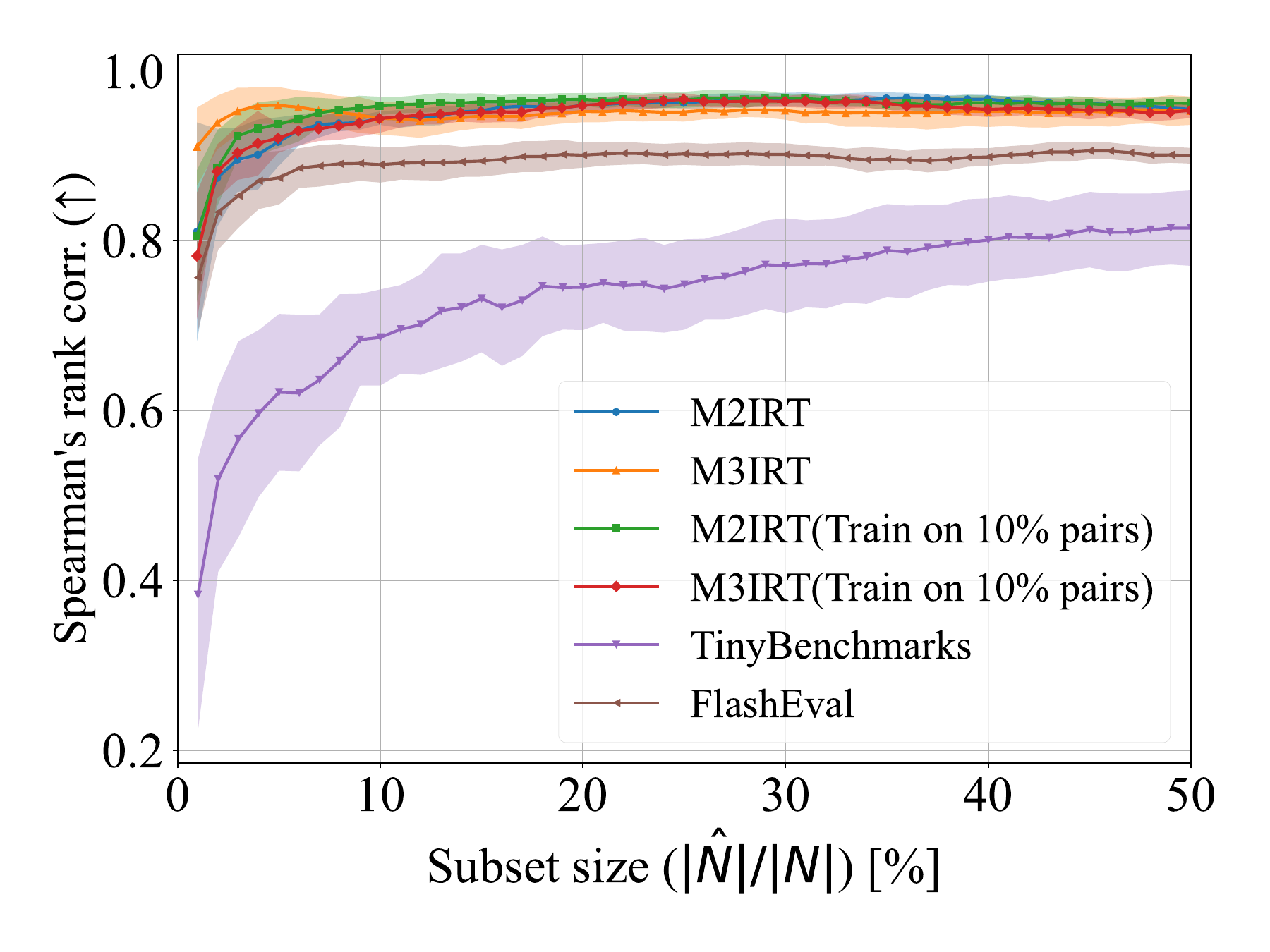}
        \caption{\textsc{SEED-Bench}}
        \label{fig:subset-rank-comp-seed-sparse}
    \end{subfigure}
    \caption{The average and standard deviation of Spearman's rank correlations between model rankings on the original benchmark and those estimated on extracted question subsets with different sizes.\mmirt{} and \mmmirt{} are trained with sparse-response matrix.}
    \label{fig:subset-rank-comp-sparse}
\end{figure*}
\begin{figure*}
    \centering
    \begin{subfigure}[b]{0.32\textwidth}
        \centering
        \includegraphics[width=\textwidth]{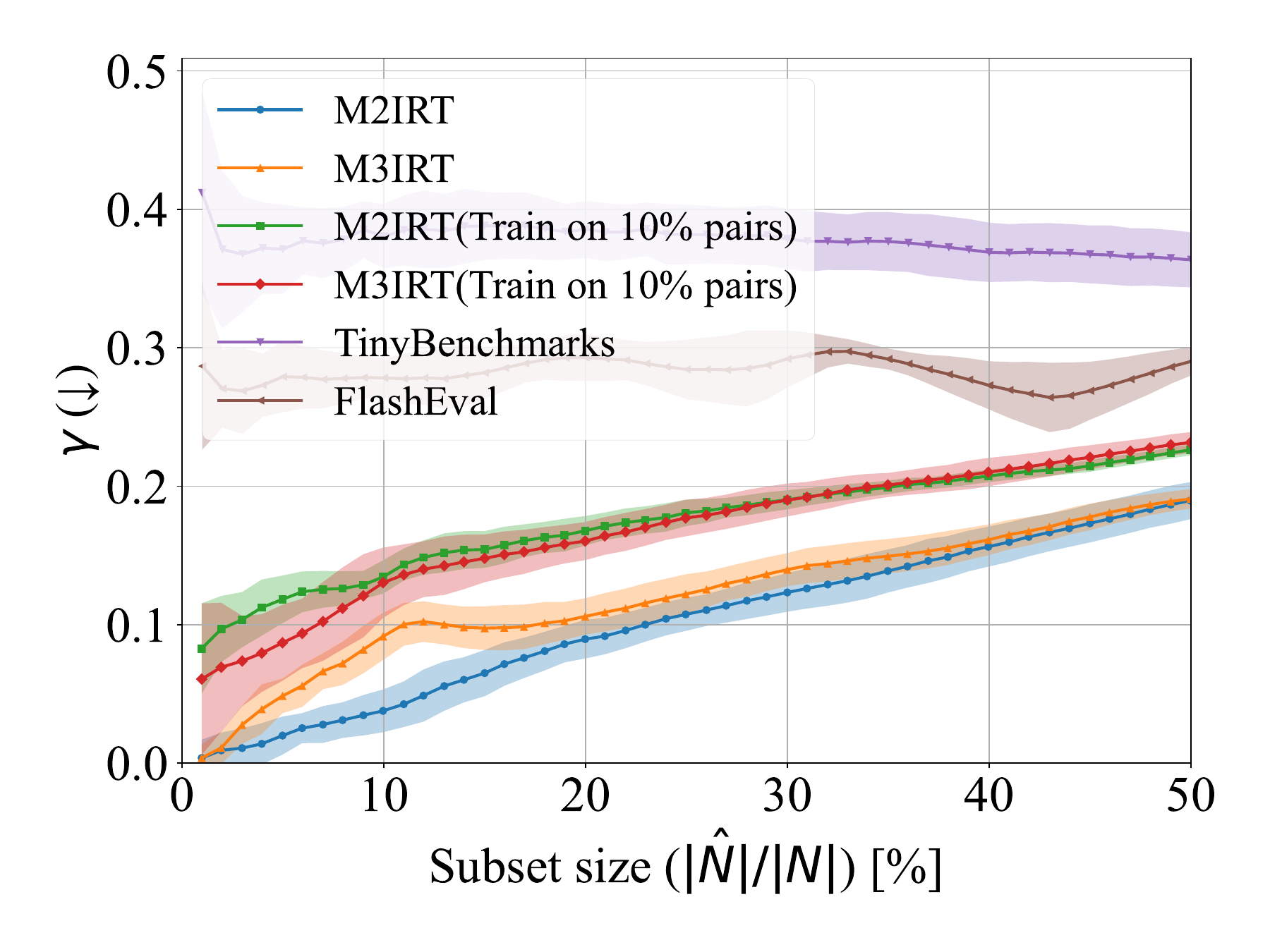}
        \caption{\textsc{MMMU}}
        \label{fig:subset-contami-tate-mmmu-sparse}
    \end{subfigure}
    \begin{subfigure}[b]{0.32\textwidth}
        \centering
        \includegraphics[width=\textwidth]{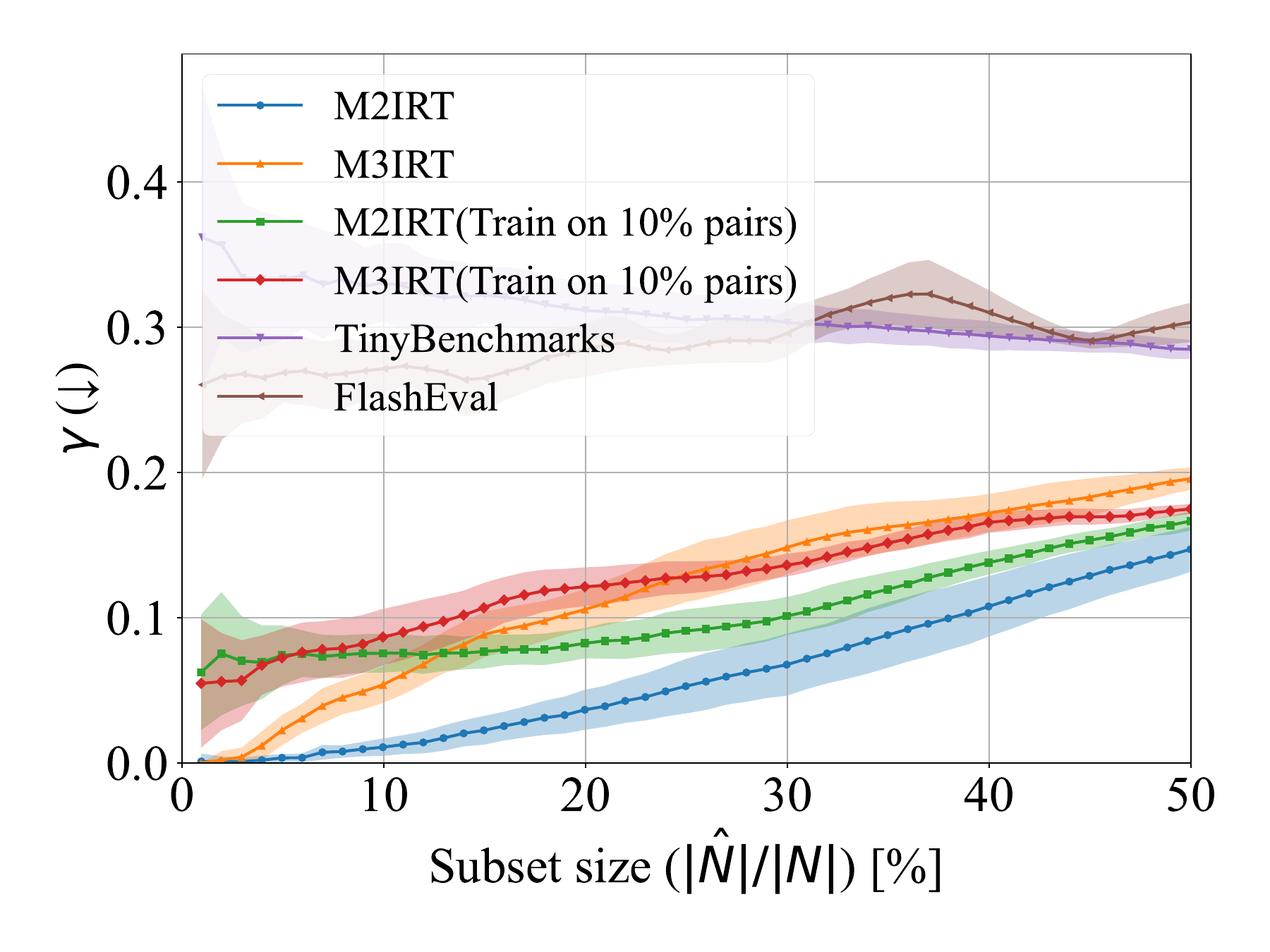}
        \caption{\textsc{MathVista}}
        \label{fig:subset-contami-tate-mathvista-sparse}
    \end{subfigure}
    \begin{subfigure}[b]{0.32\textwidth}
        \centering
        \includegraphics[width=\textwidth]{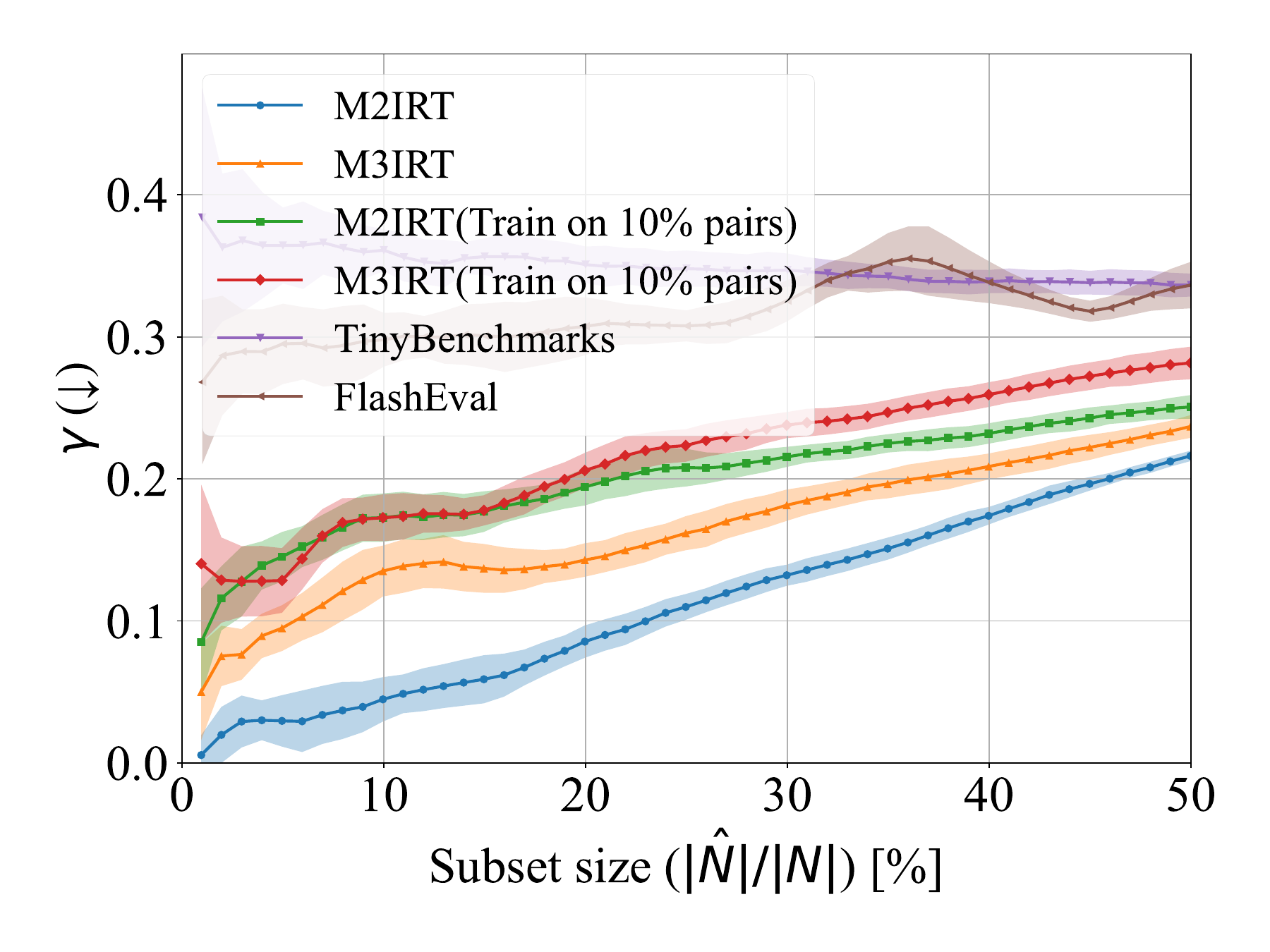}
        \caption{\textsc{SEED-Bench}}
        \label{fig:subset-contami-tate-seed-sparse}
    \end{subfigure}
    \caption{The average and standard deviation of the proportions of the low-quality questions in extracted question subsets $\gamma$ with different sizes. \mmirt{} and \mmmirt{} are trained with sparse-response matrix.}
    \label{fig:subset-contami-tate-sparse}
\end{figure*}
In \cref{subsec:learning-mmirt}, we explain that \mmirt{} and \mmmirt{} don't require all models to respond to all questions.
To demonstrate this, we conducted an additional experiment identical to Section 5.3, except that we used only 10\% of all (model, question) pairs on MMMU for training \mmirt{} and \mmmirt{}. 
We then selected informative questions for evaluating a new model.
These models trained with sparse dataset are compared with original and strong baselines.

\Cref{fig:subset-rank-comp-sparse} shows the Spearman's rank correlations between the model rankings on the original benchmark and on different sizes of subsets.
\Cref{fig:subset-contami-tate-sparse} shows the proportion $\gamma$ with varying size of subsets.

Remarkably, as shown in \cref{fig:subset-rank-comp-mmmu-sparse} using just 3\% of the total questions, \mmmirt{} trained only 10\% of all (model, question) pairs achieved a Spearman rank correlation exceeding 0.84 with the original full-dataset rankings—equivalent to the baseline performance that requires 50\% of the dataset. 
Beyond this point, \mmirt{} consistently maintained higher ranking consistency than all baselines. Moreover, the proportion of low-quality questions selected remained below 23\%, showing that \mmirt{} is not only efficient but also discriminative in identifying high-quality items.

This setting allows model comparison at only 13\% of the inference cost required for full evaluation across all models and questions, demonstrating that \mmirt{} offers substantial cost savings while preserving evaluation reliability.
\subsection{Statistical Significance Tests}
We conducted a one-sided Wilcoxon signed-rank test to evaluate the performance difference between \mmirt{} and FlashEval,\mmirt{} and TinyBenchmarks, \mmmirt{} and FlashEval,and \mmmirt{} and TinyBenchmarks. 
\Cref{tab:sigtest-mmmu}, \cref{tab:sigtest-mathvista}, and \cref{tab:sigtest-seed} show the results for 5\%, 10\%, 30\%, and 50\% of \cref{fig:subset-rank-comp} and \cref{fig:subset-contami-tate} with a confidence level of 1\%.
\textsc{MMMU} shows significant differences from the baseline method. 
\textsc{Mathvista} shows significant differences in all conditions except for Spearman's rank corr at 5\% against FlashEval's score.
\textsc{SEED-Bench} shows significant differences from the baseline method. 

\begin{table}[htbp]
\centering
\caption{Wilcoxon signed‐rank test on MMMU comparing FlashEval and TinyBench against \mmmirt{}.}
\label{tab:sigtest-mmmu}
\scalebox{0.8}{
\begin{tabular}{|l|cc|cc|cc|cc|}
\hline
Comparison 
  & \multicolumn{2}{c|}{5\% subset}
  & \multicolumn{2}{c|}{10\% subset}
  & \multicolumn{2}{c|}{30\% subset}
  & \multicolumn{2}{c|}{50\% subset} \\
\hline
  & $p$‐value & $W$
  & $p$‐value & $W$
  & $p$‐value & $W$
  & $p$‐value & $W$ \\
\hline
vs FlashEval (Rank corr.)  
  & $<0.0001$ & 0.0
  & $<0.0001$ & 0.0
  & $<0.0001$ & 0.0
  & $<0.0001$ & 0.0 \\
vs TinyBench (Rank corr.) 
  & $<0.0001$ & 0.0
  & $<0.0001$ & 0.0
  & $<0.0001$ & 0.0
  & $<0.0001$ & 0.0 \\
\hline
vs FlashEval (Shuffle ratio)  
  & $<0.0001$ & 0.0
  & $<0.0001$ & 0.0
  & $<0.0001$ & 0.0
  & $<0.0001$ & 0.0 \\
vs TinyBench (Shuffle ratio) 
  & $<0.0001$ & 0.0
  & $<0.0001$ & 0.0
  & $<0.0001$ & 0.0
  & $<0.0001$ & 0.0 \\
\hline
\end{tabular}
}
\end{table}

\begin{table}[htbp]
\centering
\caption{Wilcoxon signed‐rank test on \textsc{MathVista} comparing FlashEval and TinyBench against \mmmirt{}.}
\label{tab:sigtest-mathvista}
\scalebox{0.8}{
\begin{tabular}{|l|cc|cc|cc|cc|}
\hline
Comparison 
  & \multicolumn{2}{c|}{5\% subset}
  & \multicolumn{2}{c|}{10\% subset}
  & \multicolumn{2}{c|}{30\% subset}
  & \multicolumn{2}{c|}{50\% subset} \\
\hline
  & $p$‐value & $W$
  & $p$‐value & $W$
  & $p$‐value & $W$
  & $p$‐value & $W$ \\
\hline
vs FlashEval(Rank Corr.)
  & 0.0197 & 222.0
  & 0.0004 & 262.0
  & 0.0019 & 251.5
  & $<0.0001$ & 293.0 \\
vs TinyBench(Rank Corr.) 
  & $<0.0001$ & 300.0
  & $<0.0001$ & 300.0
  & $<0.0001$ & 300.0
  & $<0.0001$ & 300.0 \\
\hline
vs FlashEval(Shuffle ratio)  
  & $<0.0001$ & 0.0
  & $<0.0001$ & 0.0
  & $<0.0001$ & 0.0
  & $<0.0001$ & 0.0 \\
vs TinyBench(Shuffle ratio)  
  & $<0.0001$ & 0.0
  & $<0.0001$ & 0.0
  & $<0.0001$ & 0.0
  & $<0.0001$ & 0.0 \\
\hline
\end{tabular}
}
\end{table}

\begin{table}[htbp]
\centering
\caption{Wilcoxon signed‐rank test on SEEDBench comparing FlashEval and TinyBench against \mmmirt{}.}
\label{tab:sigtest-seed}
\scalebox{0.8}{
\begin{tabular}{|l|cc|cc|cc|cc|}
\hline
Comparison 
  & \multicolumn{2}{c|}{5\% subset}
  & \multicolumn{2}{c|}{10\% subset}
  & \multicolumn{2}{c|}{30\% subset}
  & \multicolumn{2}{c|}{50\% subset} \\
\hline
  & $p$‐value & $W$
  & $p$‐value & $W$
  & $p$‐value & $W$
  & $p$‐value & $W$ \\
\hline
vs FlashEval (Rank corr.)  
  & $<0.0001$ & 231.0
  & $<0.0001$ & 231.0
  & $<0.0001$ & 231.0
  & $<0.0001$ & 231.0 \\
vs TinyBench (Rank corr.) 
  & $<0.0001$ & 231.0
  & $<0.0001$ & 231.0
  & $<0.0001$ & 231.0
  & $<0.0001$ & 231.0 \\
\hline
vs FlashEval (Shuffle ratio)  
  & $<0.0001$ & 0.0
  & $<0.0001$ & 0.0
  & $<0.0001$ & 0.0
  & $<0.0001$ & 0.0 \\
vs TinyBench (Shuffle ratio) 
  & $<0.0001$ & 0.0
  & $<0.0001$ & 0.0
  & $<0.0001$ & 0.0
  & $<0.0001$ & 0.0 \\
\hline
\end{tabular}
}
\end{table}


\section{Details of Experimental Settings}\label{appendix:experiment}
\subsection{Computational resources}
The computational resources utilized in this study are presented in \cref{tab:computer_specs}.
\begin{table}[tb]
  \centering
  \caption{Computer Specifications Used for Experiments}
  \begin{tabular}{ll}
    \hline
    \textbf{Component} & \textbf{Specification} \\
    \hline
    Operating System & Ubuntu 20.04 LTS \\
    CPU & AMD EPYC Milan 7763 DP/UP (64C/128T, 2.45GHz) $\times$ 2 \\
    Memory & 2048GB \\
    python version & 3.12.9 \\
    torch version & 2.6.0 \\
    \hline
  \end{tabular}
  \label{tab:computer_specs}
\end{table}
The experiments in \cref{subsec:subset} require 2 hours per dataset, and those in \cref{subsec:robust-auc} necessitate 3 hours per dataset.

\subsection{Datasets}
\textbf{A Massive Multi-discipline Multimodal Understanding and Reasoning Benchmark for Expert AGI} (\textsc{\textbf{MMMU}})~\citep{Yue_2024_CVPR_MMMU} : The license for this dataset is "Apache License 2.0".

\textsc{\textbf{MathVista}}~\citep{lu2024mathvista} : The license for this dataset is "Creative Commons Attribution Share Alike 4.0 International".

\textsc{\textbf{VQA-AnswerTherapy} (VQAAT)}~\citep{chen2023vqa} : The license for this dataset is "Creative Commons Attribution 4.0 International License".
 
\textsc{\textbf{SEED-Bench}}~\citep{Li_2024_CVPR_seed} : The license for this dataset is "Creative Commons Attribution Non Commercial 4.0".
\subsection{VLMs}
We use 24 commonly used VLMs listed in \cref{tab:used-mllms} for our experiments.
We access open-source models and a subset of closed models through Openrouter.
\begin{table}[H]
\centering
\caption{Overview of AI Models Used}
\begin{tabular}{|l|c|l|}
\hline
\textbf{Model Name} & \textbf{Type} & \textbf{License or Terms} \\
\hline
GPT-4-turbo & Closed & \href{https://openai.com/policies/terms-of-use/}{OpenAI Terms of Use} \\
GPT-4o~\citep{openai2024gpt4ocard} & Closed & \href{https://openai.com/policies/terms-of-use/}{OpenAI Terms of Use} \\
GPT-4o-mini~\citep{openai2024gpt4omini} & Closed & \href{https://openai.com/policies/terms-of-use/}{OpenAI Terms of Use} \\
GPT-4.1~\citep{openai2025gpt41} & Closed & \href{https://openai.com/policies/terms-of-use/}{OpenAI Terms of Use} \\
GPT-4.1-mini~\citep{openai2025gpt41} & Closed & \href{https://openai.com/policies/terms-of-use/}{OpenAI Terms of Use}\\
GPT-4.1-nano~\citep{openai2025gpt41} & Closed & \href{https://openai.com/policies/terms-of-use/}{OpenAI Terms of Use} \\
Gemini-1.5-flash~\citep{google2024gemini15} & Closed & \href{https://ai.google.dev/gemini-api/terms}{Gemini API Additional Terms of Service} \\
Gemini-1.5-flash-8b~\citep{google2024gemini15} & Closed & \href{https://ai.google.dev/gemini-api/terms}{Gemini API Additional Terms of Service} \\
Gemini-1.5-pro~\citep{google2024gemini15}  & Closed & \href{https://ai.google.dev/gemini-api/terms}{Gemini API Additional Terms of Service} \\
Gemini-2.0-flash~\citep{pichai2024gemini} & Closed & \href{https://ai.google.dev/gemini-api/terms}{Gemini API Additional Terms of Service} \\
Claude-3-haiku~\citep{anthropic2024claude3}& Closed & \href{https://www.anthropic.com/legal/consumer-terms}{Anthropic Consumer Terms of Service} \\
Claude-3-sonnet~\citep{anthropic2024claude3} & Closed & \href{https://www.anthropic.com/legal/consumer-terms}{Anthropic Consumer Terms of Service} \\
Claude-3.5-sonnet~\citep{anthropic2024claude35} & Closed & \href{https://www.anthropic.com/legal/consumer-terms}{Anthropic Consumer Terms of Service} \\
Claude-3.7-sonnet~\citep{anthropic2025claude37} & Closed & \href{https://www.anthropic.com/legal/consumer-terms}{Anthropic Consumer Terms of Service} \\
Grok-2~\citep{grok2} & Closed & \href{https://x.ai/legal/terms-of-service}{xAI Terms of Service} \\
Nova-pro~\citep{Intelligence2024} & Closed & \href{https://aws.amazon.com/service-terms/?nc1=h_ls}{AWS Terms of Service} \\
Nova-lite~\citep{Intelligence2024} & Closed & \href{https://aws.amazon.com/service-terms/?nc1=h_ls}{AWS Terms of Service} \\
Qwen-2.5-vl-7b~\citep{Qwen2.5-VL} & Open & Apache 2.0 \\
Qwen-2.5-vl-72b~\citep{Qwen2.5-VL} & Open & Apache 2.0 \\
Llama-3.2-11b-instruct~\citep{meta2024llama} & Open & Llama 3.2 Community License  \\
Llama-3.2-90b-instruct~\citep{meta2024llama} & Open & Llama 3.2 Community License \\
Pixtral-12b~\citep{agrawal2024pixtral} & Open & Apache 2.0 \\
Pixtral-large~\citep{agrawal2024pixtral} & Open & Apache 2.0 \\
Minimax-01~\citep{minimax2025minimax01scalingfoundationmodels} & Open & MIT License \\
\hline
\end{tabular}
\label{tab:used-mllms}
\end{table}

\end{document}